\pdfoutput=1

\documentclass{article}

 \usepackage[preprint]{neurips_2026}
  

\usepackage[T1]{fontenc}
\usepackage[utf8]{inputenc}

\usepackage{latexsym}
\usepackage{inconsolata}
\usepackage{graphicx}
\usepackage{microtype}
\usepackage{url}
\usepackage{booktabs}
\usepackage[export]{adjustbox}
\usepackage{multirow}
\usepackage{enumitem}
\usepackage{subcaption}
\usepackage{longtable}
\usepackage{float}
\usepackage{hyperref}
\usepackage{xurl}
\usepackage{amsfonts}
\usepackage{nicefrac}
\usepackage{xcolor}
\usepackage{natbib}
\usepackage{xspace}
\usepackage{amsmath}
\usepackage{gensymb}
\usepackage{colortbl}
\usepackage{amssymb}
\usepackage{mathtools}
\usepackage{amsthm}
\usepackage{stfloats}

\usepackage[autostyle=false,style=english]{csquotes}
\MakeOuterQuote{"}

\usepackage[capitalize,noabbrev]{cleveref}
\crefname{section}{Sec.}{Secs.}
\crefname{table}{Table}{Tables}
\crefname{figure}{Fig.}{Figs.}
\crefname{app}{App.}{Apps.}
\crefformat{footnote}{#2\footnotemark[#1]#3}

\usepackage{tikz}
\usetikzlibrary{arrows.meta, positioning, shadows.blur}
\definecolor{xcol}   {HTML}{4269D0}
\definecolor{fxcol}  {HTML}{EFB118}
\definecolor{gxcol}  {HTML}{A463F2}
\definecolor{gfxcol} {HTML}{3CA951}
\definecolor{fgxcol}{HTML}{FF725C}
\pgfdeclarelayer{bg}
\pgfsetlayers{bg,main}

\usetikzlibrary{fit,calc}

\title{How Do Language Models Compose Functions?}

\author{%
  Apoorv Khandelwal \qquad Ellie Pavlick \\
  Department of Computer Science\\
  Brown University\\
  \texttt{\{apoorvkh,ellie\_pavlick\}@brown.edu}
}

\begin{document}

\maketitle

\begin{abstract}
While large language models (LLMs) appear to be increasingly capable of solving compositional tasks, it is an open question whether they do so using compositional mechanisms. In this work, we investigate how feedforward LLMs solve two-hop factual recall tasks, which can be expressed compositionally as $g(f(x))$. We first confirm that modern LLMs continue to suffer from the "compositionality gap", i.e. their ability to compute both $z = f(x)$ and $y = g(z)$ does not entail their ability to compute the composition $y = g(f(x))$. We then decode residual stream representations and identify two processing mechanisms: one which solves tasks \textit{compositionally}, computing $f(x)$ along the way to $g(f(x))$, and one which solves them \textit{directly}, without any detectable signature of the intermediate variable $f(x)$. Finally, we find that embedding space geometry is strongly related to which mechanism is employed, where the idiomatic mechanism is dominant when tasks are represented by translations from $x$ to $g(f(x))$ in the embedding spaces.
We fully release our data and code at: \url{https://github.com/apoorvkh/composing-functions}.
\end{abstract}

\section{Introduction}\label{sec:intro}

\textit{Compositional behavior} \citep{mccurdy2024:compositionality} is widely considered essential for flexible and general intelligence \citep{szabo:compositionality}. A long-running debate has asked whether compositional \textit{behavior} necessarily entails compositional \textit{representations} and \textit{processes}. One the one hand, formal languages based on compositional syntax and semantics are guaranteed to support certain types of invariance and generalization, making them compelling models for how humans might achieve abstract cognitive abilities like language and logic \citep{fodor1975:lot,quilty2023:best}.  On the other hand, critics are quick to point out that humans frequently deviate from ideal compositional and logical behavior, suggesting some other mechanism must underlie our advanced cognition \citep{kahneman1972subjective,evans2002logic}.

Large language models (LLMs) provide an opportunity to revisit this debate in a new light. LLMs exhibit behavior that is at least ostensibly compositional, and which is not easily explained away by trivially non-compositional mechanisms \citep{mccoy2023:embers,griffiths2025whither}. However, LLMs also lack the kinds of explicit symbolic architectural components that have long been assumed necessary for such compositionality. This provides an opportunity to ask: do LLMs produce compositional behavior by invoking compositional processes, or do they rely on something more idiomatic instead?

We offer an initial investigation into this question, focusing on a set of two-hop factual retrieval tasks, such as: given a book's title, output that book's author's birth year. All of the tasks we consider can be formally expressed as $y = g(f(x))$ and are thus defensibly "compositional" in the sense invoked in traditional symbolic models. We are interested in whether LLMs solve such tasks by approximating the mapping from $x$ to $y$ \textit{compositionally}, by computing the intermediate variable $z = f(x)$, or \textit{directly}, without a readily-detectable representation of any such $z$. 
We find that:

\begin{enumerate}
    \item {
        Models' ability to compute both $x \to f(x)$ and $f(x) \to g(f(x))$ does not entail their ability to compute $x \to g(f(x))$. This extends earlier findings on the "compositionality gap" \citep{press2022:compositionality-gap}, showing that the gap holds for modern models and on a larger set of tasks. This gap is not trivially reduced in larger models or necessarily by reasoning models (\cref{sec:gap}).
    }
    \item{
        Models exhibit both \textit{compositional} processing mechanisms and \textit{direct} processing mechanisms, as defined above. The type of mechanism is only weakly associated with accuracy, suggesting that LLMs are able to use both effectively to compute correct answers (\cref{sec:lens}).
    }
    \item{
        The choice of mechanism is well predicted by embedding space geometry. Specifically, when there exists a translation (for a given task) from $x$ in the input embedding space to $g(f(x))$ in the output unembedding space, the LLM tends to favor direct computation over compositional processing (\cref{sec:linearity}).
    }
\end{enumerate}

\begin{table*}[b!]
\caption{List of our tasks. The compositional function $(g \circ f)$ is constructed by  $f$ and $g$ here. We list the number of examples (\#) in each task's dataset, along with the variables $x$, $f(x)$, and $g(f(x))$ for one random example. We list $g(x)$ and $f(g(x))$ for tasks that define them in \cref{app:data}.}
\label{tab:tasks}
\begin{center}\begin{adjustbox}{width=\linewidth}
\begin{tabular}{cc|c|c}
\toprule
$f$ & $g$ & \# & $x \to f(x) \to g(f(x))$ \\
\midrule
Word → Antonym & English → Spanish & 2398 & bogus → authentic → auténtico \\
Word → Antonym & English → German & 2398 & philosophical → practical → praktisch \\
Word → Antonym & English → French & 2398 & excessive → insufficient → insuffisant \\
Book → Author & Author → Birth Year & 2228 & The Boy in the Striped Pyjamas → John Boyne → 1971 \\
Song → Artist & Artist → Birth Year & 958 & Heartbreak Hotel → Elvis Presley → 1935 \\
Landmark → Country & Country → Capital & 1385 & Taq-i Kisra → Iraq → Baghdad \\
Park → Country & Country → Capital & 743 & Mount Rainier National Park → United States → Washington, D.C. \\
Movie → Director & Director → Birth Year & 2180 & Cape Fear → Martin Scorsese → 1942 \\
Person → University & University → Year & 4992 & Andi Gutmans → Technion – Israel Institute of Technology → 1924 \\
Person → University & University → Founder & 4996 & Ezra Abbot → Bowdoin College → James Bowdoin \\
Product → Company & Company → CEO & 1904 & NES-101 → Nintendo → Shuntaro Furukawa \\
Product → Company & Company → HQ & 2276 & Toyota Alphard → Toyota → Toyota \\
x + 10 & 2x & 1000 & 699 → 709 → 1418 \\
x + 100 & 2x & 1000 & 922 → 1022 → 2044 \\
x mod 20 & 2x & 1000 & 891 → 11 → 22 \\
Word → Numeric & 2x & 1000 & one hundred and forty-eight → 148 → 296 \\
Word[:-1] & Word[::-1] & 2946 & responsible → responsibl → lbisnopser \\
Rotate(RGB, 120°) & RGB → Name & 1000 & 8a735a → 598a73 → dimgray \\ 
\bottomrule
\end{tabular}
\end{adjustbox}\end{center}
\end{table*}

\section{Task Setup}\label{sec:setup}

Our tasks involve predicting a composition $g(f(x))$ from an input $x$, using in-context learning (ICL) and where $f$ and $g$ are some pre-defined functions. See \cref{tab:tasks} for the full list of tasks we use. We choose common functions $f$ and $g$ that models might learn through their pre-training and for which the inputs and outputs are lexical units. This enables us to use well-established tools for analyzing the mechanisms and latent computations in Transformer models, focusing on a few token positions (i.e. residual streams) and a single autoregressive forward pass. 

We design the set of tasks in our investigation to cover a qualitative variety of functions, such as arithmetic, factual recall, lexical functions, translation, rotation, and string manipulation. By construction, all of our tasks can be computed by applying $f$ and then $g$, yielding the causal hops $x \to f(x) \to g(f(x))$. Some tasks (e.g. commutative tasks) can also be computed through the hops $x \to g(x) \to g(f(x))$ --- in which case, the intermediate $z$ may also equal $g(x)$. We differentiate these further, and also describe our dataset construction methodologies (including our sources and pre-processing), in \cref{app:data}.

In our experiments, we randomly sample 10 in-context examples for a given task and query. Each in-context example is formatted with a "\texttt{Q: \{input\} \textbackslash n A: \{output\} \textbackslash n\textbackslash n}" prompting structure and the test query is formatted with "\texttt{Q: \{input\} \textbackslash n A:}".

\section{Compositionality Gap}\label{sec:gap}

\citet{press2022:compositionality-gap} documented a "compositionality gap" in LLMs, showing that they consistently fail to solve compositions, despite solving the hops independently. \citet{press2022:compositionality-gap} tested the GPT-3 family of models with natural language questions about celebrities and encyclopedic knowledge that required two-hops of factual recall. We confirm and extend this finding by testing modern LLMs on a larger set of compositional tasks.

\subsection{Experimental Design}\label{sec:gap-design}

We prompt models with $\texttt{input} \to \texttt{output}$ mappings between lexical units.\footnote{Note that this represents a methodological difference from \citet{press2022:compositionality-gap}, who prompted with long-form questions. Our format is chosen to fit with the interpretability methods we use in later sections.} We measure models' predictive accuracies using the ICL prompts from \cref{sec:setup}, greedy sampling, and the exact match evaluation metric. We test additional prompt templates in \cref{app:gap-prompts} and the effect of including task-specific instructions in the prompt in \cref{app:gap-instructions}. The \textit{compositionality gap} is defined as the proportion of examples for which a model answers both $x \to f(x)$ and $f(x) \to g(f(x))$ correctly,\footnote{We extend this definition to further require success at $x \to g(x)$ and $g(x) \to g(f(x))$ in tasks where these are valid hops.} but $x \to g(f(x))$ incorrectly.

We test a few Llama 3 and OLMo 2 models on all of our tasks using all available examples. We emphasize results for Llama 3 (3B) in this section and include the other models in \cref{app:gap-models}. We also test an even wider set of models (also including DeepSeek and GPT models) on 4 tasks: \texttt{antonym-spanish}, \texttt{plus-100-times-2}, \texttt{park-country-capital}, and \texttt{book-author-birthyear} (which capture a representative set of processing signatures from \cref{sec:lens}). We aggregate metrics over these tasks and use 100 examples per task for testing.

\begin{figure*}[t]
\begin{center}
\includegraphics[width=0.85\linewidth]{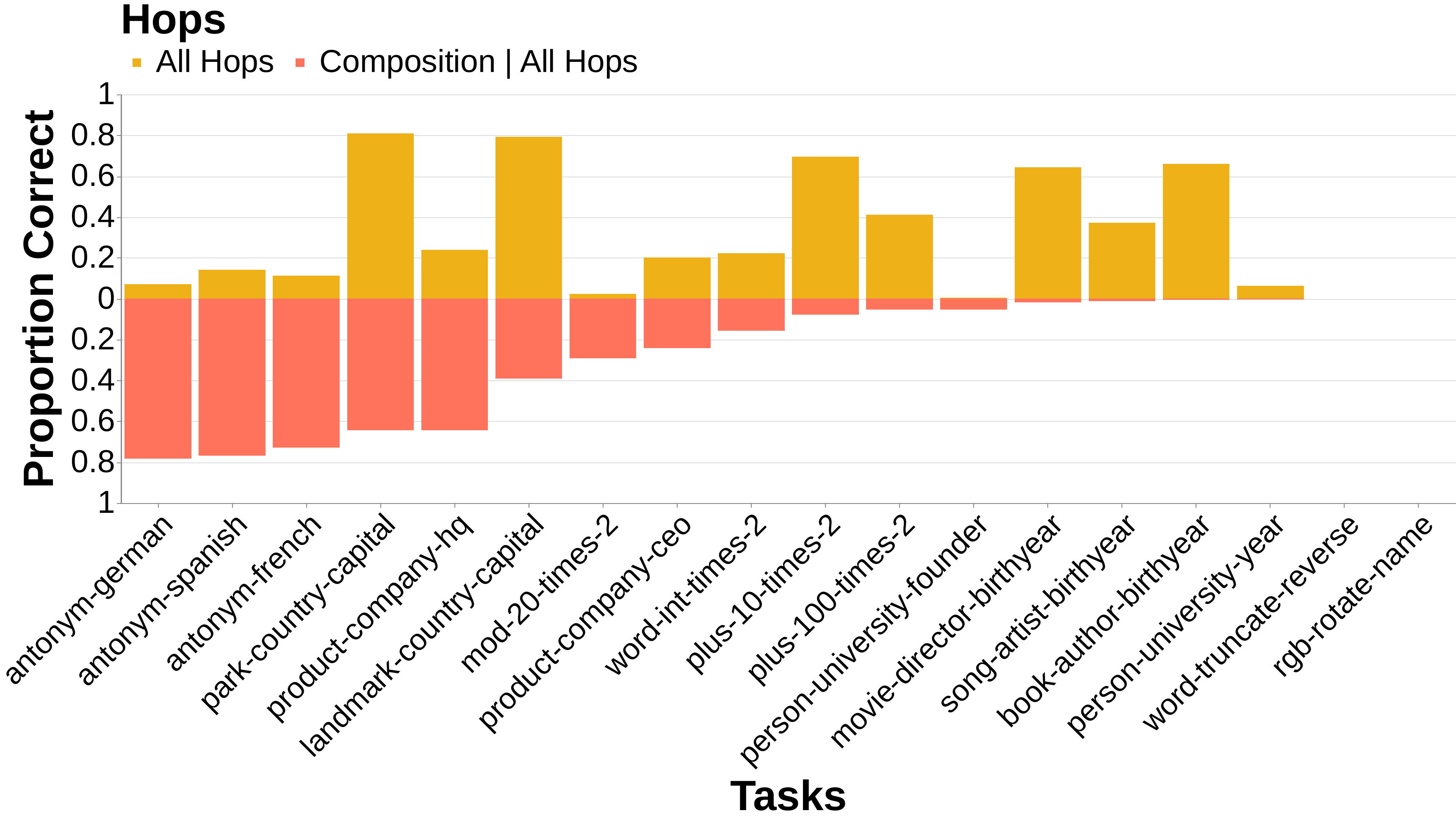}
\end{center}
\caption{Compositionality gap for Llama 3 (3B) on our tasks.
Yellow indicates the proportion of examples for which the model correctly solves all causal hops. Red indicates the (strict) subset of those examples for which the model additionally solves the composition. Correlation between red and yellow bars is $r^2 = \protect\input{artifacts/llama_3_3b/corr/compositionality_gap}$.}
\label{fig:compositionality-gap}
\end{figure*}

\begin{figure*}[t]
\centering
\includegraphics[width=0.85\linewidth]{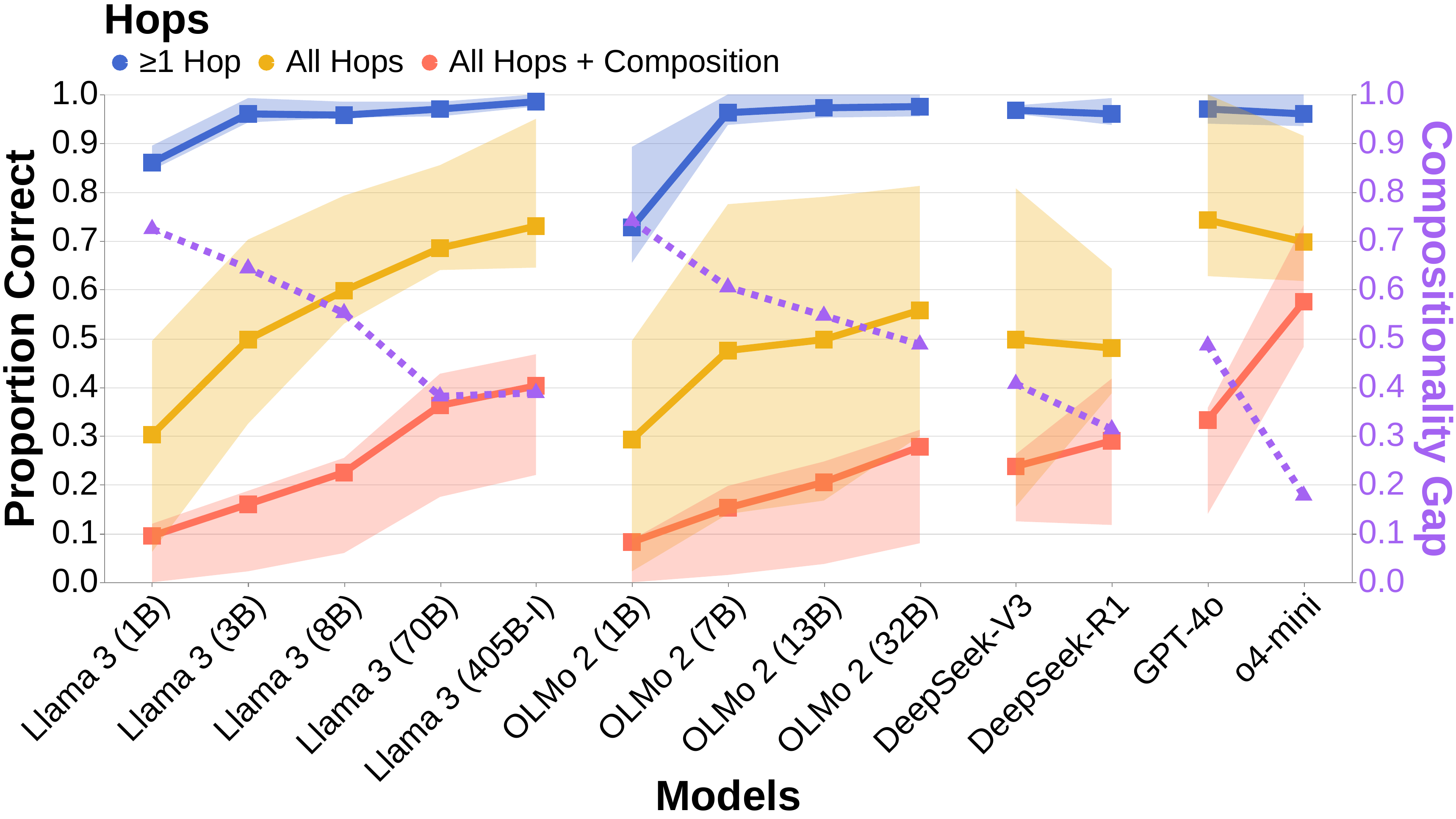}
\caption{Compositionality gap (dashed purple line; lower is better) of various models aggregated over 4 tasks (100 examples each). Blue, yellow, and red lines show proportions of examples for which models correctly solve combinations of hops and the composition. Purple line shows the relative gap between yellow and red: the proportion of examples for which the model cannot solve the composition, out of those for which it can solve all hops. "-I" indicates the instruction-tuned variant of Llama 3 (405B). Error bands show interquartile range.}
\label{fig:compositionality-gap-by-model}
\end{figure*}

\subsection{Results}

We show performance of the Llama 3 (3B) model on our tasks in \cref{fig:compositionality-gap}. We clearly find a compositionality gap: the model is unable to solve the composition in 20--100\% (varying by task) of examples for which it can solve all hops. We show the performances of our other models in \cref{fig:compositionality-gap-by-model}. We find the compositionality gap does reduce with size from 72\% $\to$ 39\% (Llama 3, 1B $\to$ 405B) and 74\% $\to$ 49\% (OLMo 2, 1B $\to$ 32B). However, the gap clearly remains and we find monotonically diminishing improvements for both model families with respect to size. We plot the gap against model parameters and layers in \cref{app:gap-size}. In fact, the gap shows no improvement at all between the 70B and (instruction-tuned) 405B parameter Llama 3 models.

We also compare reasoning models (o4-mini and DeepSeek-R1; allotted a budget of 2000 reasoning tokens) against same-generation, non-reasoning models (GPT-4o and DeepSeek-V3) in \cref{fig:compositionality-gap-by-model}. We find some reduction (41\% $\to$ 31\%) in the compositionality gap in the case of DeepSeek's reasoning model and significant reduction (49\% $\to$ 18\%) in the case of o4-mini. As o4-mini is proprietary (and both this and GPT-4o have additional "external tool-use" capabilities), it is difficult to speculate about the exact causes for these improvements. However, it is notable that even with advanced reasoning models, the gap does not necessarily disappear entirely.

Finally, we include a random sample of correct and incorrect predictions for every model in our supplemental materials.

\begin{figure*}[t]
    \refstepcounter{figure}
    \foreach \x in {lens-correct,lens-incorrect,lens-spanish,lens-movie,lens-10,lens-100} {\refstepcounter{subfigure}\label{fig:\x}}
    \addtocounter{figure}{-1}
    \begin{center}\begin{adjustbox}{width=\linewidth}
    \begin{tabular}{c||cc}
        \includegraphics[width=0.3\linewidth]{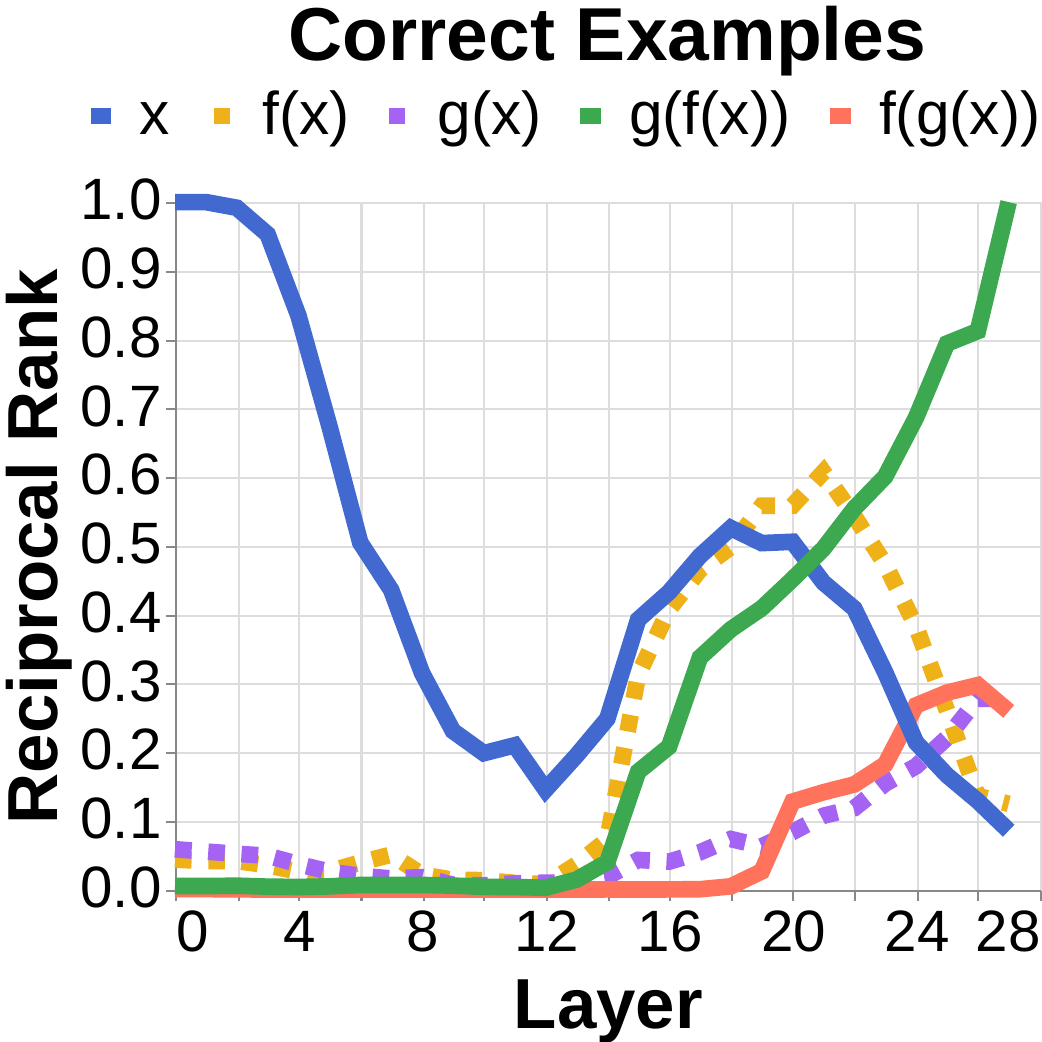} & \includegraphics[width=0.3\linewidth]{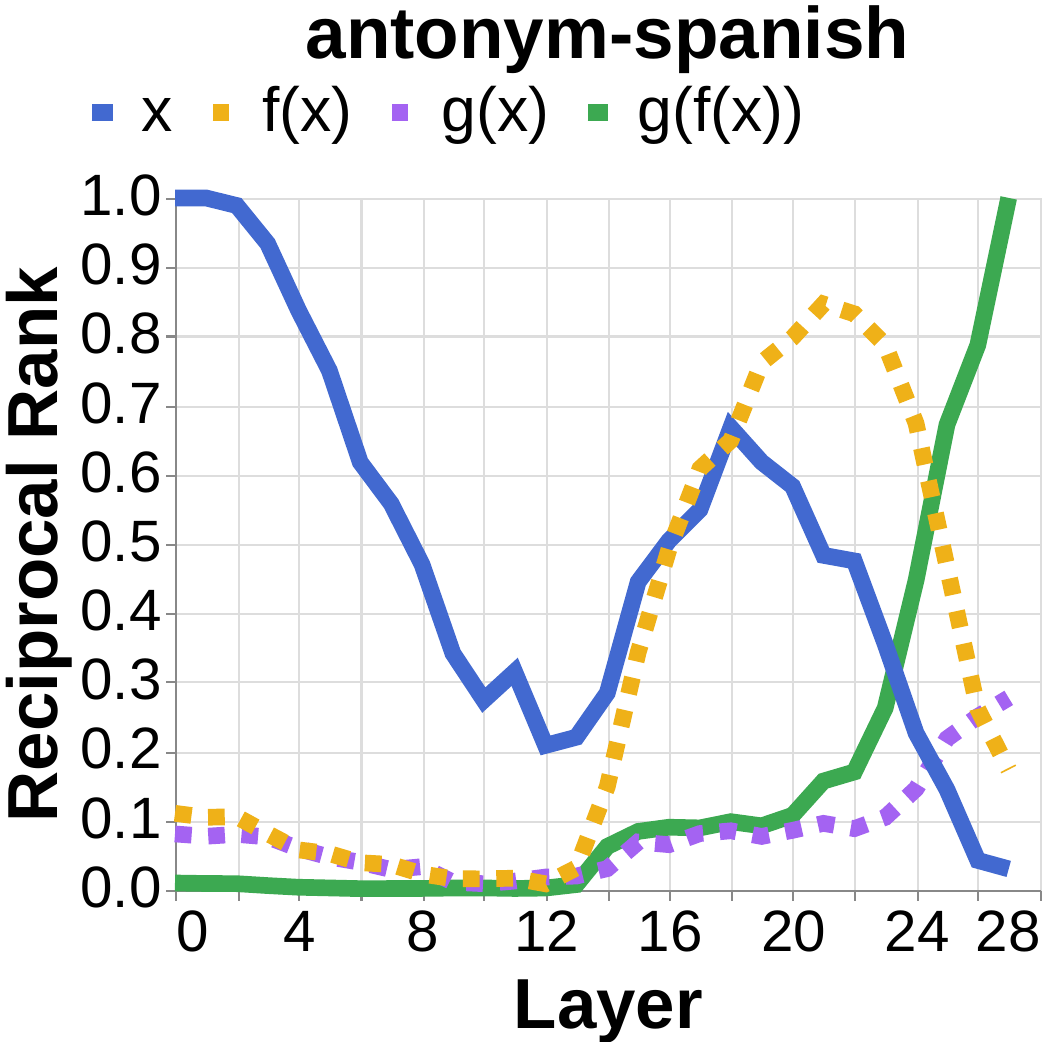} & \includegraphics[width=0.3\linewidth]{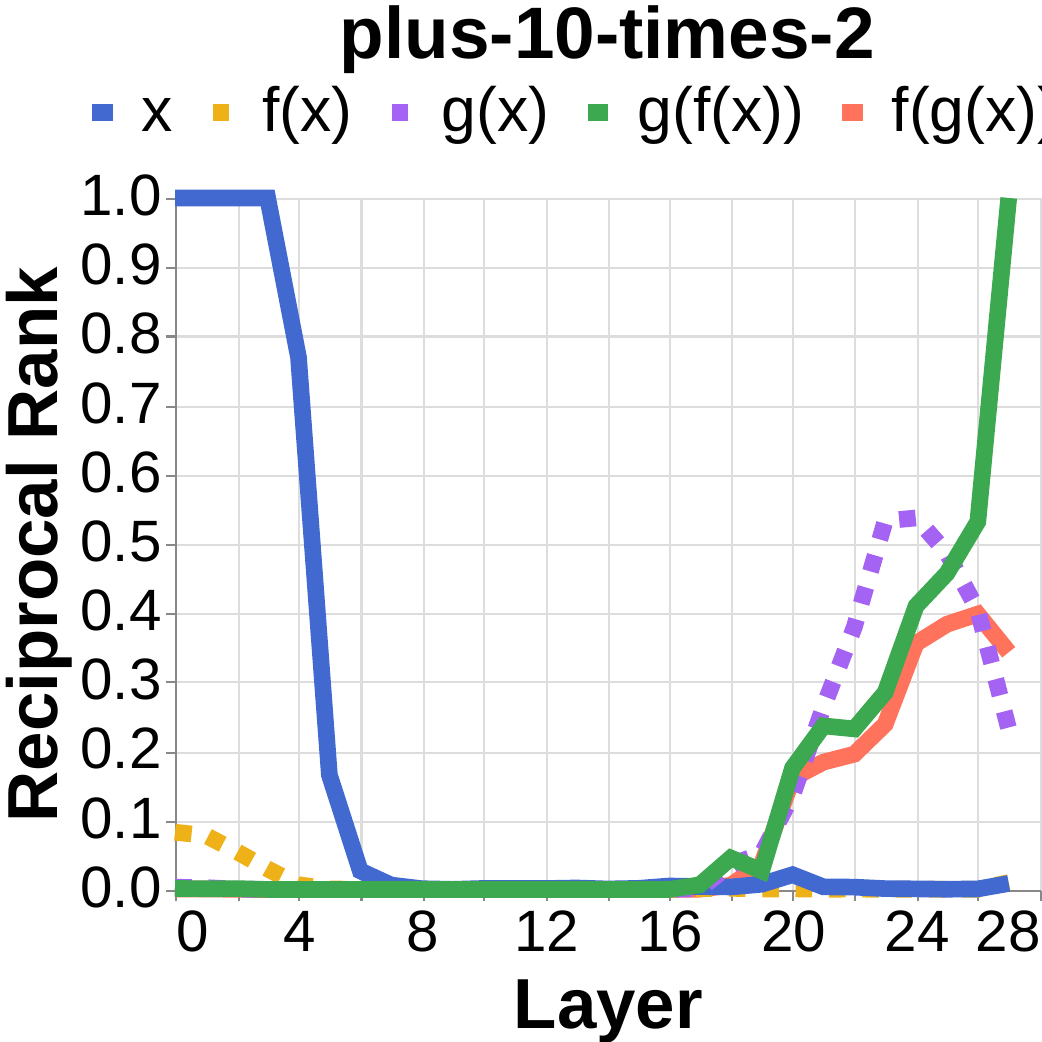} \\
        (a) & (c) & (e) \\
        \includegraphics[width=0.3\linewidth]{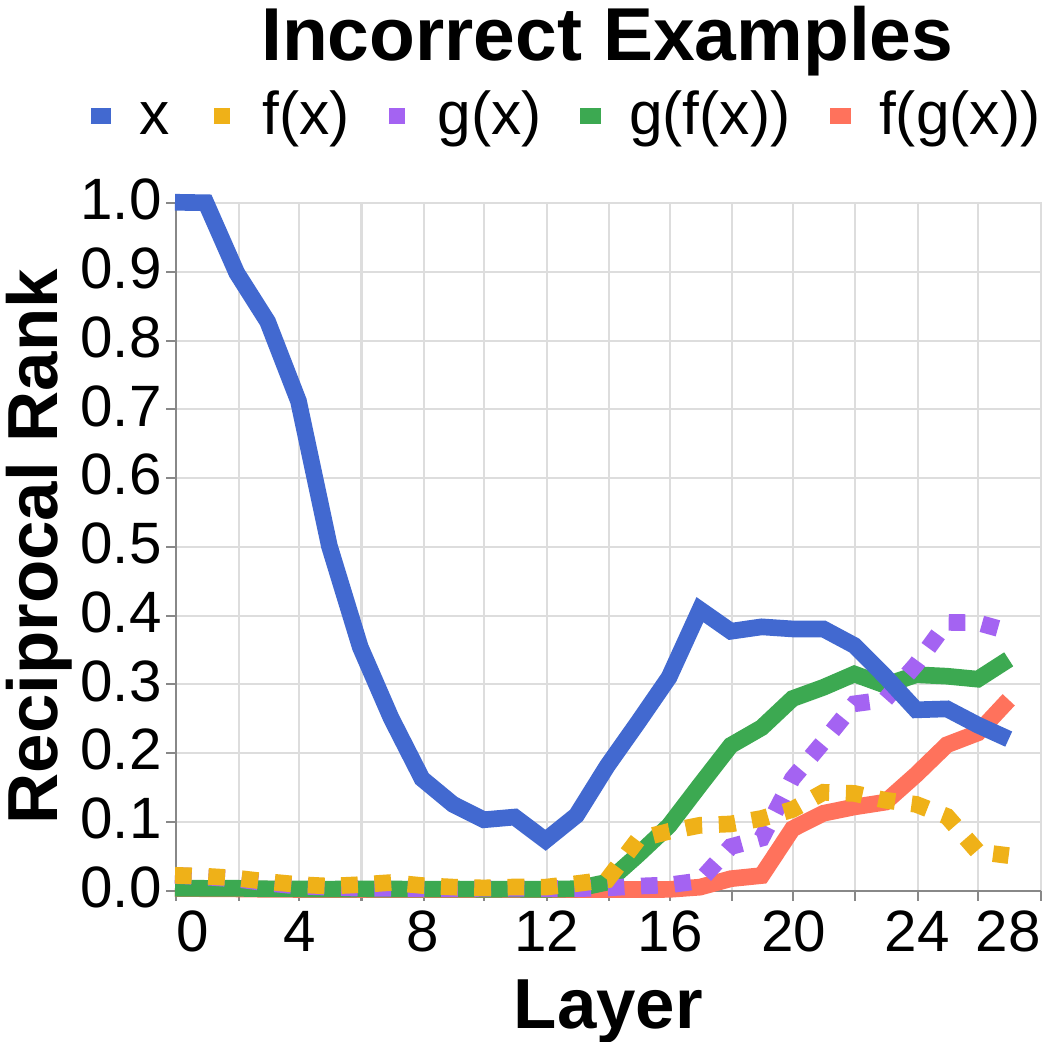} & \includegraphics[width=0.3\linewidth]{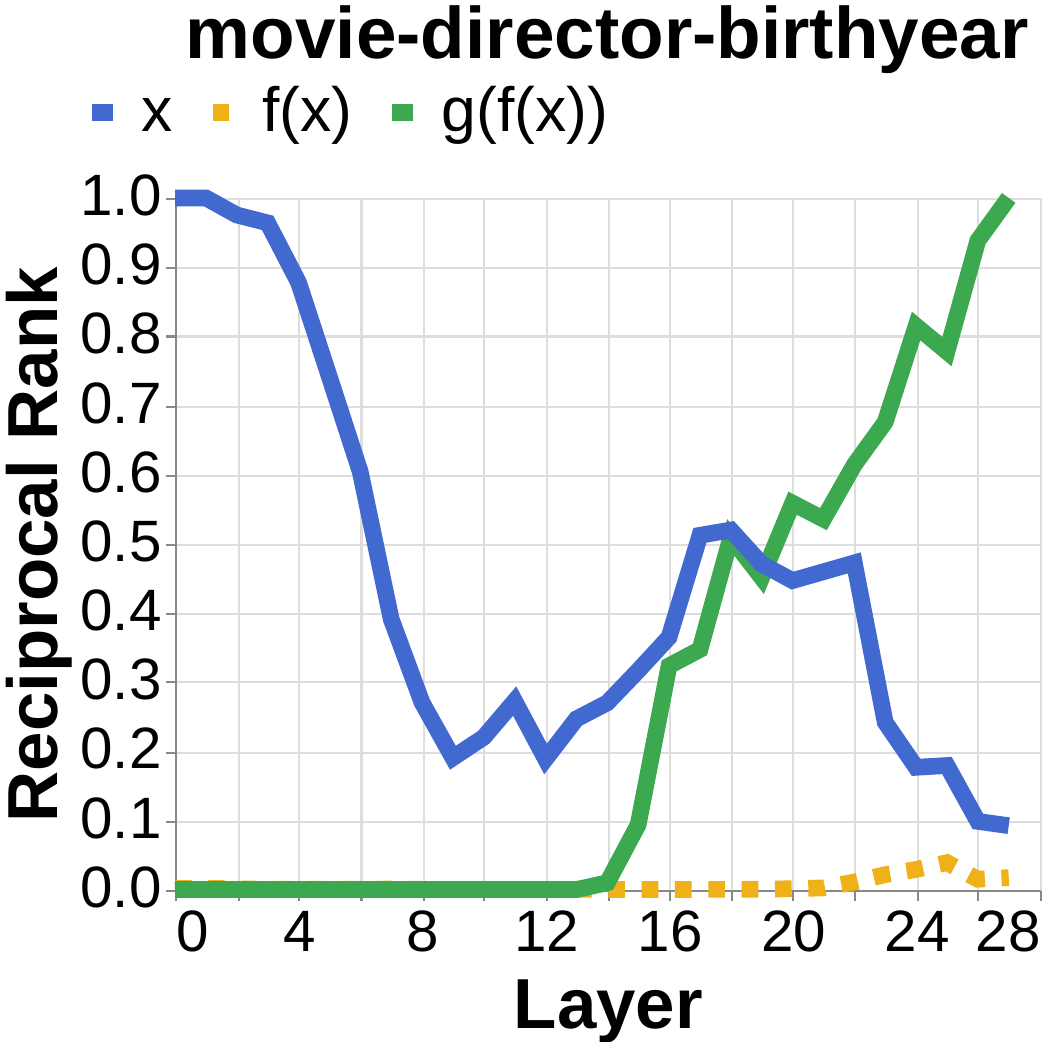} & \includegraphics[width=0.3\linewidth]{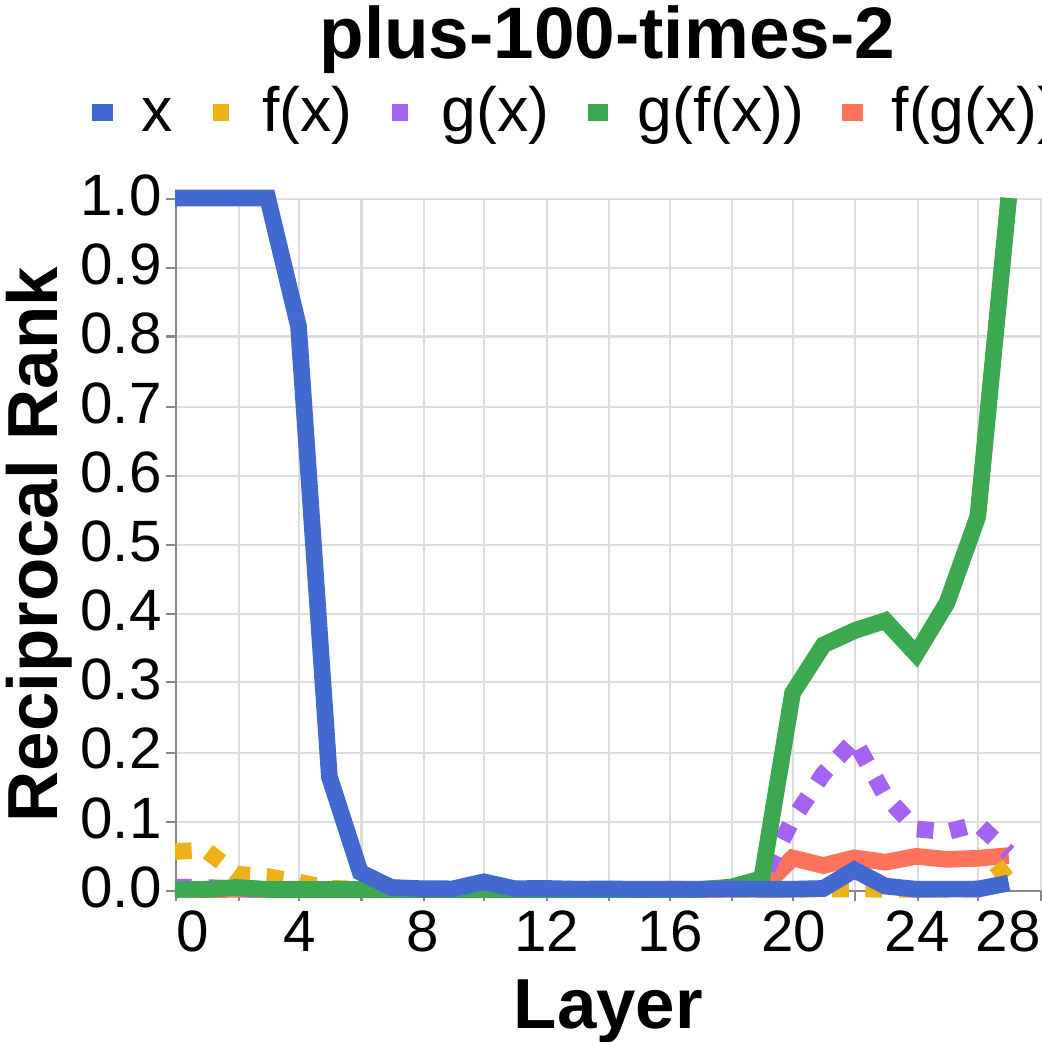} \\
        (b) & (d) & (f) \\
    \end{tabular}
    \end{adjustbox}\end{center}
    \caption{(a--b) Processing signatures aggregated over examples (across all tasks) in which Llama 3 (3B) solves all hops correctly, but the composition (a) correctly or (b) incorrectly. (c--f) Processing signatures for particular tasks --- aggregated over examples where this model correctly solves all hops and the composition. (a--f) Lines show reciprocal ranks of relevant variables (decoded using logit lens) from residual streams corresponding to $x \to g(f(x))$. Intermediate variables are shown with dashed lines. The incorrect composition, $f(g(x))$, is shown by the red line when not distinct from $g(f(x))$.}
    \label{fig:lens-grid}
\end{figure*}

\section{Analyzing Processing Mechanisms}\label{sec:lens}

We next try to understand \textit{how} the model correctly computes compositions in cases where it is successful. Our intuition is based on prior work from \citet{merullo2023:vector-arithmetic} which identifies a processing signature in models that solve one-hop relational tasks. That work shows that models predicting $y = f(x)$ iteratively surface vocabulary representations --- first for $x$ and then for $y$ --- in the residual stream. This "crossover" point was interpreted as evidence of the function $f$ being applied to the argument $x$ in order to yield the final answer $f(x)$ and was localized to specific computations in the MLPs.

In this section, we ask whether an analogous signature will emerge in the case of compositional functions, $g(f(x))$. That is, can we find distinct intermediate representations for $x$, followed by $f(x)$, and then $g(f(x))$ during the model's processing?

Here, we employ analyses most similar to \citet{biran2024:hopping} and \citet{yang2025:internal-cot} in the context of our evaluation (see \cref{sec:related} for further discussion on these works). We join other works in identifying stages of processing within language models \citep{tenney2019:bert,merullo2023:vector-arithmetic,lepori2024:racing,lu2025:paths}.

\subsection{Experimental Design}\label{sec:lens-exp}

We rely on existing methods which allow us to analyze processing signatures that are interpretable using the vocabulary space of the model \citep{nostalgebraist2020:logit-lens,geva2022:promote}. We specifically use \textit{logit lens} \citep{nostalgebraist2020:logit-lens}, a method which projects intermediate representations into the vocabulary space using the language modeling head. Logit lens reveals information that models store in a linear representational subspace (i.e. the unembedding space) that is fundamental to the model. While these signals alone do not guarantee causal use, we find strong signals for our specific variables of interest in this important subspace very compelling and also demonstrate their causal effects in \cref{app:patching}. Such methods are not able to perfectly decompose representations and so we also include results in \cref{app:token-identity} using the token identity patchscope \citep{ghandeharioun2024:patchscopes} to explore an alternative representational subspace. Both methods yield similar findings concerning the mechanisms investigated in this section. Our intervention in \cref{app:patching} further supports these findings, demonstrating existence (or lack thereof) of intermediate variable representations.

We follow the approach from \citet{merullo2023:vector-arithmetic} to identify the processing signature of models that solve our compositional tasks and, in particular, representations of the intermediate variables, $f(x)$ and $g(x)$, prior to those for $g(f(x))$. We specifically use logit lens to analyze the residual streams corresponding to the computation $x \to g(f(x))$ and measure the reciprocal rank of our variables at each layer (see \cref{app:implementation} for more details). We also use the maximum reciprocal rank of our intermediate variables across the layers as a heuristic for their overall presence in the computation.

We focus this section's analysis on Llama 3 (3B) and summarize results for other models in \cref{app:task-lens-models}. We exclusively analyze examples where the model can solve all requisite hops. To ensure sufficient sample sizes, we exclude any task with fewer than 10 such examples where the model can also successfully solve the composition. In particular, the excluded tasks for Llama 3 (3B) are \texttt{song-artist-birthyear}, \texttt{person-university-year}, \texttt{person-university-founder}, \texttt{mod-20-times-2}, \texttt{word-truncate-reverse}, and \texttt{rgb-rotate-name}. We show results for these tasks in \cref{app:lens-correct,app:lens-incorrect}.

\subsection{Results}

\cref{fig:lens-correct} shows the relative presence of each of the variables, across layers and aggregated over all instances in which the model ultimately produced the correct answer. In such cases, we see a very clear peak signal for the intermediate variable $f(x)$, as expected, between those for $x$ and $g(f(x))$. Interestingly, this signal is much less clear for cases in which the model ultimately produces the incorrect answer (\cref{fig:lens-incorrect}). However, upon further inspection, there is little evidence of a causal relationship here, which we discuss further in \cref{app:lens-incorrect}.

There are also plenty of individual examples in which the model produces a correct answer without showing any signature of the intermediate variables, and there is only a weak correlation by task ($r^2 = 0.22$) between predictive accuracy (measured as in \cref{sec:gap-design}) and the presence of intermediate variables as measured by our heuristic (\cref{sec:lens-exp}).

\cref{fig:lens-spanish,fig:lens-movie,fig:lens-10,fig:lens-100} show model processing signatures for a few tasks, aggregated over cases in which the model produces correct answers. We see, for example, that there is a clear signature in the \texttt{antonym-spanish} task (\cref{fig:lens-spanish}) for the intermediate computation of $f(x)$, the word's antonym, before it is translated into Spanish. In contrast, for the \texttt{movie-director-birthyear} task (\cref{fig:lens-movie}), there is no decodable signal for $f(x)$, the movie's director, before the model produces their birth year. This variation can be seen in qualitatively similar tasks as well: tasks with the same basic arithmetic structure (\cref{fig:lens-10,fig:lens-100}) only sometimes carries detectable signatures of $f(x)$ or $g(x)$, depending on the task's operand (e.g. 10 or 100). We show processing signatures for the remaining tasks in \cref{app:lens-correct} and for all tasks, aggregated over unsuccessful cases, in \cref{app:lens-incorrect}.

\begin{figure*}[t]
    \refstepcounter{figure}
    \refstepcounter{subfigure}\label{fig:task-lens-correlation}
    \refstepcounter{subfigure}\label{fig:intermediate-distribution}
    \addtocounter{figure}{-1}
    \centering
    \begin{tabular}{cc}
    \includegraphics[height=0.33\linewidth]{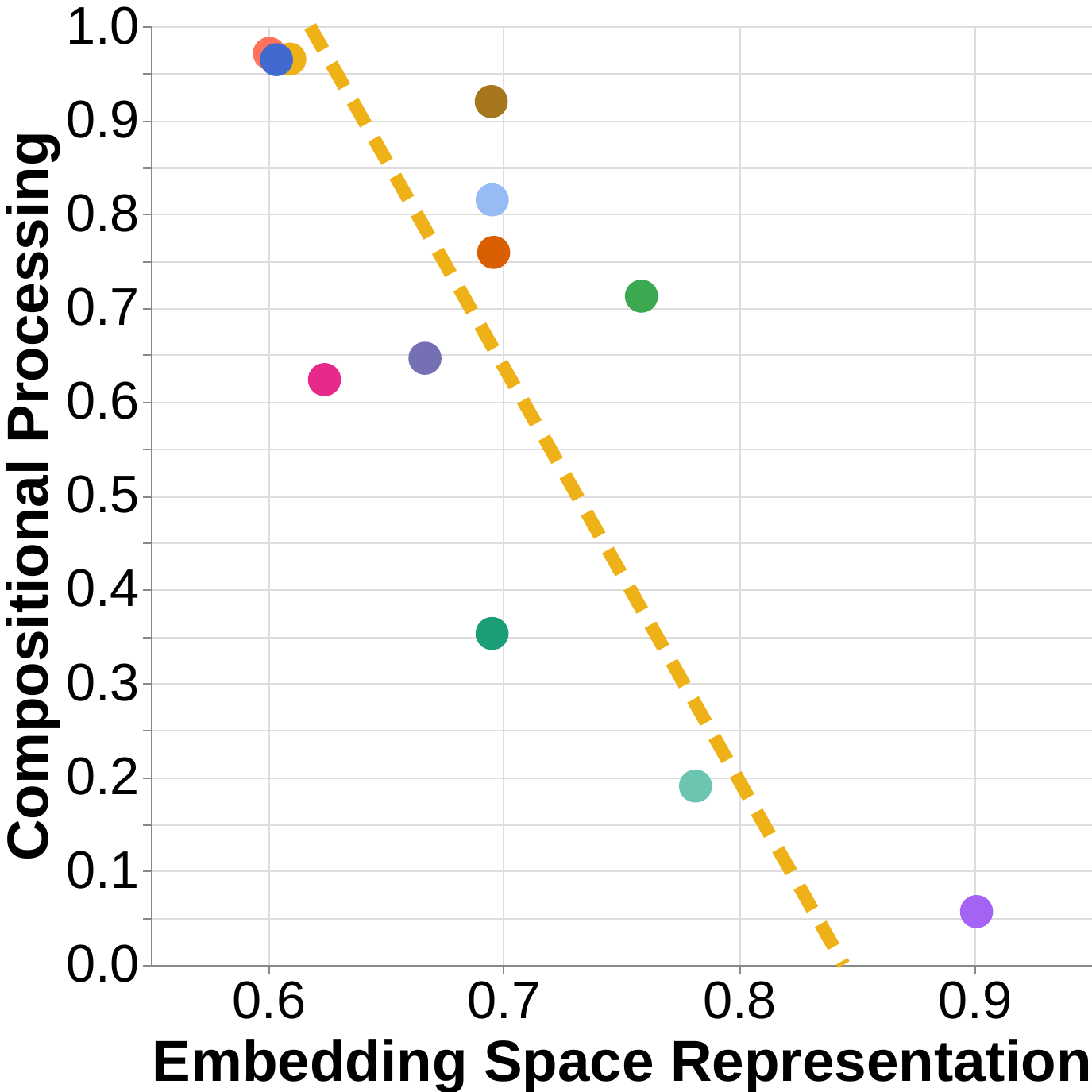} &  \includegraphics[height=0.33\linewidth]{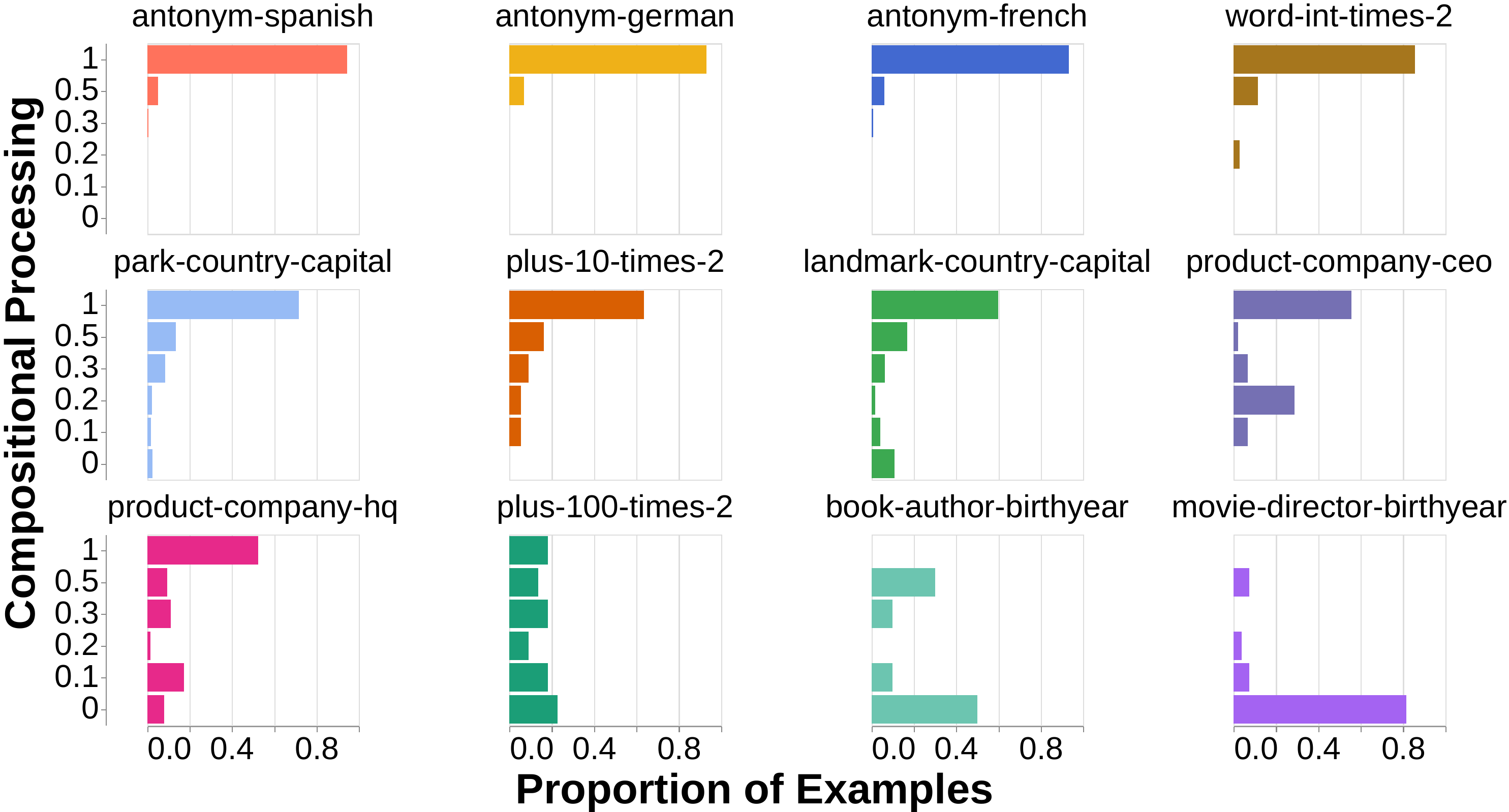} \\
    (a) & (b) \\
    \end{tabular}
    \caption{Relationship between geometry and mechanisms of Llama 3 (3B). (a) Tasks correlate strongly between use of compositional mechanisms (heuristic from \cref{sec:lens-exp} based on reciprocal rank; on average across examples) and whether tasks are represented as translations in the embedding spaces ($r^2 = \protect\input{artifacts/llama_3_3b/corr/lens_task}$). Conversely, accuracy is weakly correlated with these mechanism ($r^2 = \protect\input{artifacts/llama_3_3b/corr/acc_lens}$) and representation ($r^2 = \protect\input{artifacts/llama_3_3b/corr/acc_task}$) metrics. (b) Distribution of examples for each task, shown as a histogram of intermediate variable reciprocal ranks. (a--b) Colors refer to corresponding tasks between points in (a) and histograms in (b).}
\end{figure*}

\section{Compositional Processing and Embedding Space Geometry}\label{sec:linearity}

Given that there is significant variation in whether or not the LLM solves a task compositionally (i.e. how strongly they appear to compute the intermediate variables), we next ask why this variation occurs. It is well-known that some relations are represented by translations in classical word embedding spaces \citep{mikolov2013:word2vec,hewitt2019:structural}. \citet{hernandez2024:linearity} also shows that, when LLMs compute certain \texttt{subject $\to$ object} relations, several layers of computation can be well approximated by simple, linear transformations. We extend these ideas together and posit that language models could directly represent some tasks as translations between the embedding and unembedding spaces (spanning language models' entire forward computation). We hypothesize that some compositional functions could be represented more fundamentally via these translations and that models may compute such functions \textit{directly}, rather than \textit{compositionally} by the hops.

\subsection{Experimental Design}

To investigate our hypothesis, we fit translations for each task using least squares regression from $x$ (average embedding across tokens) to $g(f(x))$ (first token unembedding) on 100 examples.\footnote{See \cref{app:hops} for additional analyses which consider correlations with the representations of the individual hops (rather than the compositional task).} We quantify how well the translation fits the task using its reconstruction accuracy (measured via cosine similarity) on the remaining examples. We quantify how "compositional" the processing is using our heuristic metric which captures the strength of the signal for the intermediate variables, $f(x)$ and $g(x)$ (see \cref{sec:lens}). We again restrict our analysis to examples where the model is successful on all hops and the composition, as well as tasks with at least 10 such examples.

\subsection{Results}

\cref{fig:task-lens-correlation} shows the that there is a a strong inverse correlation ($r^2 = 0.65$) between embedding space task geometry and compositional processing for Llama 3 (3B). That is, the more a task is represented as a translation in the embedding spaces, the more likely the model is to display \textit{idiomatic} (as opposed to \textit{compositional}) processing.

This correlation is computed by averaging the indicator for compositional processing across instances for each task. \cref{fig:intermediate-distribution} shows the de-aggregated distribution of our "compositionality" metric across the examples. For some tasks, it appears that nearly all individual examples behave the same way. For example, nearly every instance of \texttt{antonym-spanish} displays a compositional signature, while almost every instance of \texttt{movie-director-birthyear} displays an idiomatic one. On the other hand, this distribution is more uniform for other tasks, such as \texttt{plus-100-times-2}. This distribution appears to be bimodal across all examples: 82\% have very low ($< 0.1$) or high ($\geq 0.5$) values for compositionality.

We include results for several other models in \cref{app:task-lens-models}, i.e. Llama 3 (3B) Instruct, Llama 3 (8B), OLMo (7B), and OLMo 2 (13B). All the models models we test continue to show strong correlations for the relationship presented here, with an average $r^2 = 0.66$. These models continue to show bimodal distributions of the choice of mechanism across examples. We also explore the hypothesis in \cref{app:linear} that linear representations could be predictive of the mechanism, but find a weaker fit than with the translation-based representations shown here.

\section{Discussion}\label{sec:discussion}

\paragraph{Summary of Findings} Our results suggest that tasks which appear to have the same computational structure may nonetheless be processed differently by LLMs. In particular, we consider functions which appear compositional in a formal sense --- i.e. they can be represented as $y = g(f(x))$ for some reasonably defined $f$ and $g$. We find evidence that LLMs only sometimes process such functions compositionally, showing evidence of representing or computing the value of $z = f(x)$ on the way to computing $y$. In other cases, LLMs appear to map $x$ to $y$ directly. Which of these processes is invoked appears to be related to how well the relationship between $x$ and $y$ is represented as a translation in the embedding space, e.g. as a result of training conditions.

\paragraph{Implications for Theories of Compositionality} There is a long-running debate about the degree to which compositional behavior \citep{mccurdy2024:compositionality} requires compositional mechanisms. The two sides of this debate have often talked past each other, using different types of computational architectures in order to model different aspects of behavior, for example, explicitly compositional symbolic systems for formal domains \citep{lake2017:building,ellis2023dreamcoder} vs. distributional or neural systems for humans' more idiomatic performance \citep{erk2012vector,lampinen2024:content-effects}.

Attempts to find compromises or "hybrid" systems often consist of neuro-symbolic systems which are designed top-down \citep{andreas2016neural,ellis2018learning}. Large language models offer an alternative approach for advancing this debate. LLMs have proven capable of a range of behaviors that have traditionally required compositionality --- e.g. generating language and writing formal computer code. However, LLMs lack the explicit symbolic mechanisms traditionally associated with such behaviors. Using methods from interpretability to understand how LLMs represent such functions internally enables us to approach the question in a "bottom up" manner, potentially offering novel hypotheses about the mechanisms that can generate behavior that is sometimes systematic and other times heuristic, as is the case in humans \citep{russin2025parallel}.

Our results suggest that LLMs employ a mix of compositional and idiomatic processing, and that the choice of mechanism is related to the representations of the functions that result from pretraining. This offers an interesting perspective on one question that is frequently at the heart of discussions of compositionality --- i.e. what are the primitives and where do they come from \citep{carey2011}? The relationship between embedding space task geometry and compositionality of processing presented here suggests an attractive hypothesis that the primitives are those things which are well represented as a result of (pre-)training, and that compositional mechanisms are invoked to handle those things which are not sufficiently well represented. Future work in this direction would likely yield interesting new results and topics for debate.

\paragraph{Relationship to work on compositional generalization}

The work presented here concerns the (apparent) compositionality of the processing mechanism, but does not directly relate this mechanism to an LLM's capacity for compositional generalization. The majority of work on compositionality in neural networks (and LLMs) concerns compositional generalization, and the compositionality researchers surveyed by \citet{mccurdy2024:compositionality} overwhelmingly agree that existing language models are insufficient in this regard. This belief is supported by evidence from many prior works (\cref{sec:related}) and our investigation in \cref{sec:gap}.

Our work suggests that models employ both compositional and direct mechanisms to solve tasks. Intuitively, we would expect there to be a relationship between the use of the mechanism and the ability to generalize --- i.e. the compositional mechanism should support generalization better than the idiomatic mechanism ("memorization"). However, we do not test this intuition directly in this paper. Future work could do so by employing causal interventions on the intermediate variables. This would likely present new complexities and challenges that would enrich our understanding of compositionality, and of the relationships between mechanisms and behaviors in LLMs in general.

\paragraph{Practical Implications}

Our work identifies which functions LLMs treat as "primitive" based on their geometry. Models allocate representational capacity to memorizing these primitives. If mechanisms are required for generalization, one may prefer models that fit truer primitives (e.g., $f$, $g$, $h$ rather than $g \circ f$) and invoke the compositional mechanism, enabling unseen compositions such as $f \circ h$. Our analysis also provides processing signatures that reveal which functions a model attempts to invoke (successfully or not, e.g. \cref{app:lens-incorrect}), and when individual data points are poorly fit by a function’s embedding space representation, possibly indicating weak memorization during training.

\section{Related Work}\label{sec:related}

\paragraph{Latent multi-hop reasoning} Our work is most closely related to recent works which also study latent two-hop reasoning in large language models.
\citet{yang2024:latent} use causal interventions to identify the existence of the hops in the latent computation and whether they co-occur. \citet{biran2024:hopping} employ the entity description patchscope \citep{ghandeharioun2024:patchscopes} to inspect intermediate representations and localize the hops, finding they are resolved in different layers and token positions. They propose a representational intervention ("backpatching") to correct failures based on this finding. Finally, \citet{yang2025:internal-cot} use both causal interventions and logit lens to analyze intermediate representations and consistently find a "compositional" processing signature across their tasks. Our work employs all of these interpretability methods (\cref{sec:lens,app:token-identity,app:patching}) to analyze the hops, but specifically highlights and investigates the duality of the compositional vs. direct processing mechanisms. All works (including our own) test different sets of tasks, make experimental design decisions according to their independent goals,%
\footnote{One notable example is that, while \citet{yang2024:latent} and \citet{biran2024:hopping} prompt their models with $f$ and $g$ (e.g. "\texttt{The mother of the singer of \{x\} is \{y\}}"), we typically omit this information from our prompts (i.e. more simply "\texttt{Q: \{x\} \textbackslash n A: \{y\}}"), besides in our ablation, to avoid inducing bias towards the compositional mechanism.}%
 and make findings in context of their own experiments.

Among other works in this domain, \citet{wang2024:grokked} trains a language model on synthetic compositional data and identifies a multi-hop reasoning circuit in this model. \citet{shalev2024:distributional} conduct a distributional analysis (considering semantic category spaces, rather than individual tokens) using logit lens. \citet{li2024:understanding,yu2025:backattn} also propose interventions on intermediate representations and mechanisms to solve failure cases. \citet{yang2024:latent-exploiting} conduct an evaluation that is intentionally designed to omit opportunities for models to exploit shortcuts.

\paragraph{Compositionality} Compositionality is long-studied \citep{fodor1988:connectionism,partee2004} but exact definitions evade general consensus. \citet{russin2024:frege} and \citet{mccurdy2024:compositionality} offer recent overviews on the topic in the context of large language models. \citet{russin2024:frege} provide a historic account of compositionality and review studies of compositionality generalization in neural networks. \citet{mccurdy2024:compositionality} survey compositionality researchers on how to define and evaluate compositional behavior in neural networks. These researchers agree that our current representational analyses are insufficient for evaluating models, but are divided about whether our behavioral analyses are sufficient.

In a partial effort towards defining compositionality, \citet{hupkes2020:compositionality} identify five particular aspects of compositionality and propose tests for each using a synthetic, fully compositional translation task. Systematicity is one such aspect and is prominently studied: see \citet{vegner2025:systematicity} for a survey of benchmarks for systematic generalization. Our work --- in which we test whether $f(x)$ is evaluated before $g(f(x))$ --- is closest to \citet{hupkes2020:compositionality}'s aspect of localism, in which "smaller constituents are evaluated before larger constituents".

Among many other works, \citet{johnson2017:clevr}, \citep{keysers2019:cfq}, \citet{lake2018:scan}, \citet{hupkes2020:compositionality}, and \citet{kim2020:cogs} offer prominent benchmarks that behaviorally test for compositional generalization in neural networks trained from scratch on compositional data. These works generally show that such models perform poorly on generalization, or at least poorly implement the compositional processes that underlie the data. \citet{press2022:compositionality-gap} and \citet{ma2023crepe} continue to show significant failures in compositional behavior and generalization in pre-trained models. On the other hand, \citet{furrer2020} points out that pre-training a masked language model rivals or outperforms architectures specifically designed for the SCAN \citep{lake2018:scan} and CFQ \citep{keysers2019:cfq} generalization benchmarks. \citet{lepori2023:break} finds that neural networks learn to implement compositionality structurally in their weights, supporting this claim against the need for specialized symbolic mechanisms.

\paragraph{Compositionality of functions} Several works consider how LLMs solve compositions of functions (rather than specifically multi-hop reasoning tasks). \citet{dziri2023:faith} studies how LLMs autoregressively solve such tasks, like multi-digit multiplication, by inspecting their scratchpads. \citet{wattenberg2024:relational} propose mechanisms which neural networks could use to implement relational compositions. \citet{yu2023:zs-fn-comp,todd2024:fv} propose zero-shot methods to invoke compositions of functions in LLMs that have learned the primitive functions. \citet{zhou2024:comp-functions} find that language models can compose functions with meta-learning in a way that imitates human behavior.

\section*{Reproducibility}

We make our data and code fully available so that all of our experiments can be replicated as closely as possible and all computational artifacts (datasets, plots, results) can be reconstructed. We do our best to include all experimental details in the main text and appendices of our paper.

\section*{Limitations}

In this work, we primarily analyze the computation that occurs in a single forward pass of LLMs. It is also necessary to understand how other models (e.g. reasoning models or different inductive biases) implement compositional functions. Our findings reflect our specific tasks and experimental design. Further work should test other kinds of compositional functions, and try to more deeply understand the relationship between compositional mechanisms, behavior, and generalization. We investigate a limited subset of mechanisms in language models and use current methods to conduct our analyses. These permit us to decode some, but not all, relevant representational structure. Some signals that we do decode may be a result of feature multiplicity or are not guaranteed to be causal. Finally, some of our tasks (e.g. arithmetic) may be solved by algorithms that we do not consider.

\section*{Acknowledgments}
We are very grateful to the members of the Language Understanding and Representation (LUNAR) Lab at Brown University --- especially Jacob Russin and Michael A. Lepori --- and to Chen Sun, Najoung Kim, and Sohee Yang for their generous feedback. This research was conducted using computational resources and services at the Center for Computation and Visualization, Brown University. This work was supported in part by a Brown University Presidential Fellowship for Apoorv Khandelwal and in part by a Young Faculty Award from the Defense Advanced Research Projects Agency Grant \#D24AP00261. Ellie Pavlick is a paid consultant for Google DeepMind. The content of this article does not necessarily reflect the views of the US Government or of Google, and no official endorsement of this work should be inferred.

\bibliography{_b_references}
\bibliographystyle{icml2026}
\newpage
\appendix
\onecolumn
\raggedbottom

\section{Data Creation}\label{app:data}

\begin{table}[H]
\caption{List of our tasks, showing $x$, $g(x)$, and $f(g(x))$ for the random example in \cref{tab:tasks}. Tasks with neither $g(x)$ nor $f(g(x))$ are omitted. $f(g(x))$ only shown if distinct from $g(f(x))$.}
\label{tab:tasks_gx_fgx}
\begin{center}\begin{adjustbox}{width=\linewidth}
\begin{tabular}{cc|c|c|c}
\toprule
$f$ & $g$ & $x$ & $g(x)$ & $f(g(x))$ \\
\midrule
Word → Antonym & English → Spanish & bogus & false & --- \\
Word → Antonym & English → German & philosophical & philosophisch & --- \\
Word → Antonym & English → French & excessive & excessive & --- \\
x + 10 & 2x & 699 & 1398 & 1408 \\
x + 100 & 2x & 922 & 1844 & 1944 \\
x mod 20 & 2x & 891 & 1782 & 2 \\
Word → Numeric & 2x & one hundred and forty-eight & two hundred and ninety-six & --- \\
Word[:-1] & Word[::-1] & responsible & elbisnopser & elbisnopse \\
Rotate(RGB, 120°) & RGB → Name & 8a735a & dimgray & --- \\ 
\bottomrule
\end{tabular}
\end{adjustbox}\end{center}
\end{table}

All tasks in \cref{tab:tasks_gx_fgx} permit the additional computational pathway $x \to g(x) \to g(f(x))$. Those which don't list $f(g(x))$ are commutative and so $f(g(x)) = g(f(x))$ and applying $f$ to $g(x)$ results in $g(f(x))$. The remaining tasks are not commutative, but their formal construction permits the hop $g(x) \to g(f(x))$ anyway. In particular, $g(f(x))$ equals $g(x) + 20$ in \texttt{plus-10-times-2}, $g(x) + 200$ in \texttt{plus-100-times-2}, $g(x) \bmod 40$ in \texttt{mod-20-times-2}, and $g(x)$[1:] in \texttt{word-truncate-reverse}.

\subsection{Task Construction}

\paragraph{Antonyms \& Translations} We obtain a list of antonyms from \citet{todd2024:fv} --- further derived from \citet{nguyen2017:antonyms} --- and obtain translations from Opus-MT \citep{tiedemann-2020:opus-mt}.

\paragraph{Factual Relations} We obtain various factual relations from WikiData and IMDb Non-Commercial Datasets \citep{vrandevcic2014:wikidata,imdb,bast2017:qlever}. We apply a number of heuristics to obtain well-known and unambiguous mappings. For example, we filter entities by their "sitelinks" on WikiData or "votes" on IMDB (heuristics for popularity) to obtain well-known subjects. To avoid ambiguity, we identify subjects (songs, books, movies, people, etc.) with a single corresponding object (authors, attended universities, etc.). We omit parks and landmarks that exist in their country's capital. Our exact queries for generating each task can be found in our source code.

\paragraph{Arithmetic} We use the range of numbers from 0 to 999 as $x$ in our tasks. These numbers typically result in one token. We use the \href{https://github.com/savoirfairelinux/num2words}{\texttt{num2words}} library to obtain a mapping between words and numeric values. We use the list of antonyms as our list of words for the \texttt{word-truncate-reverse} task.

\paragraph{\texttt{Colors}} In the \texttt{rgb-rotate-name} task, we randomly sample RGB colors, rotate them $120\degree$ by their hue, and map the resulting color value to that color's name (using the \href{https://webcolors.readthedocs.io}{\texttt{webcolors}} library and the common CSS 3 specification).

\section{Implementation Details}\label{app:implementation}

\paragraph{Examples \& Prompts} We prevent sampling of in-context examples that intersect in the variables $\{x, f(x), g(x), g(f(x)), f(g(x))\}$ with the query. And, as mentioned in \cref{sec:lens-exp}, we exclude examples in \cref{sec:lens,sec:linearity} which overlap in the first token among their variables. So, although $x = \textrm{"excessive"}$ for the \texttt{antonym-french} task is listed in \cref{tab:tasks_gx_fgx}, this trivially shares the same first token as $g(x) = \textrm{"excessive"}$ and would be omitted from our analyses.

Our prompts are tokenized differently when predicting numbers or words, e.g. "\texttt{... \textbackslash n A: 99}" results in \texttt{[ ][99]} whereas "\texttt{... \textbackslash n A: modern}" results in \texttt{[ modern]}. We accordingly include the trailing space in our prompts when predicting numbers and omit it otherwise. We would then test for the single-token prediction of [99] and [ modern] in this example.

\paragraph{Representational analysis} In \cref{sec:lens}, we analyze the model's computation from $x \to g(f(x))$. Consider the query for "Heartbreak Hotel" $\to$ "1935": i.e. "\texttt{... Q: Heartbreak Hotel \textbackslash n A: }". Here, multiple tokens (\texttt{[ Heart][break][ Hotel][ \textbackslash][n][ A:][ ]}) are central to the computation. We therefore analyze all residual streams for these tokens. At each layer, we measure the signal for each variable by its maximum reciprocal rank across the streams. This procedure yields processing signatures, which quantify the presence of our variables at every layer.

We additionally represent each variable by its first token (since our decoding methods can only produce single-token probabilities). We exclude examples where different variables share the same first token and would be hard to differentiate. For example, $f(x) = \textrm{" modern"}$ and $g(f(x)) = \textrm{" moderno"}$ both share the first token \texttt{[ modern]}.

\section{Compositionality Gap by Prompt Template}\label{app:gap-prompts}

\begin{figure}[H]
    \begin{center}
        \includegraphics[width=\linewidth]{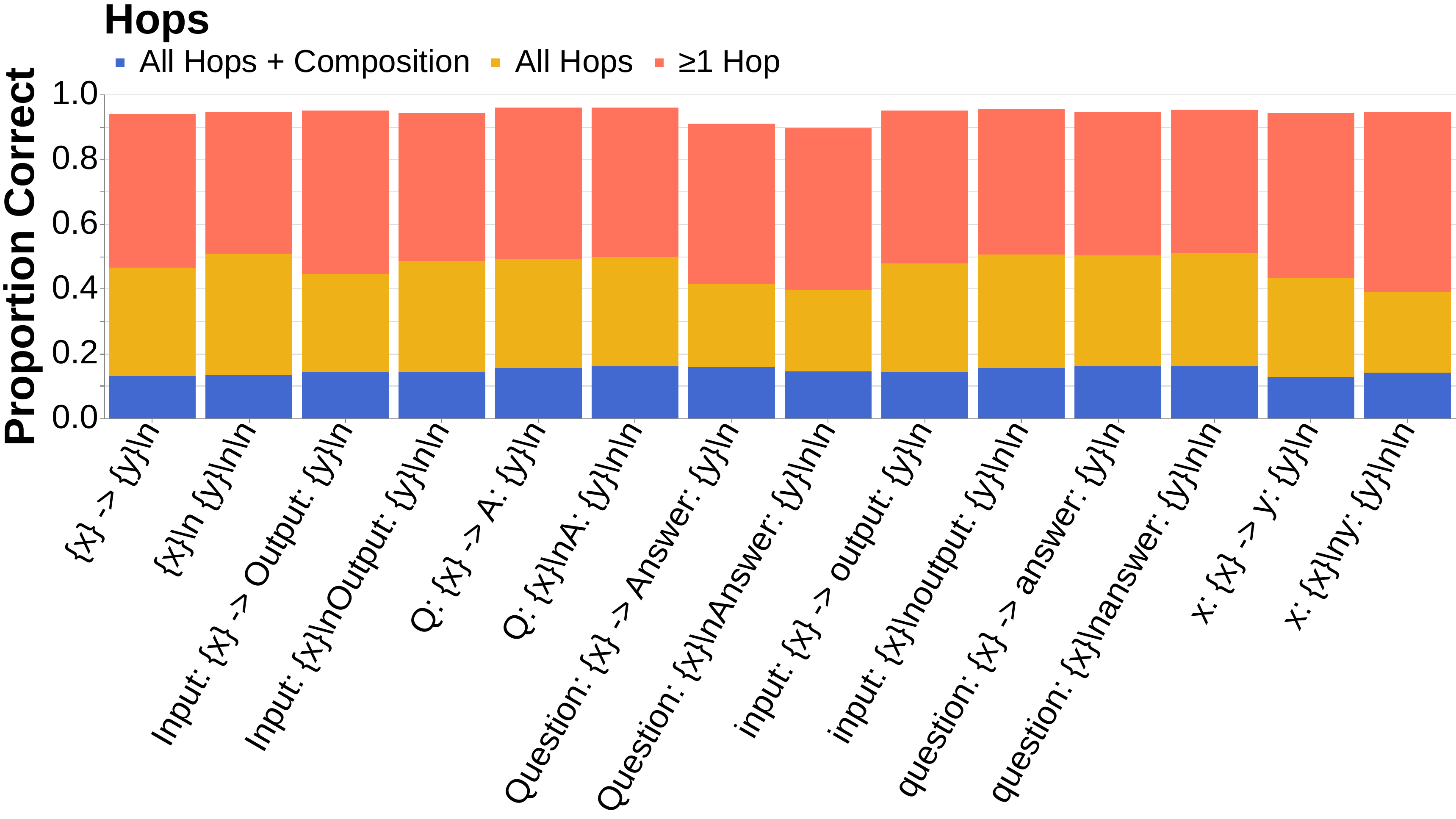}
        \end{center}
    \caption{Ablation of Llama 3 (3B) model with different ICL prompt templates under the settings in \cref{fig:compositionality-gap-by-model}. We find no significant difference in performance by choice of prompt.}
\end{figure}

\section{Compositionality Gap with Task Instructions}\label{app:gap-instructions}

\begin{figure}[H]
    \begin{center}
        \includegraphics[width=0.9\linewidth]{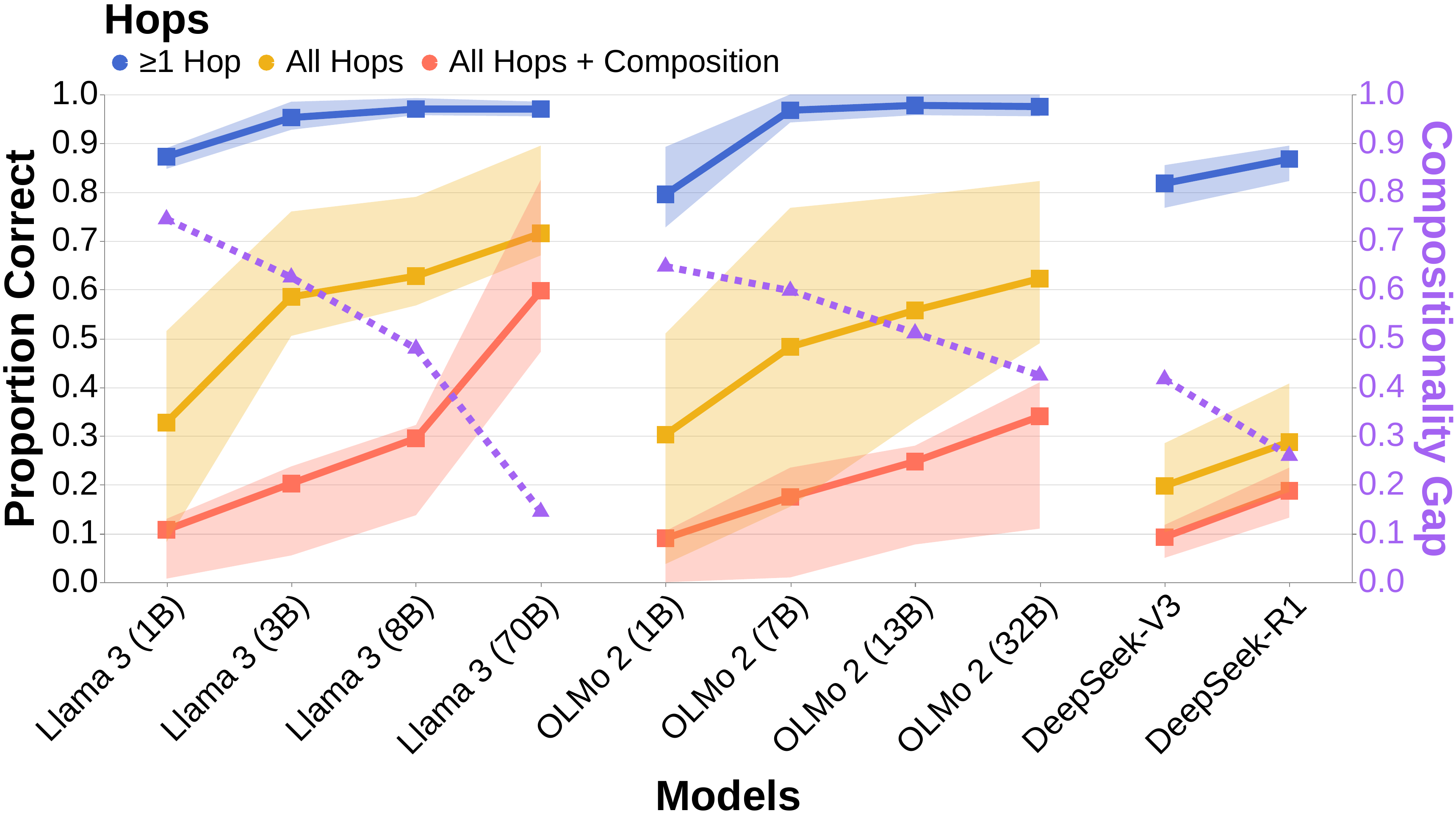}
        \end{center}
    \caption{We repeat our experiments in \cref{fig:compositionality-gap-by-model}, but always prefix the existing ICL prompt with a task instruction: e.g. "Add 100." for $x \to f(x)$ or "Add 100, multiply by 2." for $x \to g(f(x))$. We continue to find a clear compositionality gap in our results. We observe the gap diminishes more steeply w.r.t. model size or reasoning than in the original experiments. Llama 3 (70B) is also substantially closer to closing the gap. We omit the instruction-tuned Llama model and OpenAI models, whose outputs are harder to constrain (and thus evaluate) given the additional instruction prompt.}
\end{figure}

\section{Compositionality Gap by Size}\label{app:gap-size}

\begin{figure}[H]
    \begin{center}
        \includegraphics[width=0.4\linewidth]{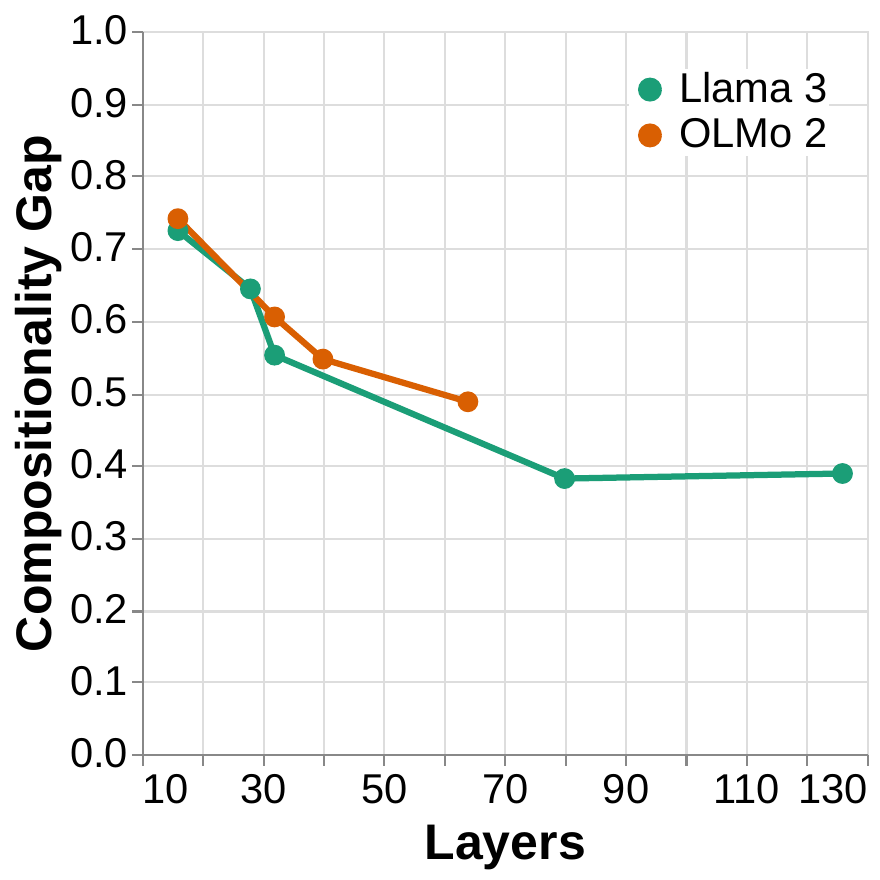}
        \includegraphics[width=0.4\linewidth]{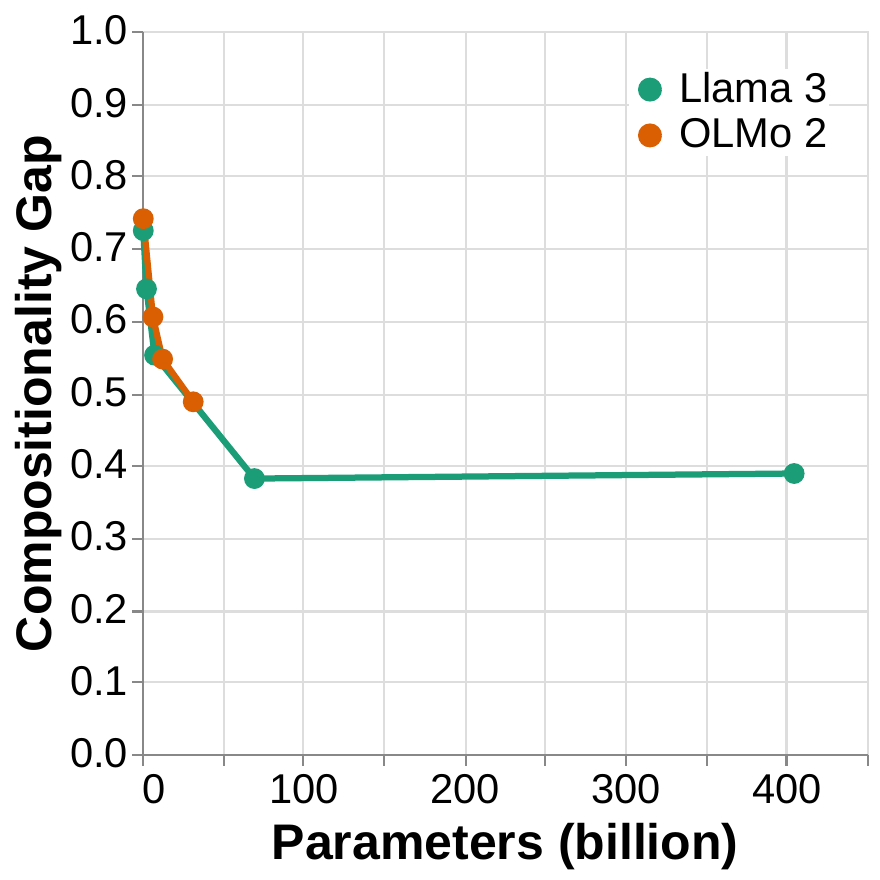}
        \end{center}
    \caption{We illustrate the monotonically diminishing improvements to the compositionality gap resulting from increased model size (layers and parameters). We re-visualize results for the OLMo 2 and Llama 3 model families from \cref{fig:compositionality-gap-by-model}.}
\end{figure}

\section{Compositionality Gap on Additional Models}\label{app:gap-models}

\begin{figure}[H]
\begin{center}
\includegraphics[width=0.9\linewidth]{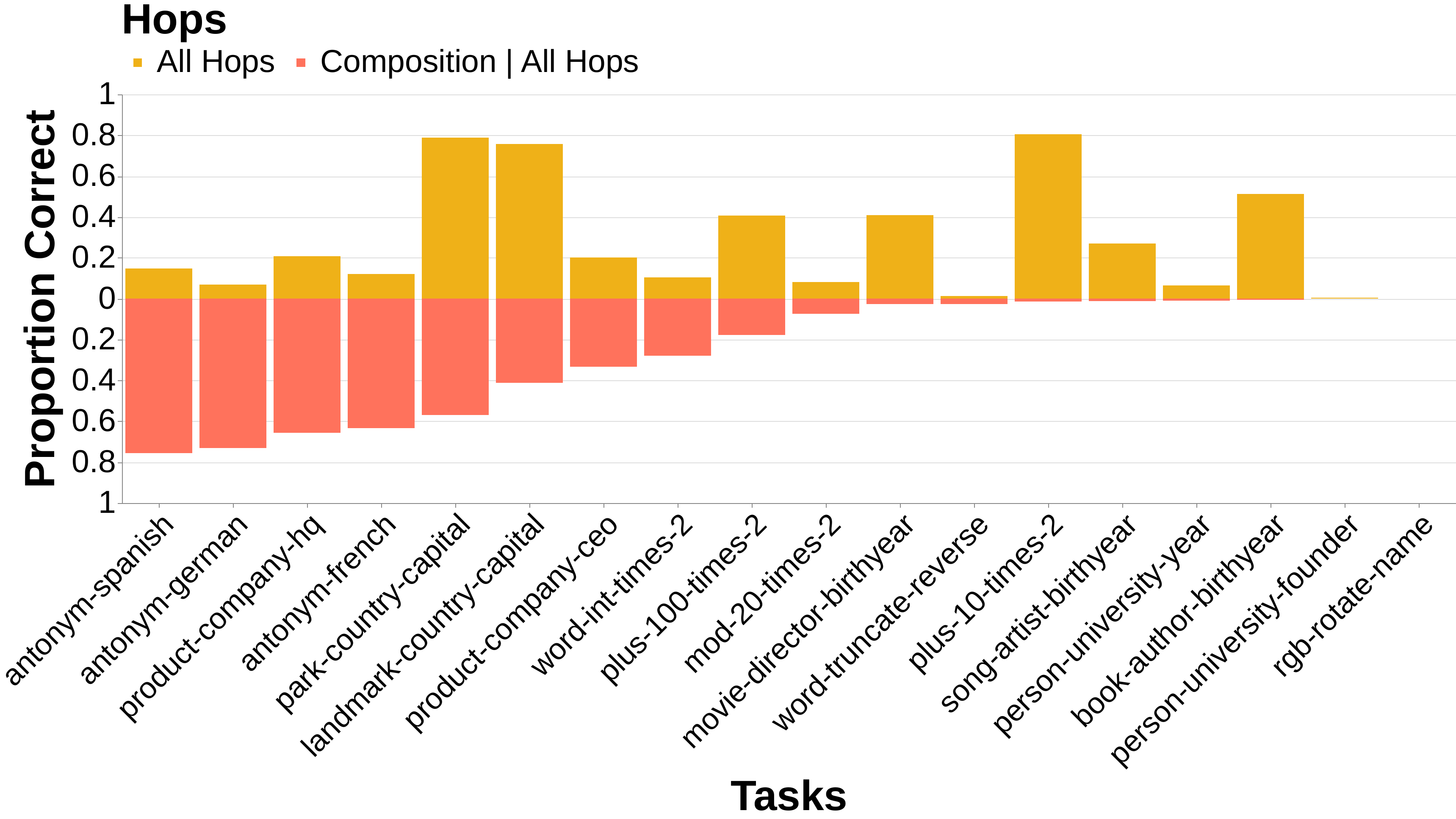}
\end{center}
\caption{Compositionality gap for Llama 3 (3B) Instruct on all our tasks. Correlation between red and yellow bars is $r^2 = \protect\input{artifacts/llama_3_3b_instruct/corr/compositionality_gap}$.}
\end{figure}

\begin{figure}[H]
\begin{center}
\includegraphics[width=0.9\linewidth]{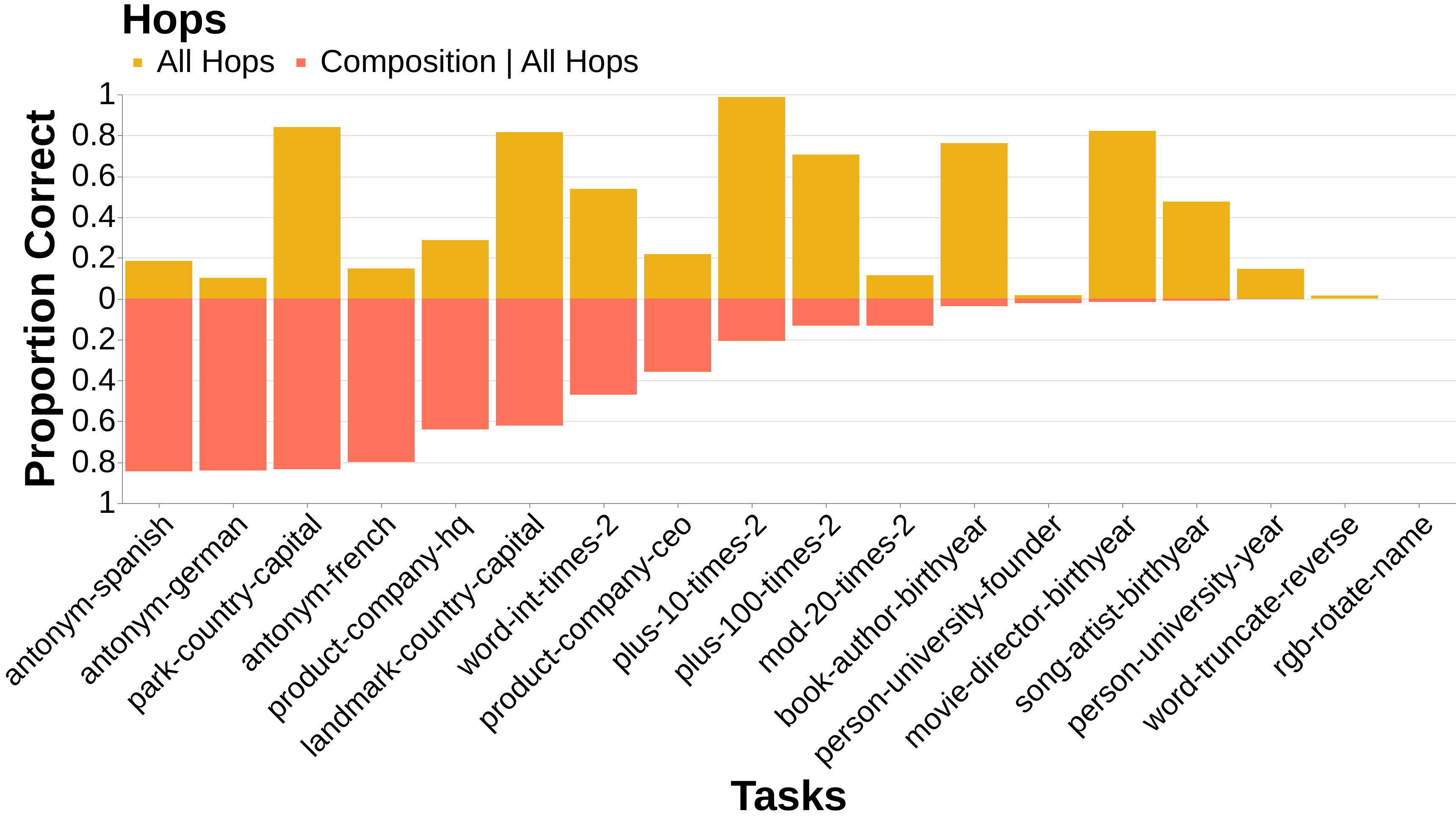}
\end{center}
\caption{Compositionality gap for Llama 3 (8B) on all our tasks. Correlation between red and yellow bars is $r^2 = \protect\input{artifacts/llama_3_8b/corr/compositionality_gap}$.}
\end{figure}

\begin{figure}[H]
\begin{center}
\includegraphics[width=0.9\linewidth]{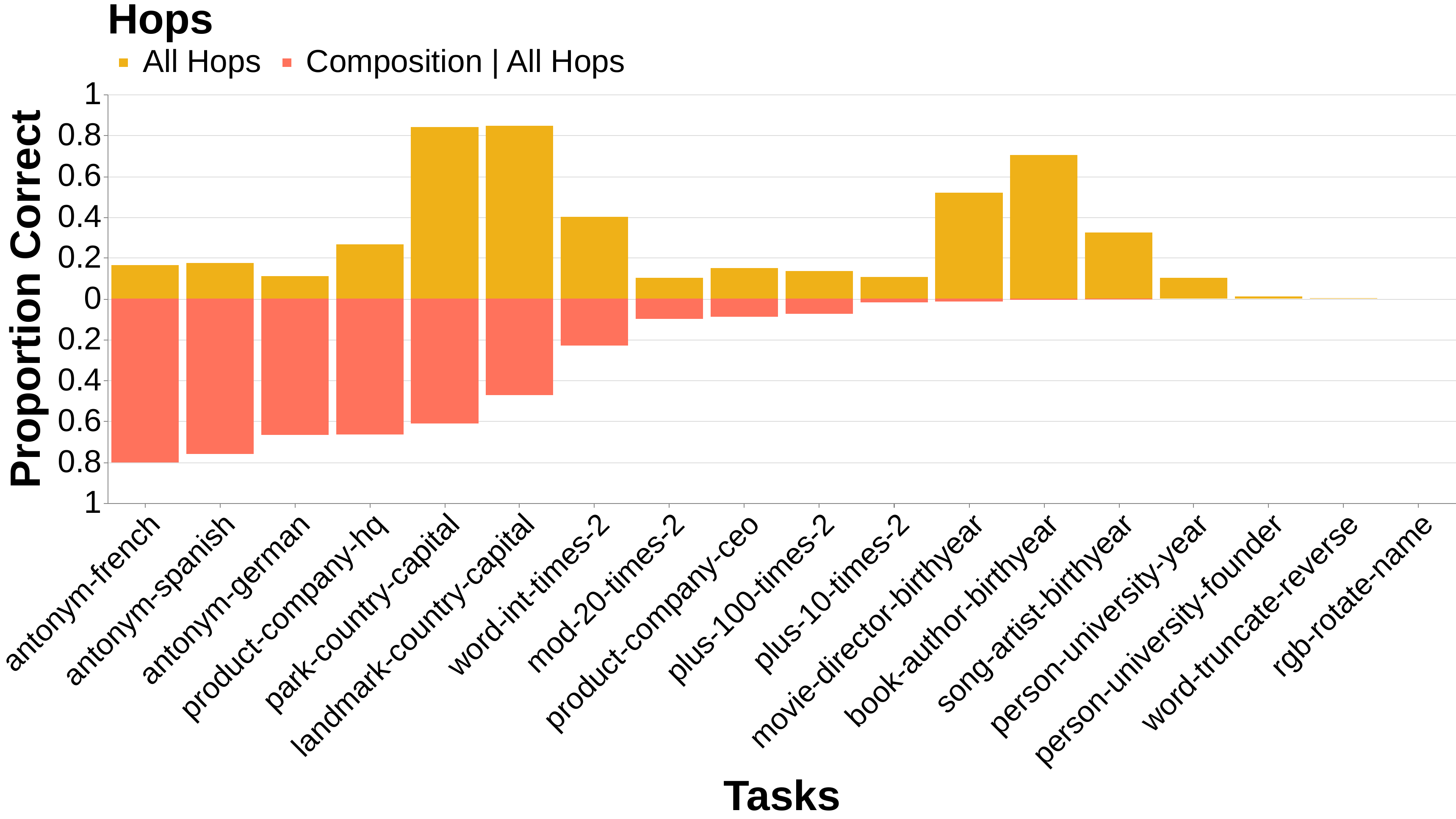}
\end{center}
\caption{Compositionality gap for OLMo 2 (7B) on all our tasks. Correlation between red and yellow bars is $r^2 = \protect\input{artifacts/olmo_2_7b/corr/compositionality_gap}$.}
\end{figure}

\begin{figure}[H]
\begin{center}
\includegraphics[width=0.9\linewidth]{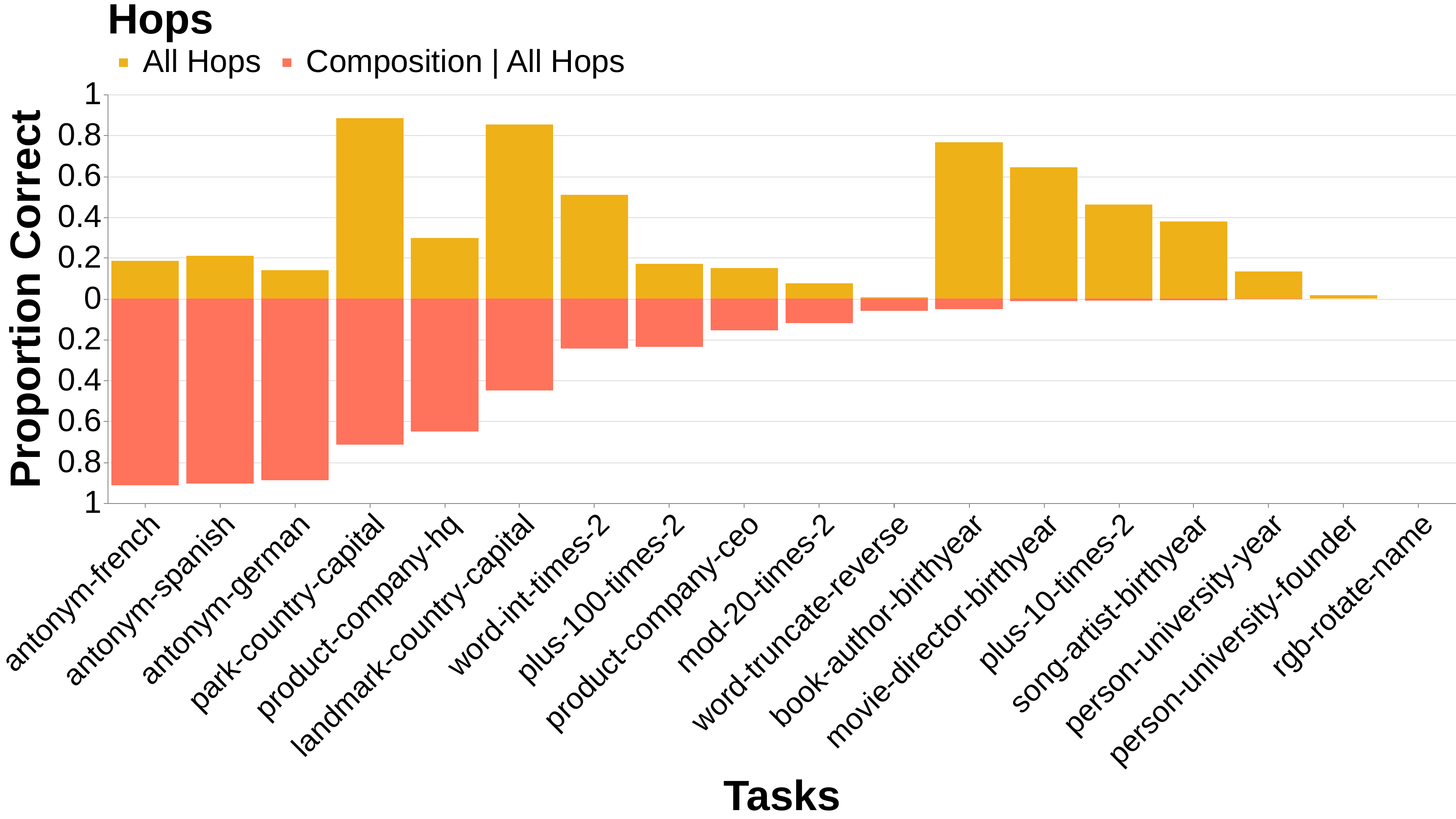}
\end{center}
\caption{Compositionality gap for OLMo 2 (13B) on all our tasks. Correlation between red and yellow bars is $r^2 = \protect\input{artifacts/olmo_2_13b/corr/compositionality_gap}$.}
\end{figure}

\section{Processing Signatures (Correct)}\label{app:lens-correct}

\begin{figure}[H]
    \begin{center}\begin{adjustbox}{width=\linewidth}
    \begin{tabular}{cccc}
        \includegraphics[width=0.25\linewidth]{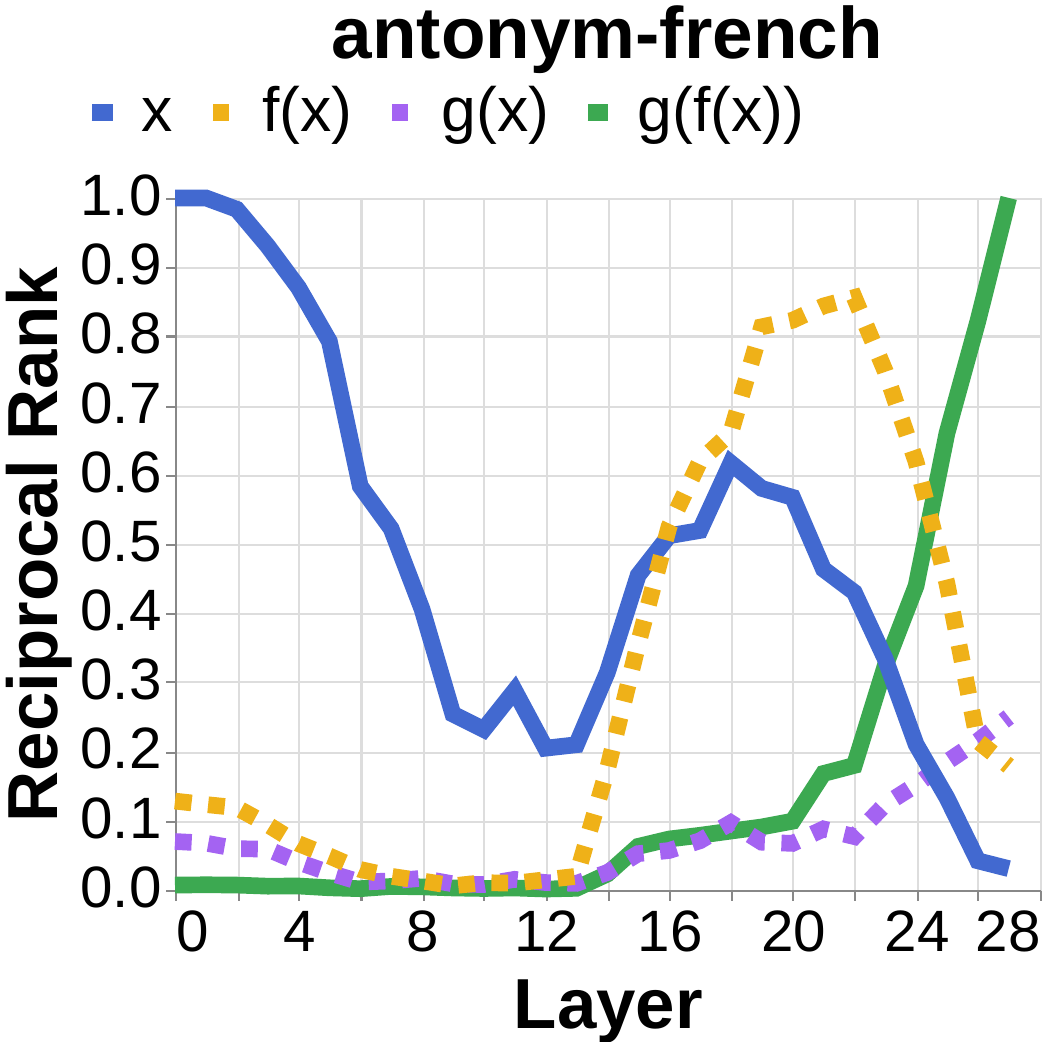}
        & \includegraphics[width=0.25\linewidth]{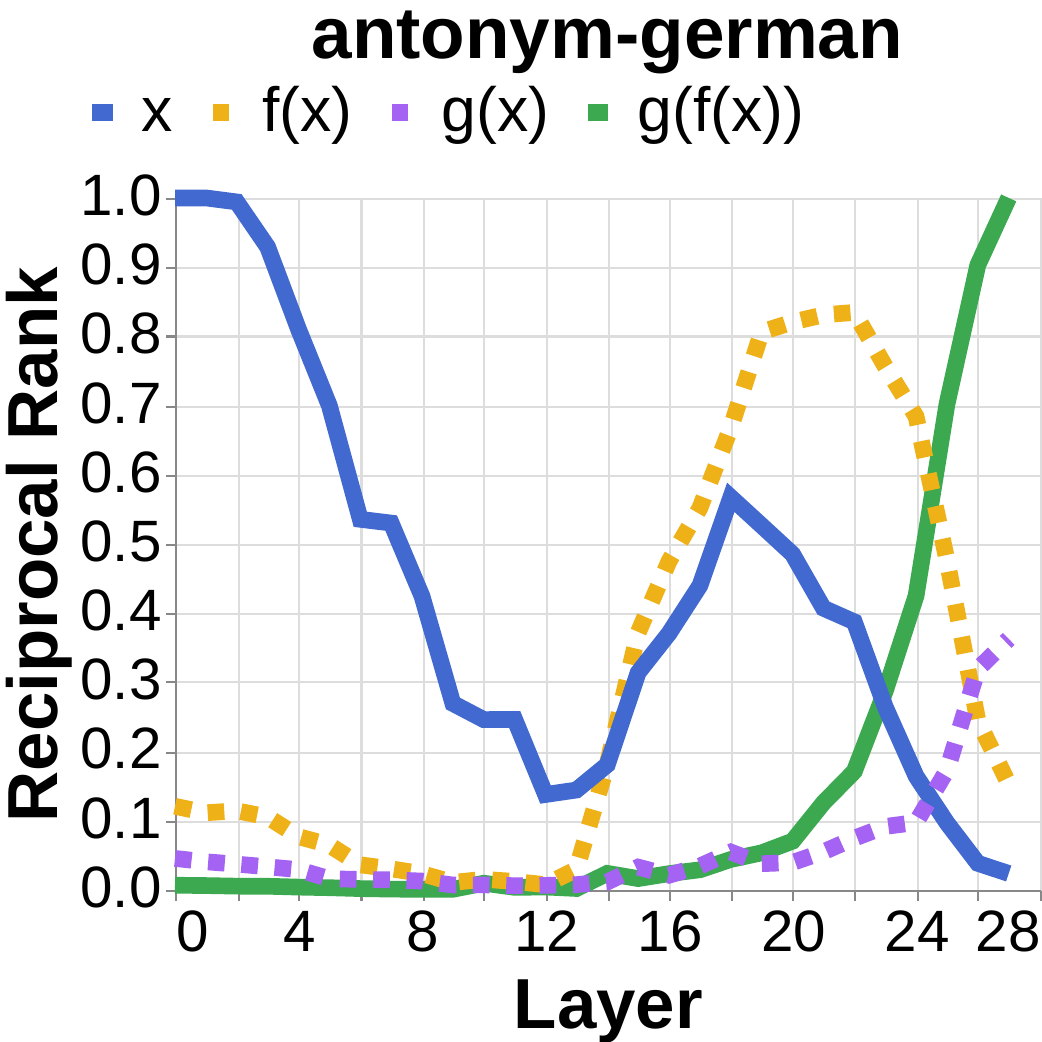}
        & \includegraphics[width=0.25\linewidth]{artifacts/llama_3_3b/lens/antonym-spanish_correct.pdf}
        & \includegraphics[width=0.25\linewidth]{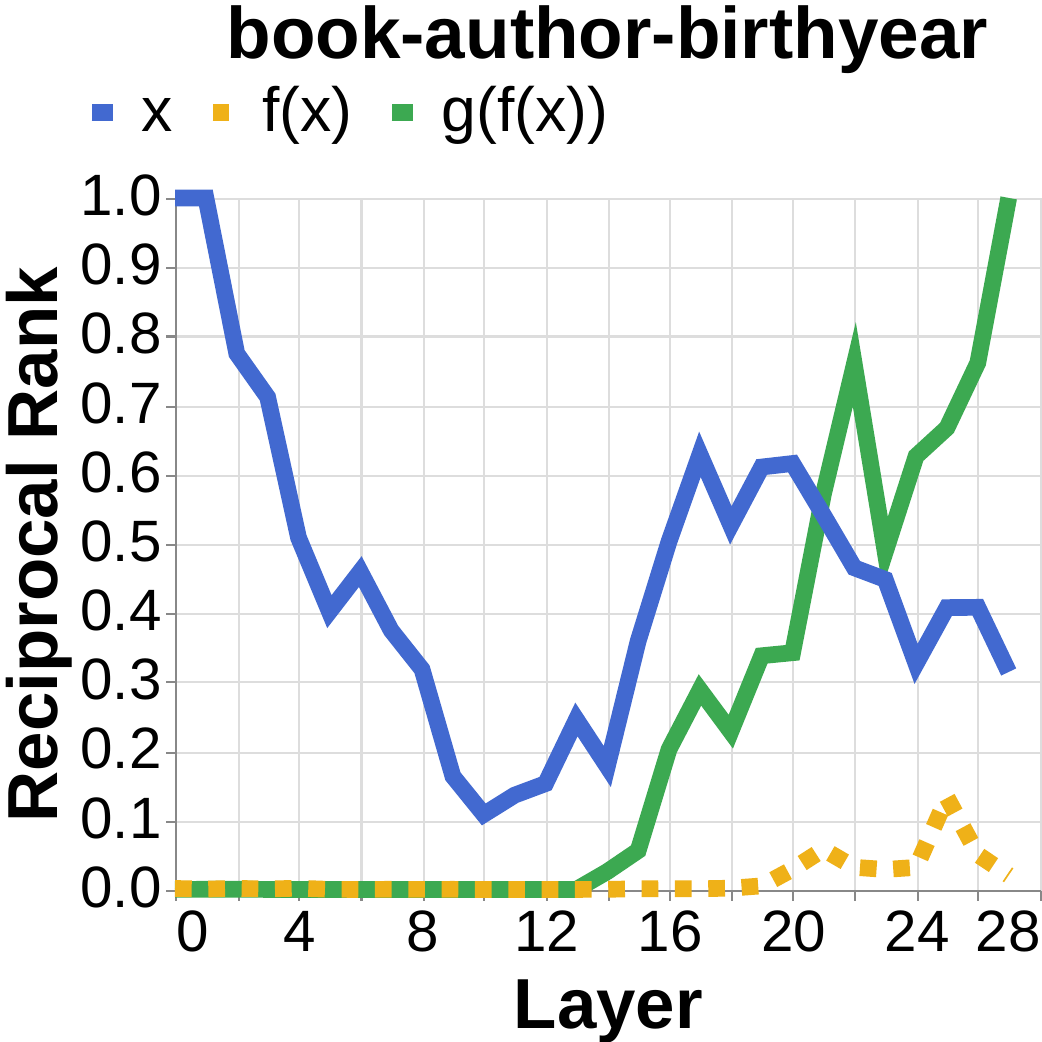} \\
        \includegraphics[width=0.25\linewidth]{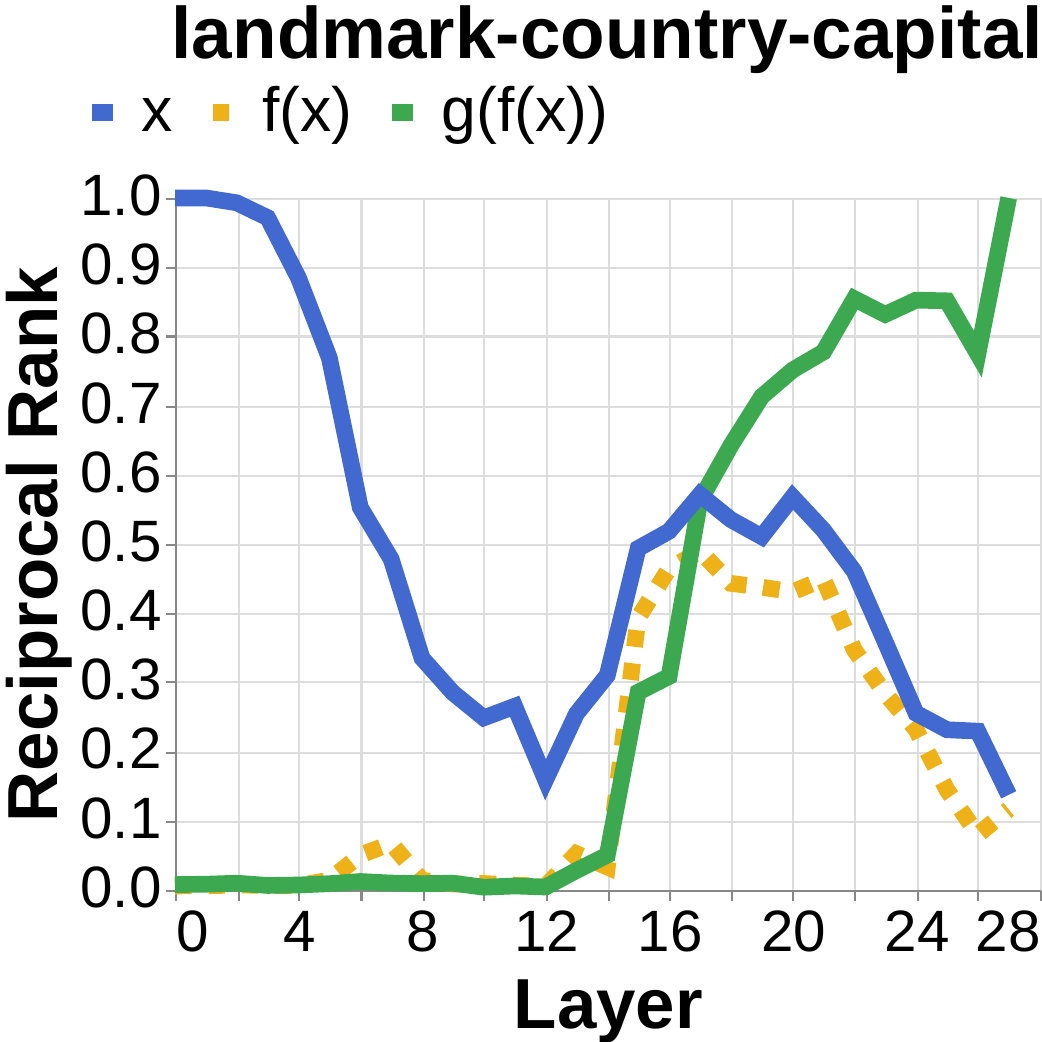}
        & \includegraphics[width=0.25\linewidth]{artifacts/llama_3_3b/lens/movie-director-birthyear_correct.pdf}
        & \includegraphics[width=0.25\linewidth]{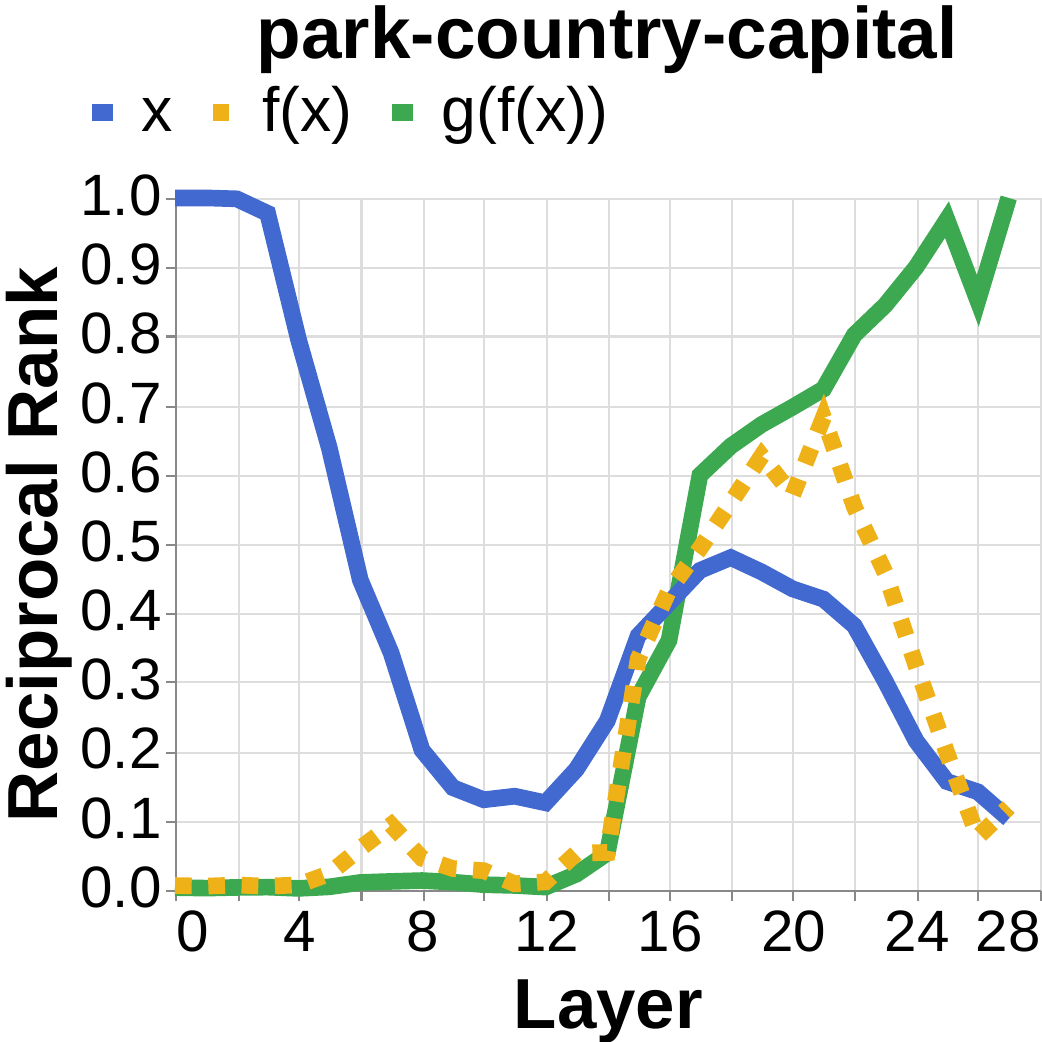}
        & \includegraphics[width=0.25\linewidth]{artifacts/llama_3_3b/lens/plus-hundred-times-two_correct.pdf} \\
        \includegraphics[width=0.25\linewidth]{artifacts/llama_3_3b/lens/plus-ten-times-two_correct.pdf}
        & \includegraphics[width=0.25\linewidth]{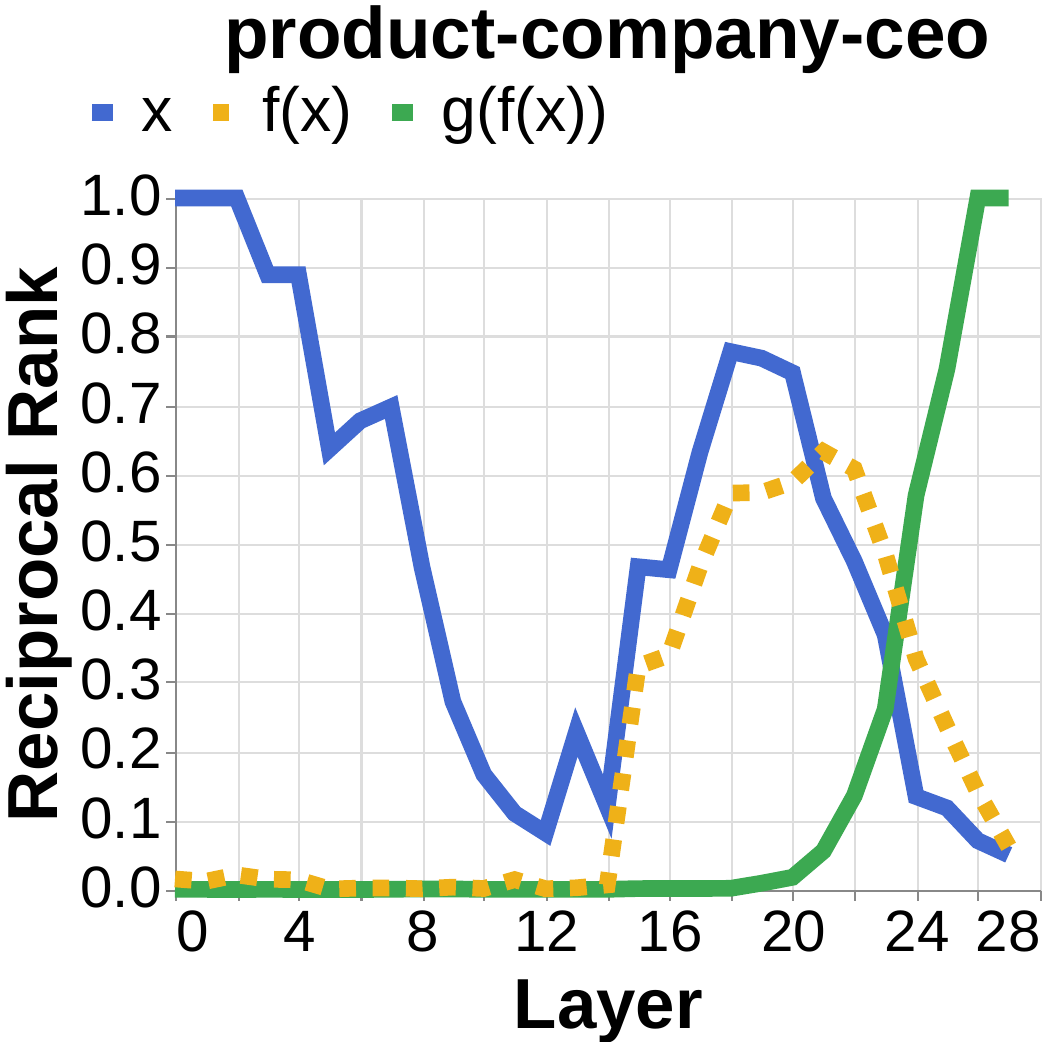}
        & \includegraphics[width=0.25\linewidth]{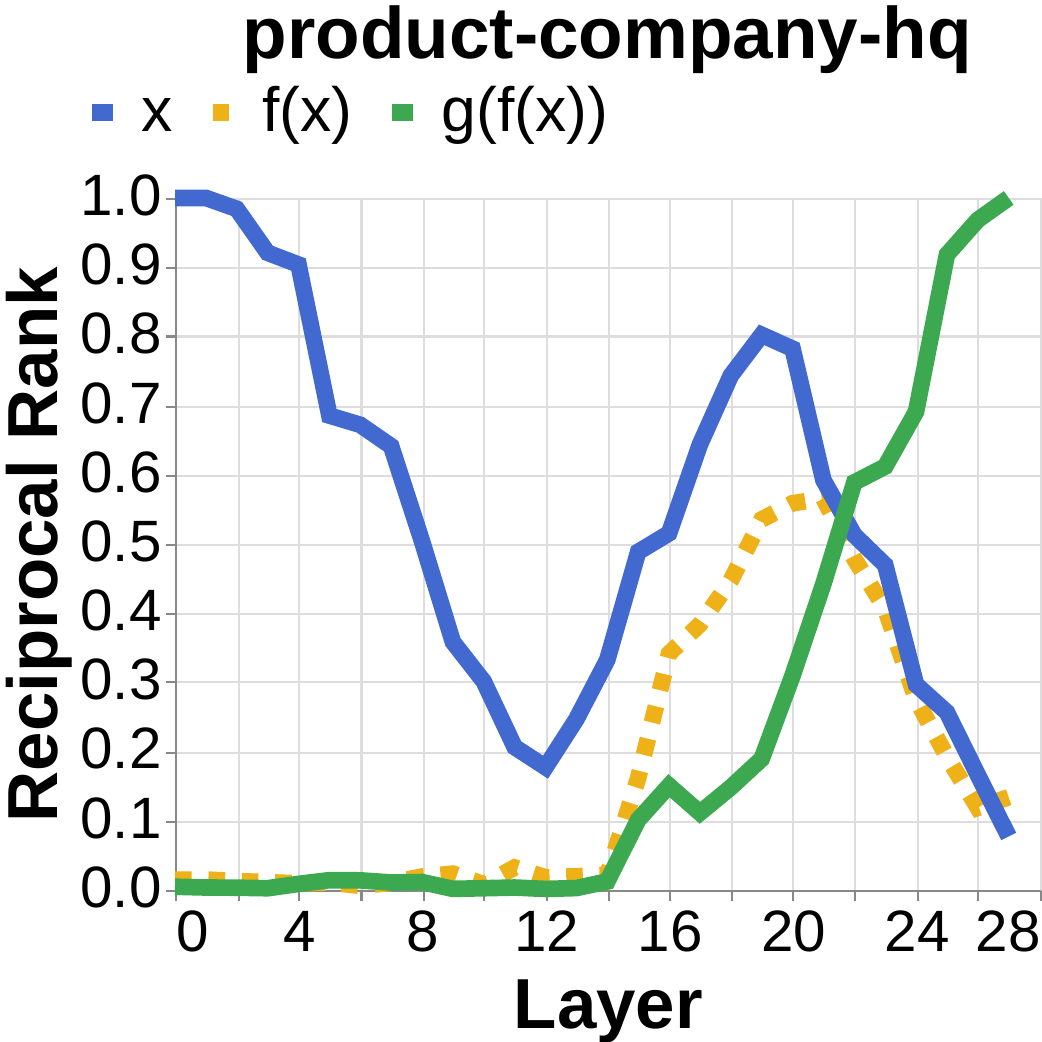}
        & \includegraphics[width=0.25\linewidth]{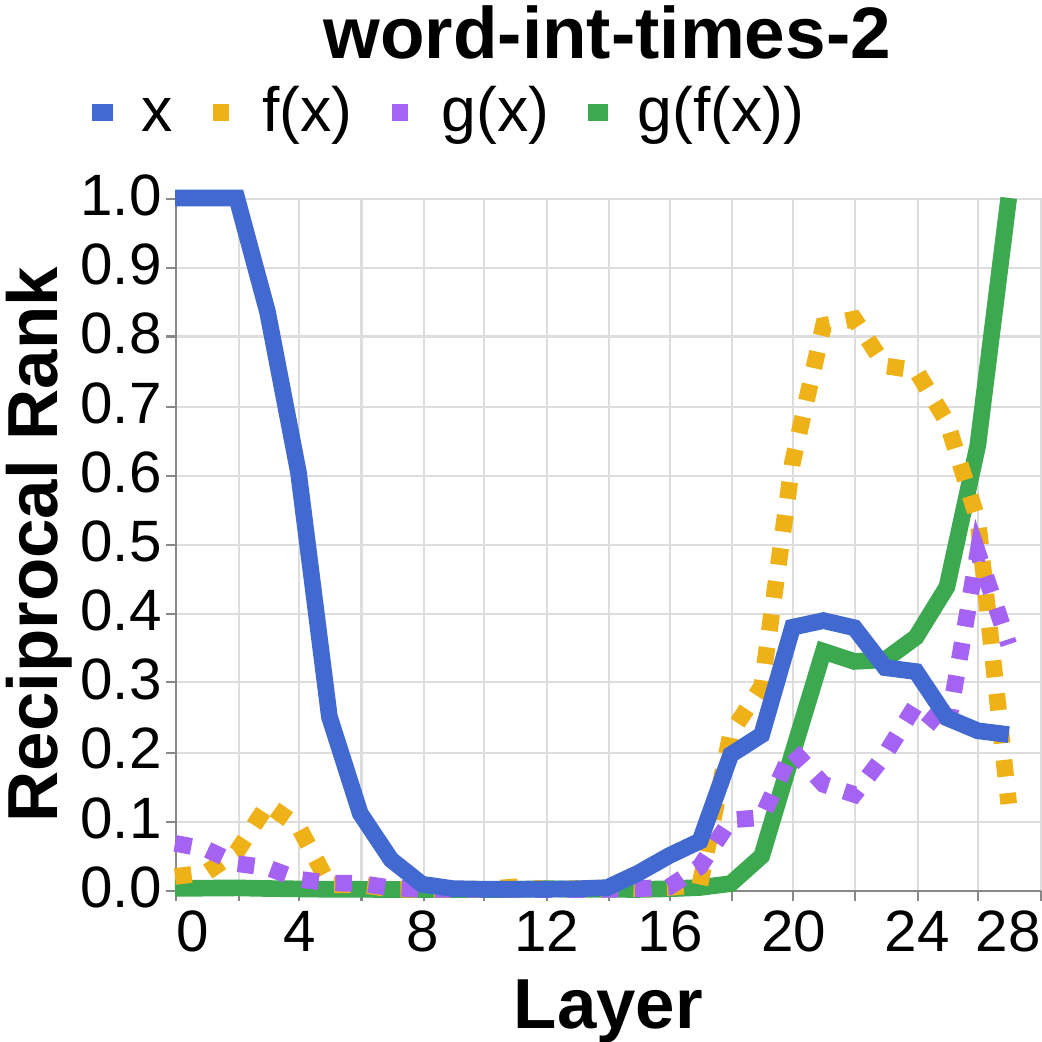} \\
    \end{tabular}
    \end{adjustbox}\end{center}
    \caption{Aggregate processing signatures for each of our tasks, in which Llama 3 (3B) correctly solves all hops and the composition for at least 10 examples.}
\end{figure}

\begin{figure}[H]
    \begin{center}\begin{adjustbox}{width=\linewidth}
    \begin{tabular}{cccc}
        \includegraphics[width=0.25\linewidth]{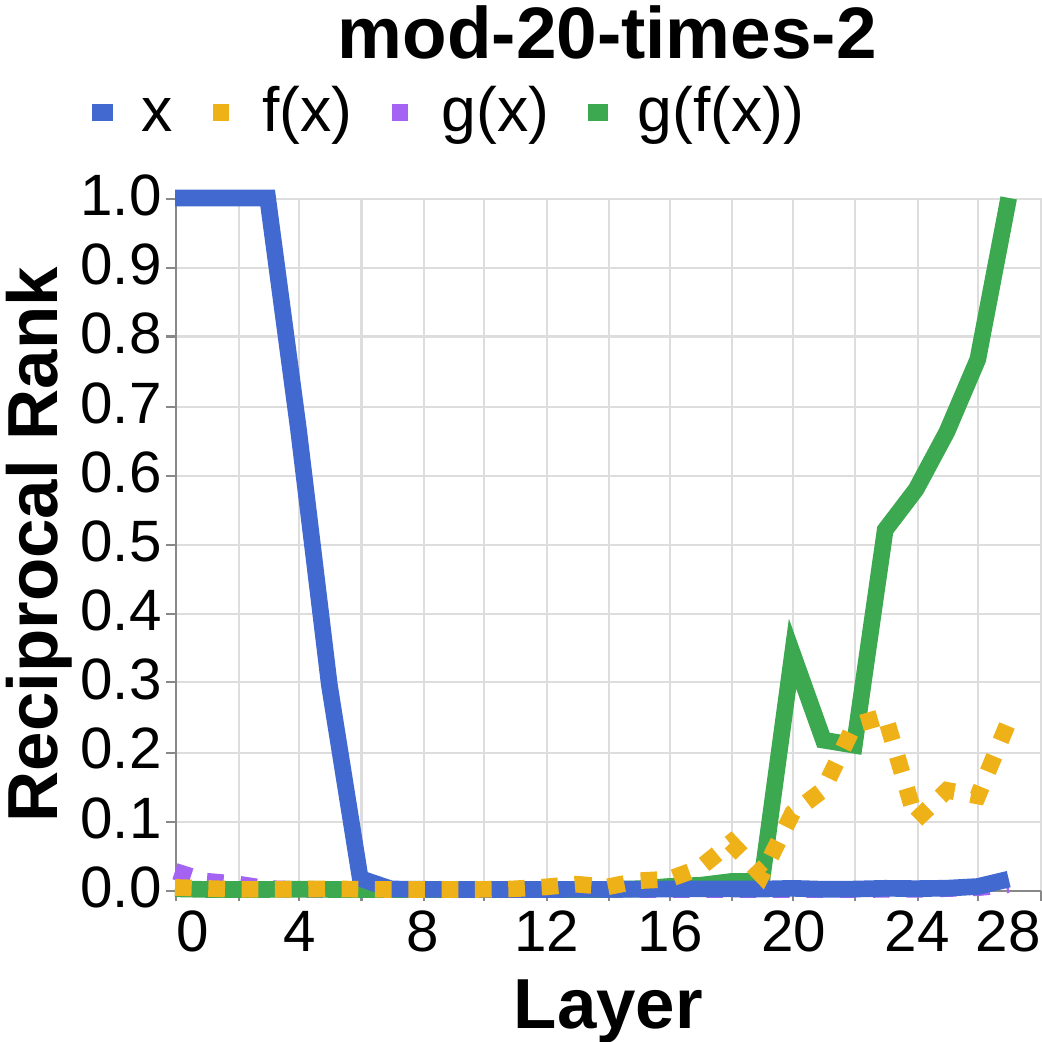}
        & \includegraphics[width=0.25\linewidth]{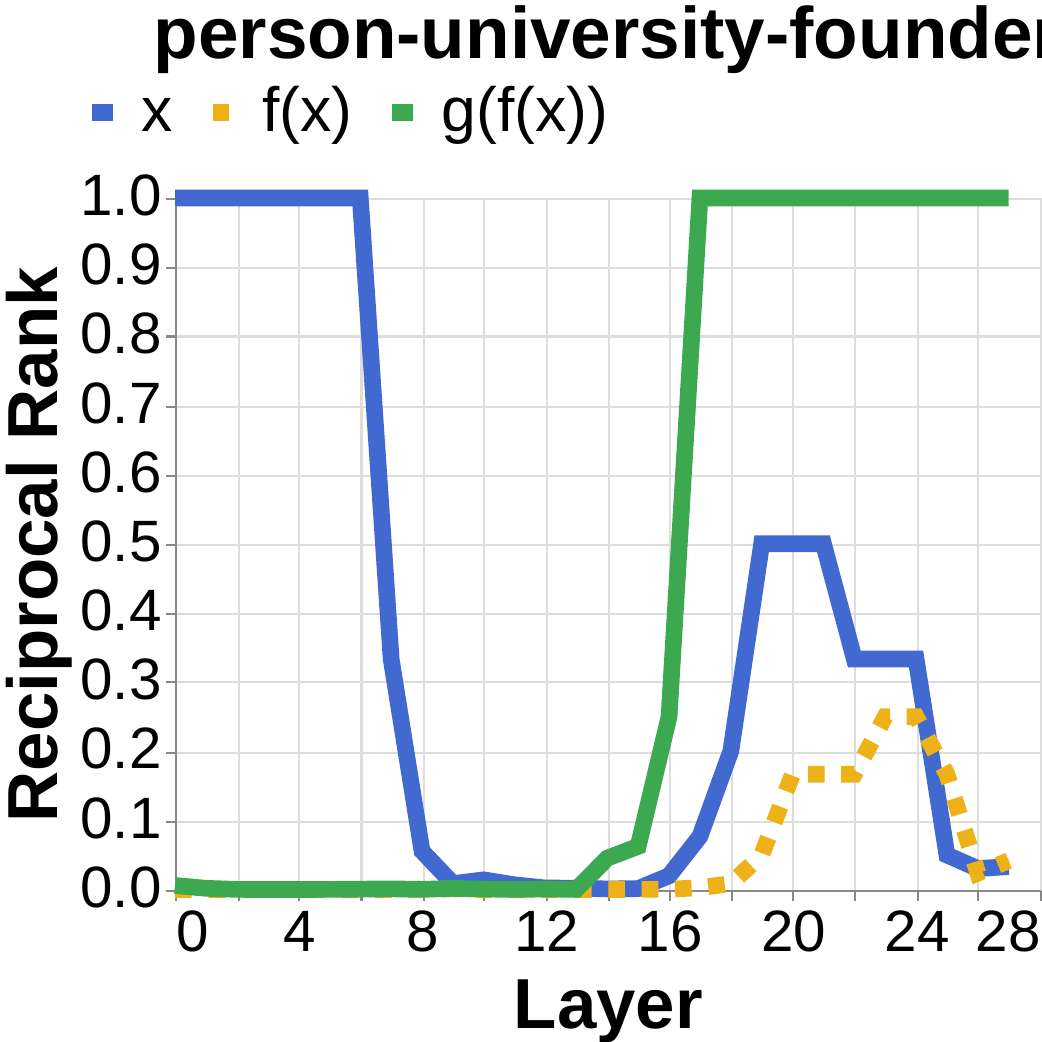}
        & \includegraphics[width=0.25\linewidth]{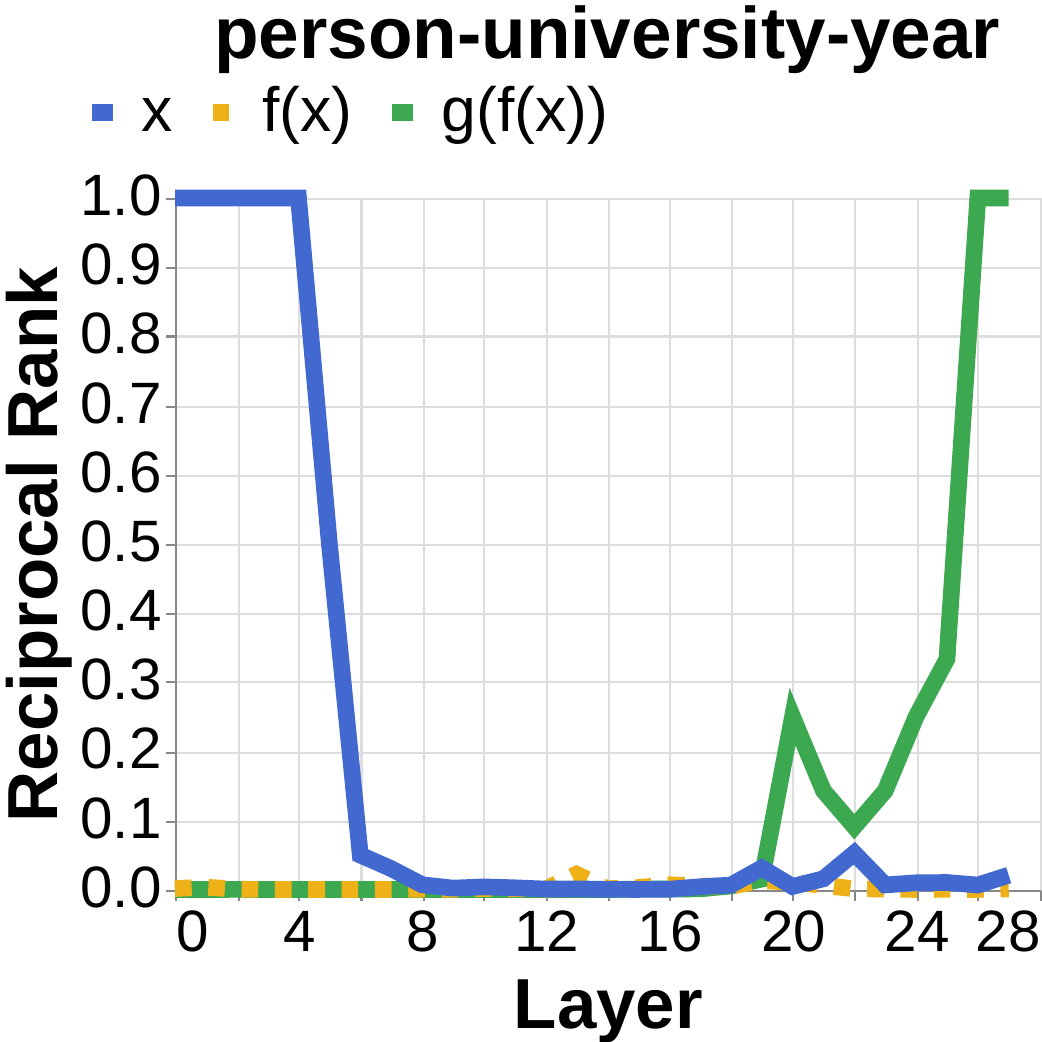}
        & \includegraphics[width=0.25\linewidth]{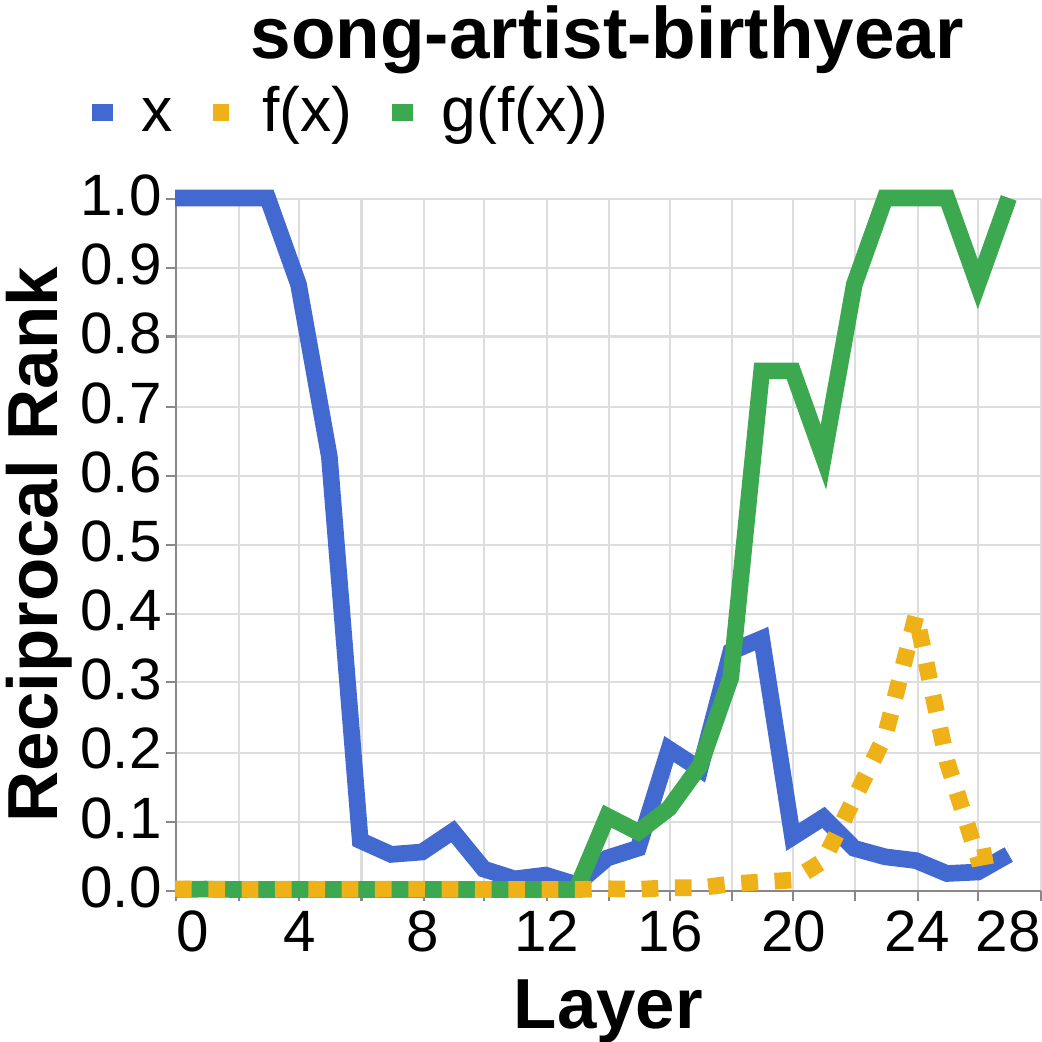} \\
        \includegraphics[width=0.25\linewidth]{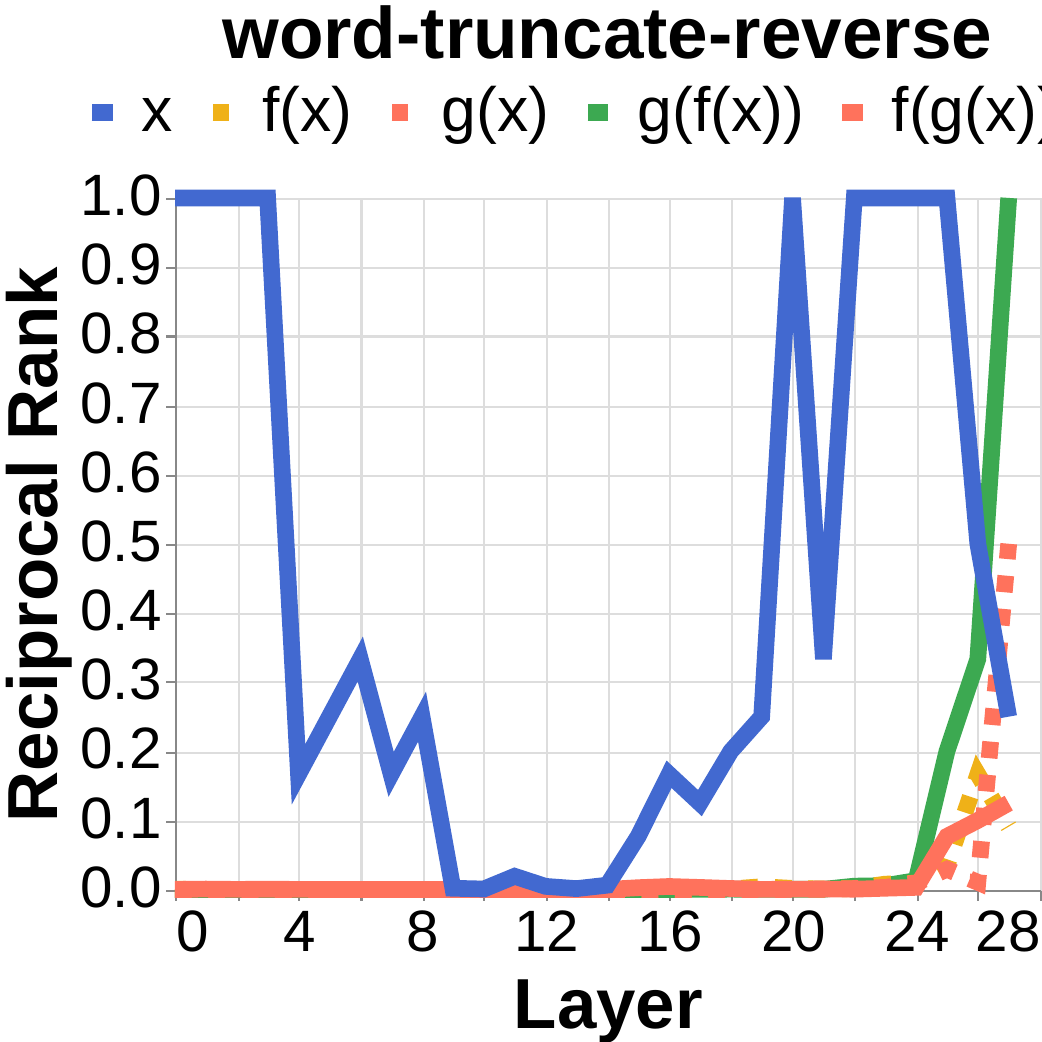} & & & \\
    \end{tabular}
    \end{adjustbox}\end{center}
    \caption{Aggregate processing signatures for each of our tasks, in which Llama 3 (3B) correctly solves all hops and the composition for less than 10 examples.}
    \label{fig:lens-excluded-correct}
\end{figure}

\newpage
\section{Processing Signatures (Incorrect)}\label{app:lens-incorrect}

\begin{figure}[H]
    \begin{center}\begin{adjustbox}{width=\linewidth}
    \begin{tabular}{cccc}
        \includegraphics[width=0.25\linewidth]{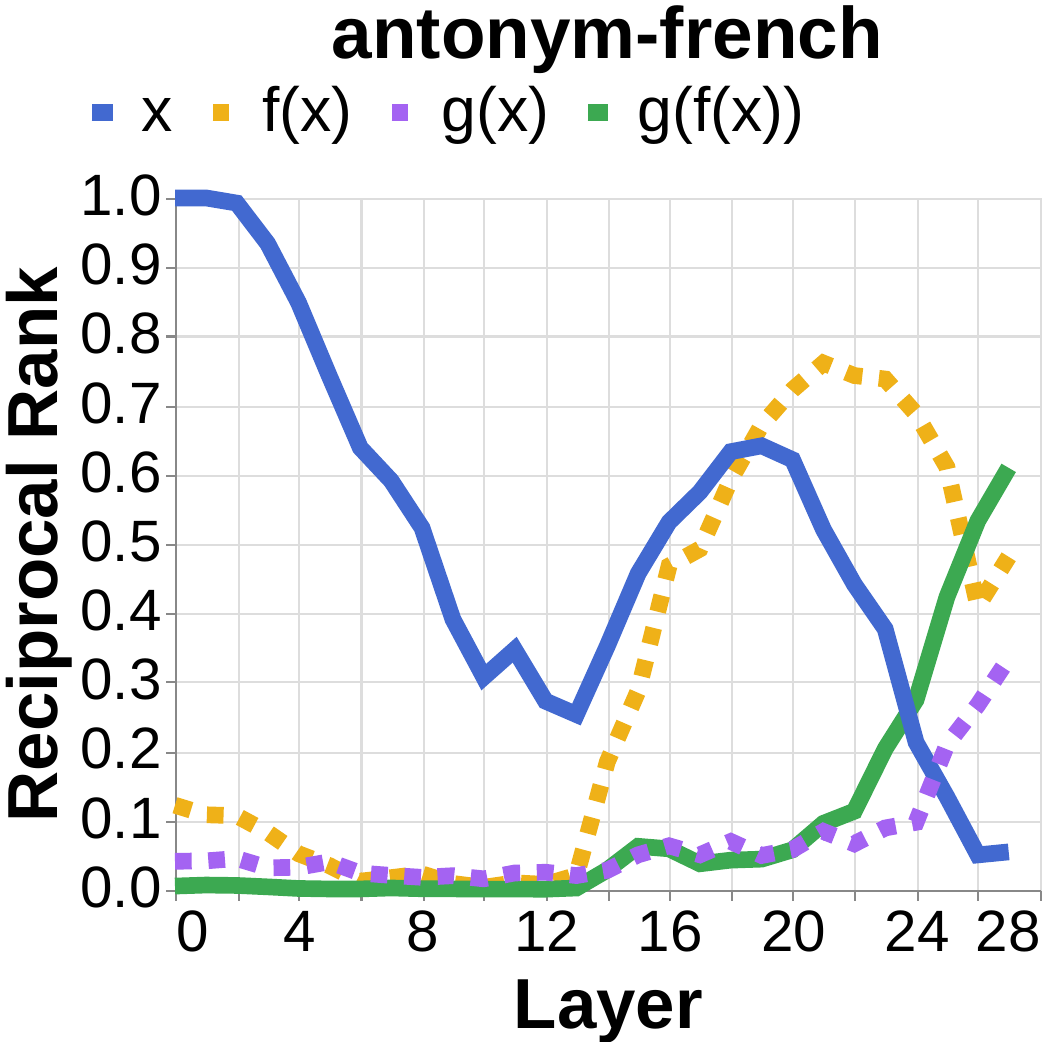}
        & \includegraphics[width=0.25\linewidth]{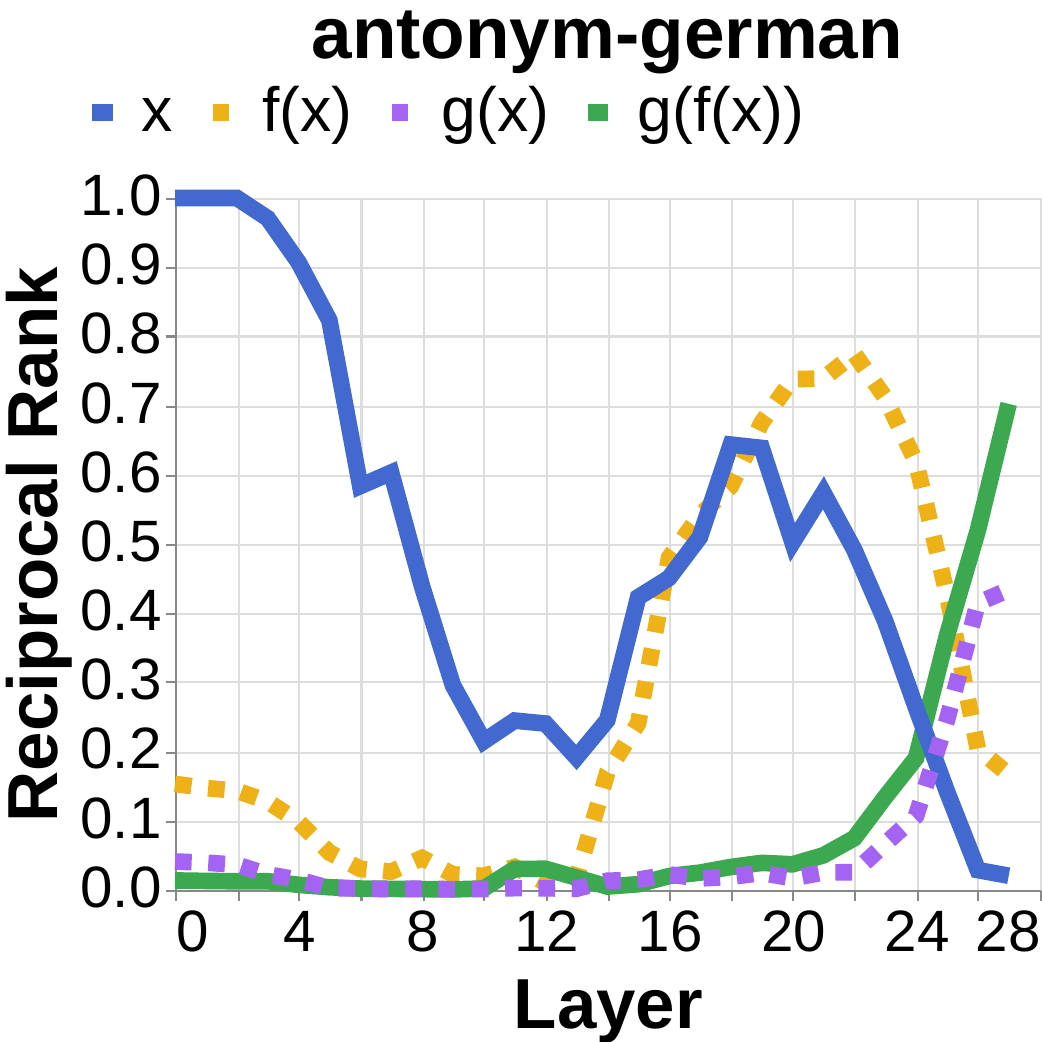}
        & \includegraphics[width=0.25\linewidth]{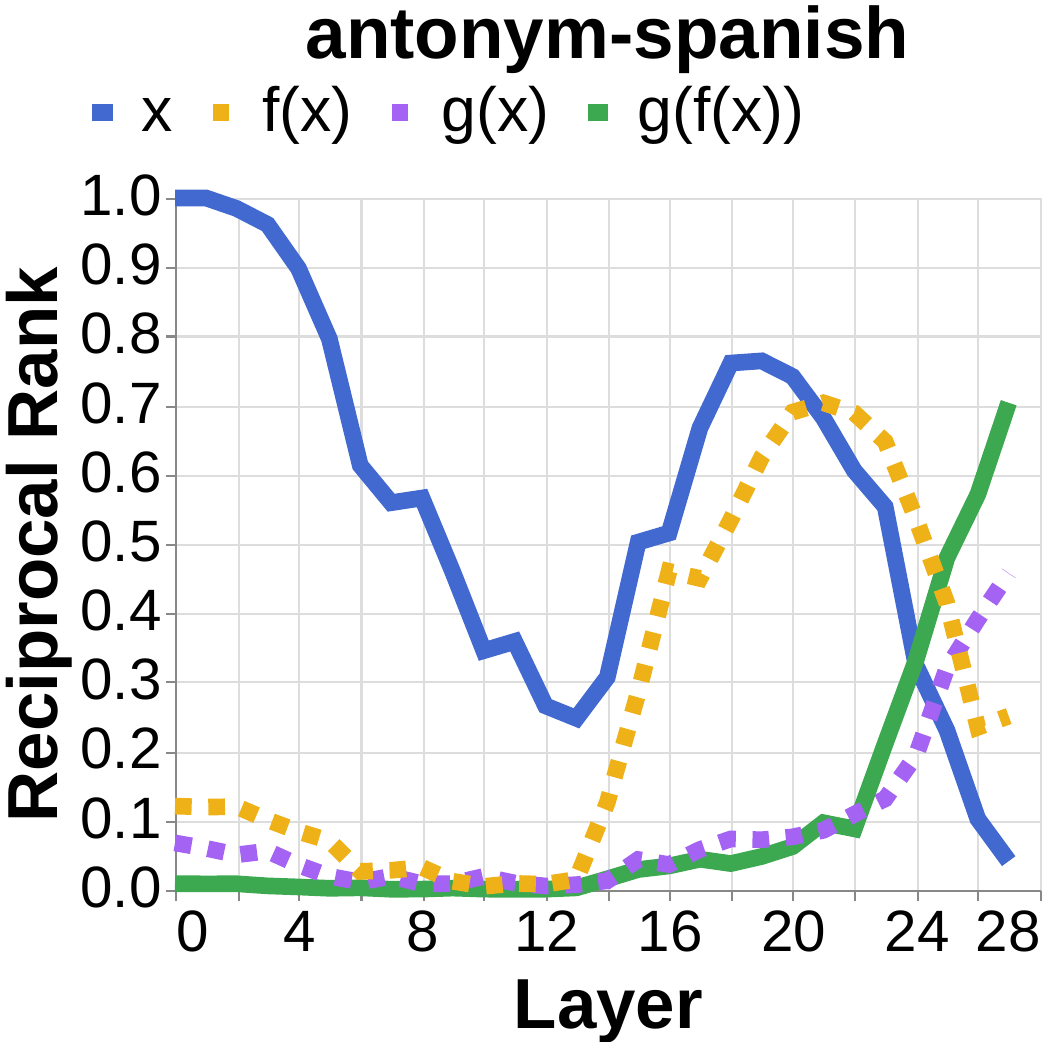}
        & \includegraphics[width=0.25\linewidth]{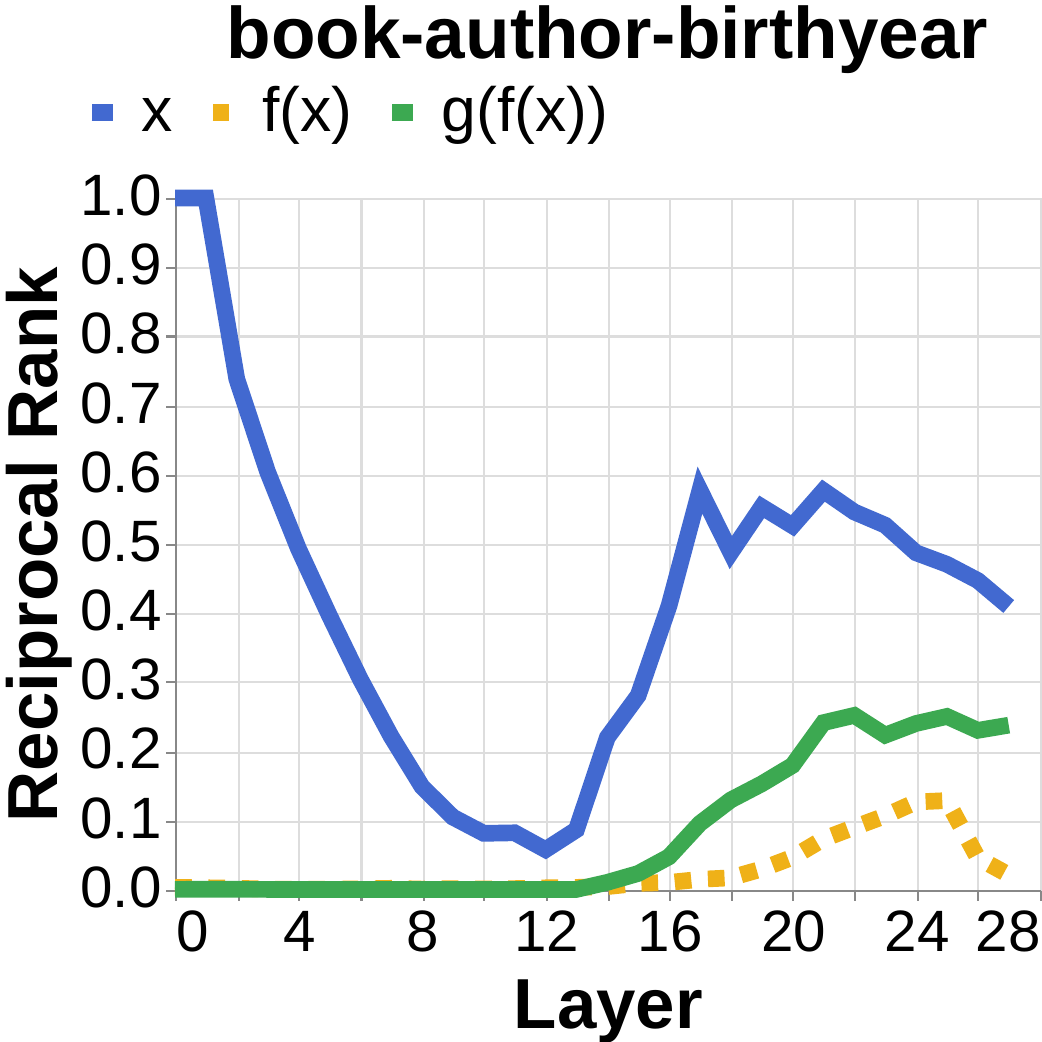} \\
        \includegraphics[width=0.25\linewidth]{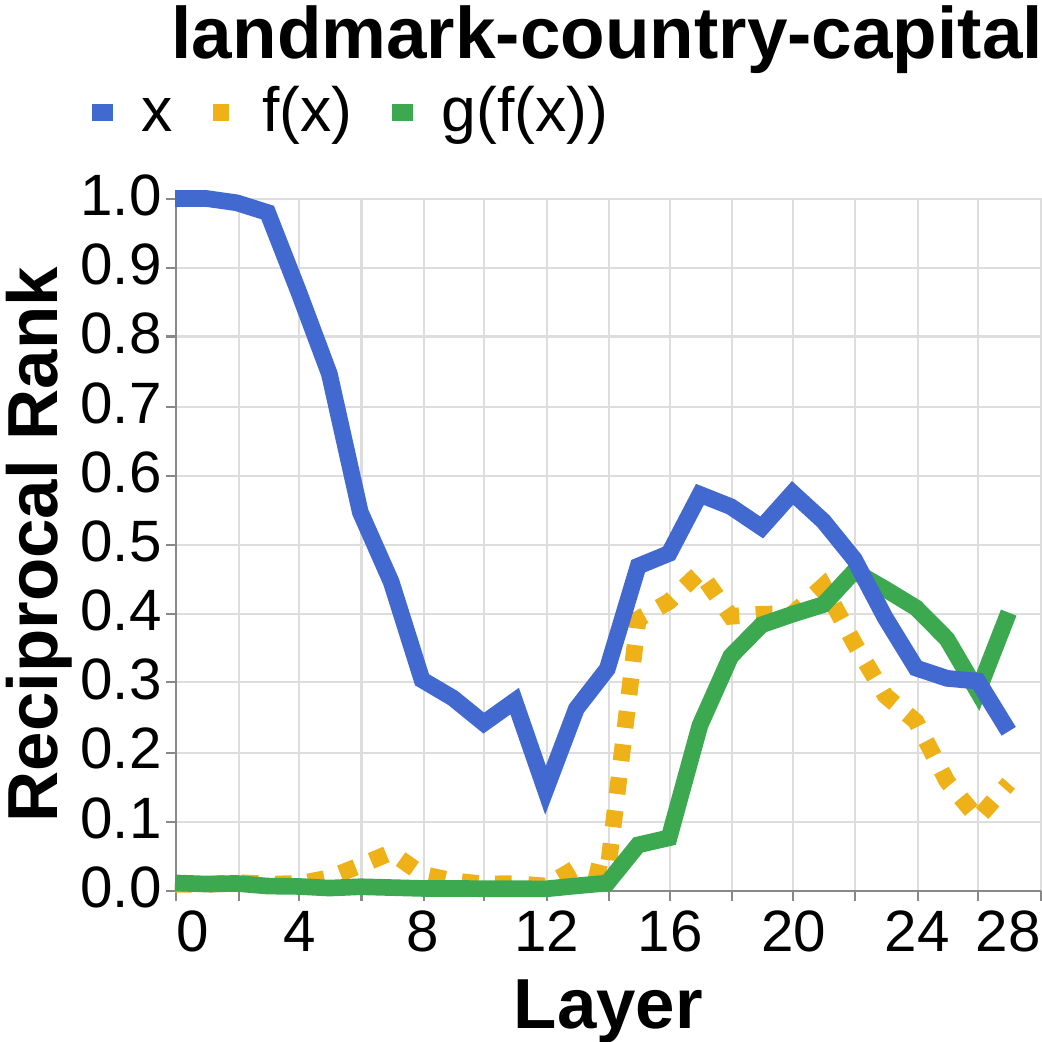}
        & \includegraphics[width=0.25\linewidth]{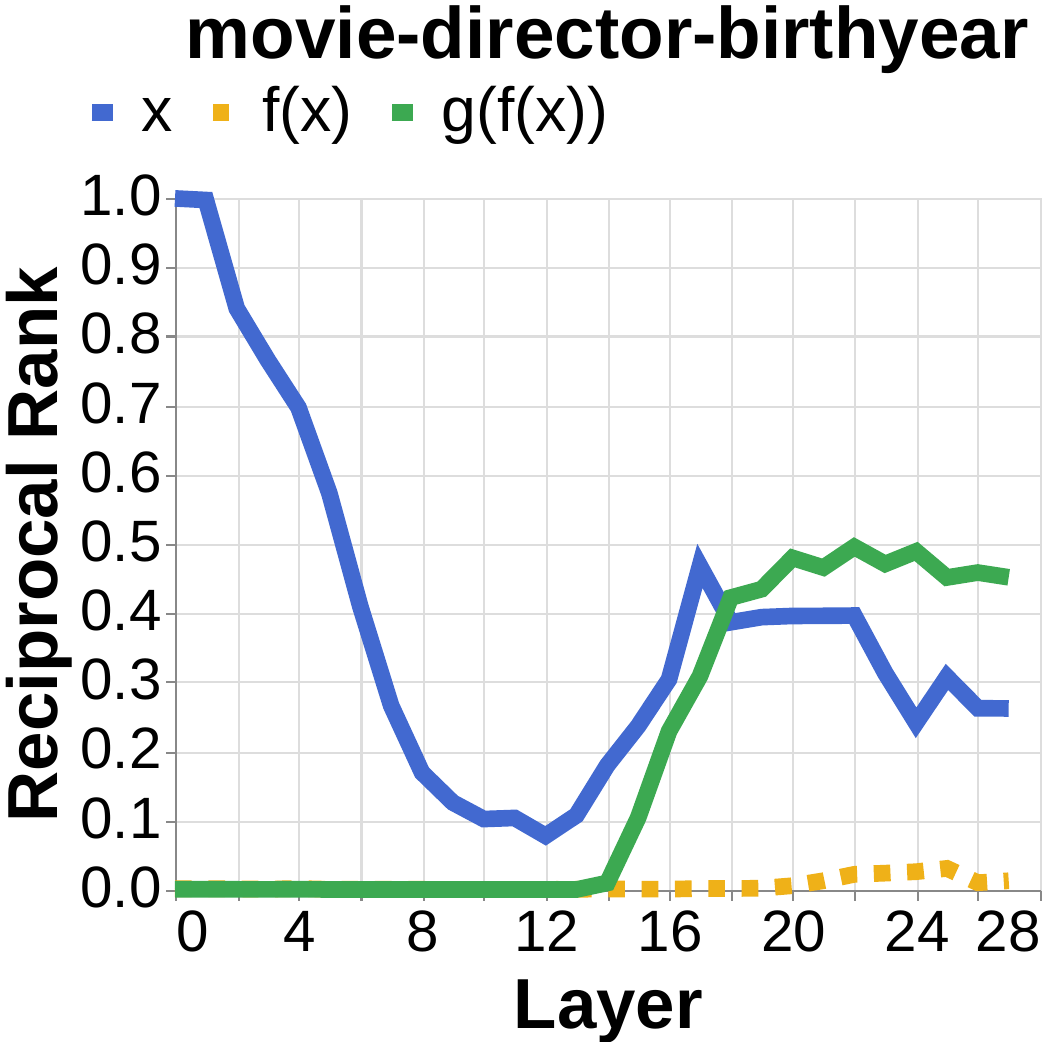}
        & \includegraphics[width=0.25\linewidth]{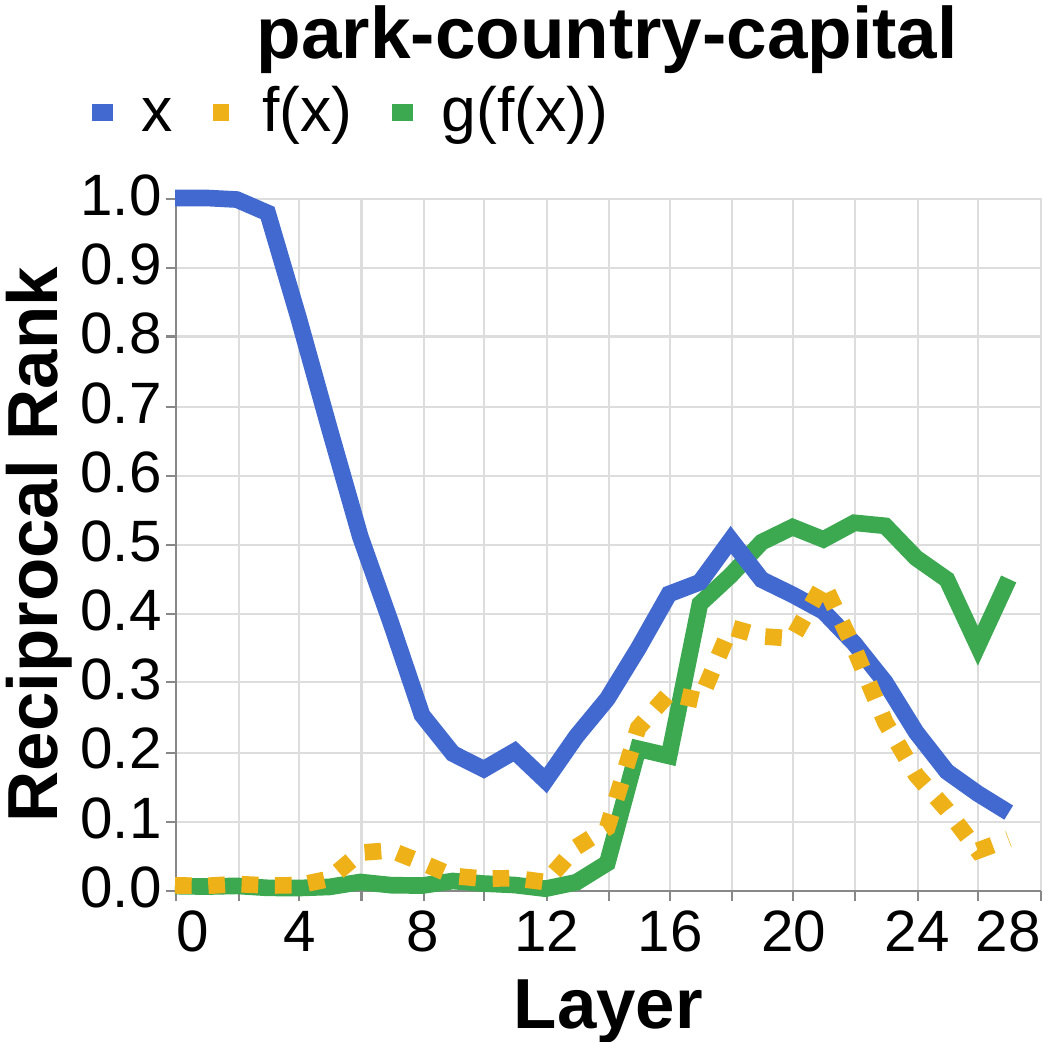}
        & \includegraphics[width=0.25\linewidth]{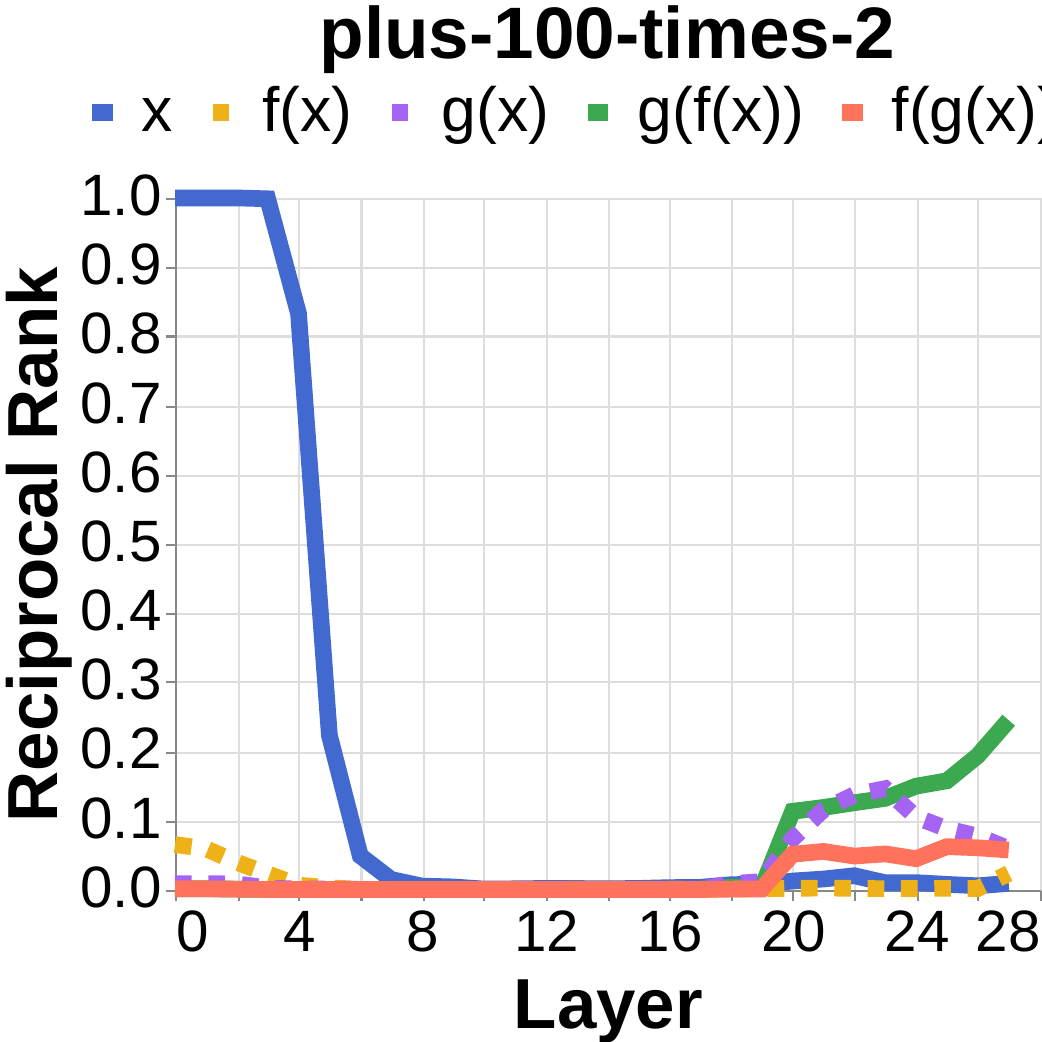} \\
        \includegraphics[width=0.25\linewidth]{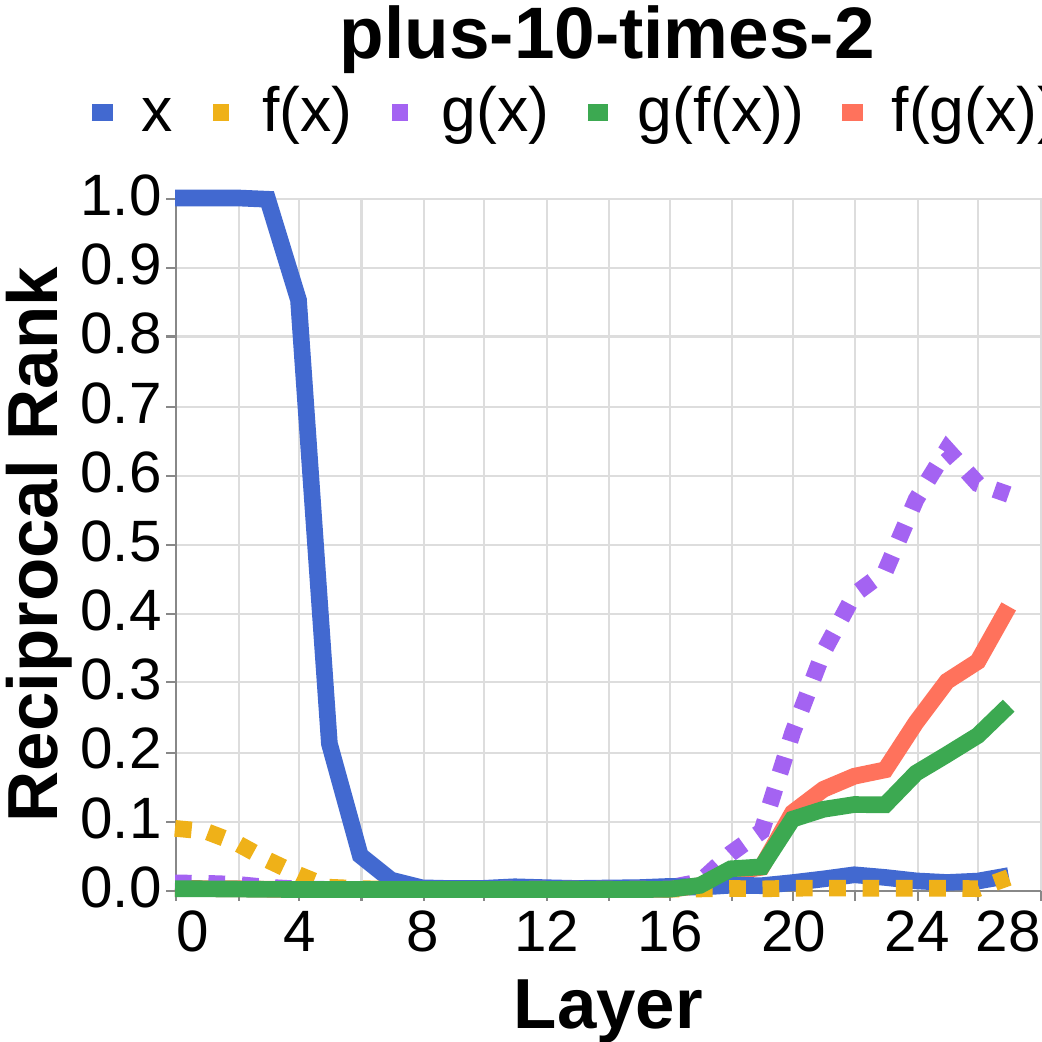}
        & \includegraphics[width=0.25\linewidth]{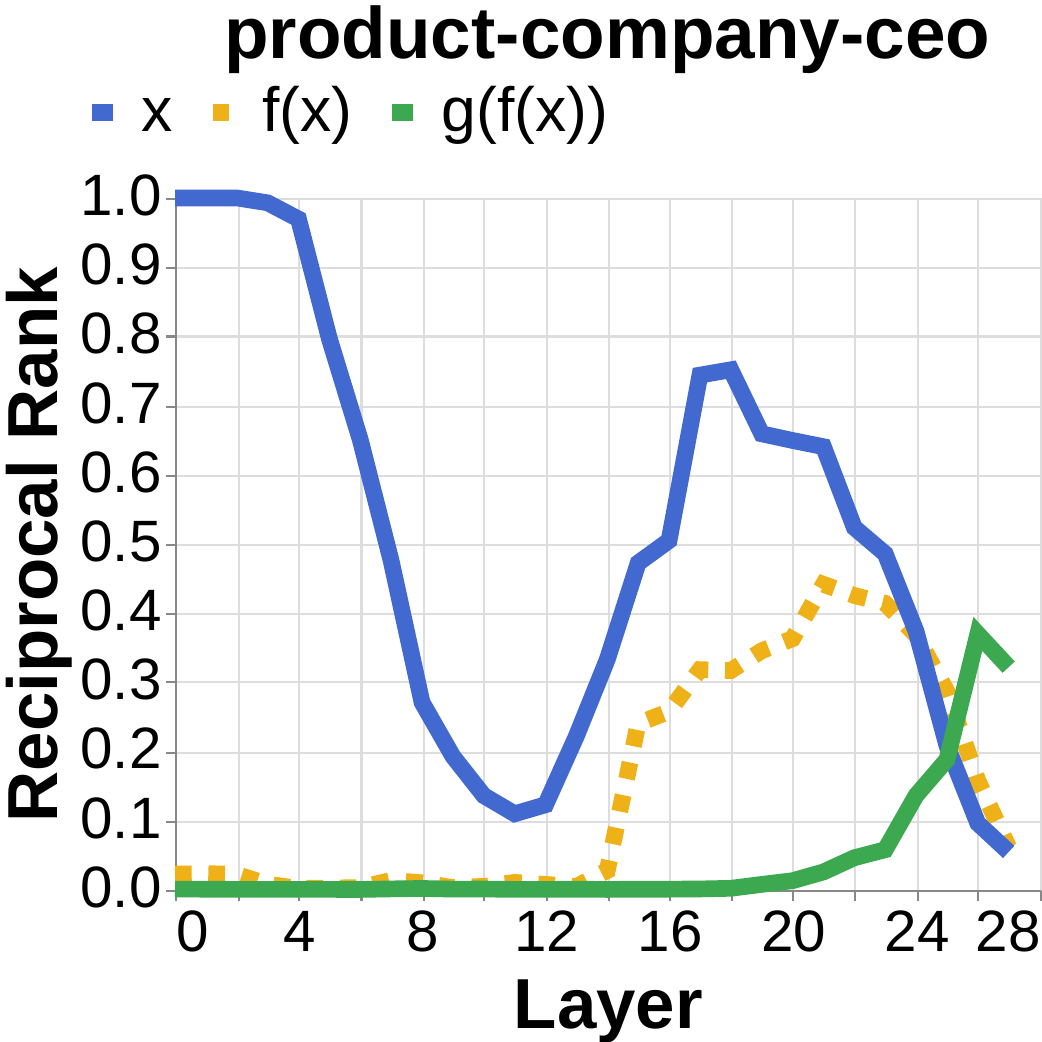}
        & \includegraphics[width=0.25\linewidth]{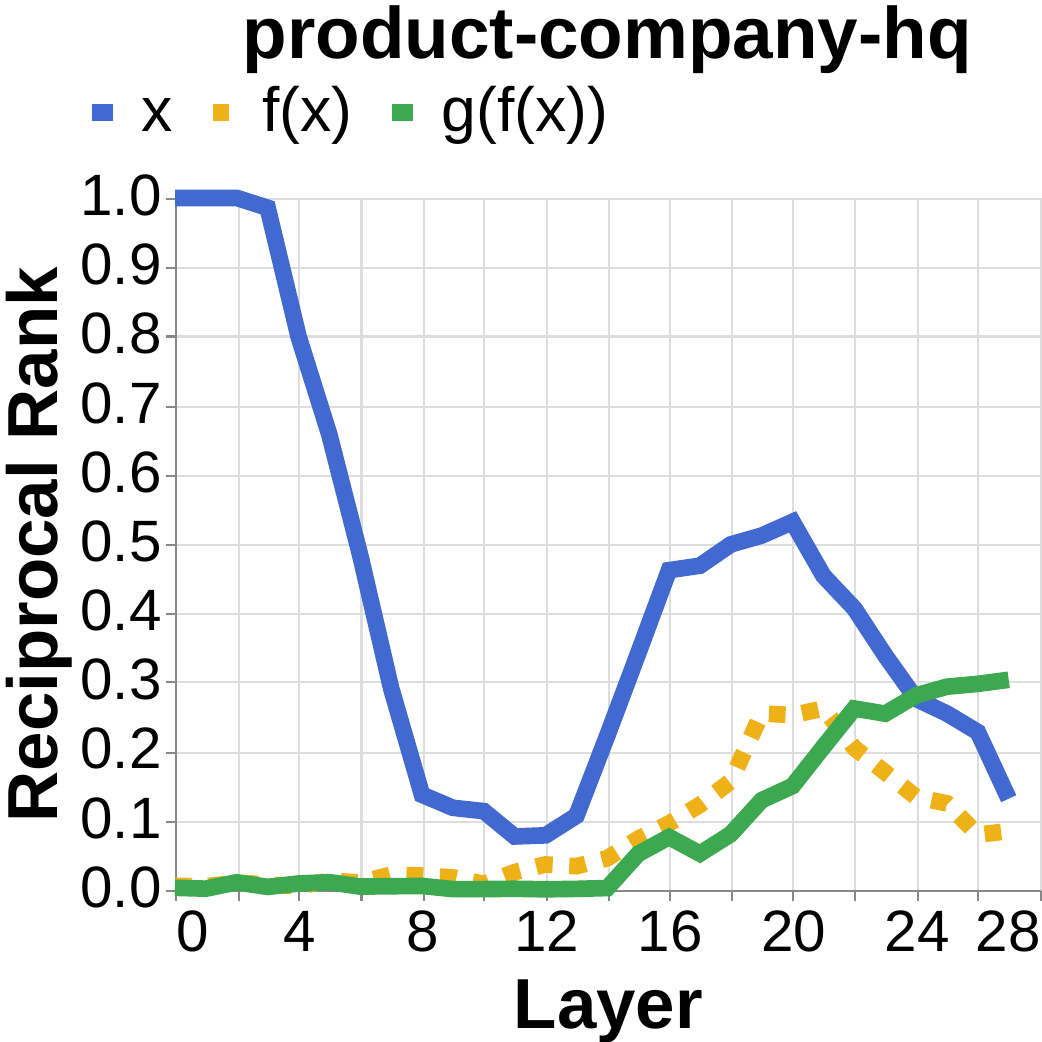}
        & \includegraphics[width=0.25\linewidth]{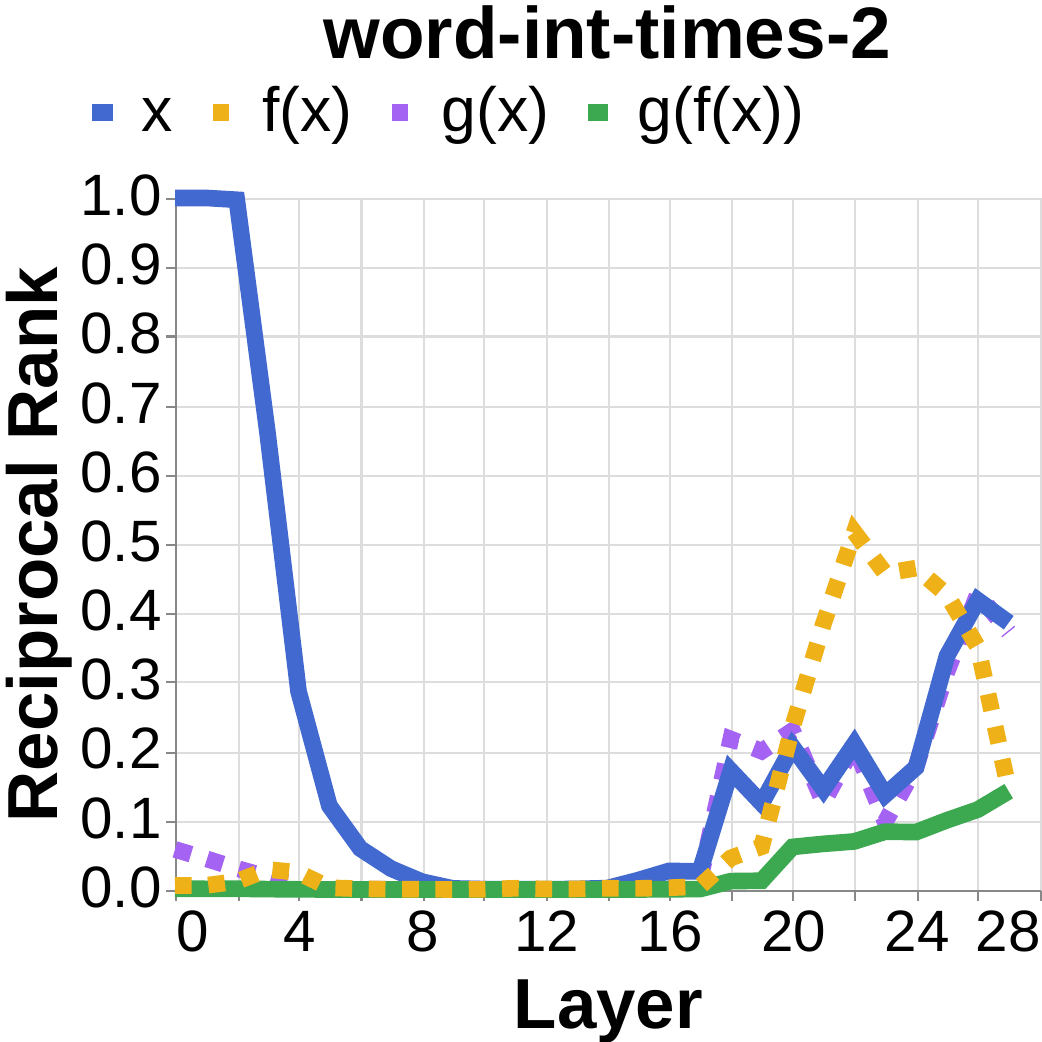} \\
        \includegraphics[width=0.25\linewidth]{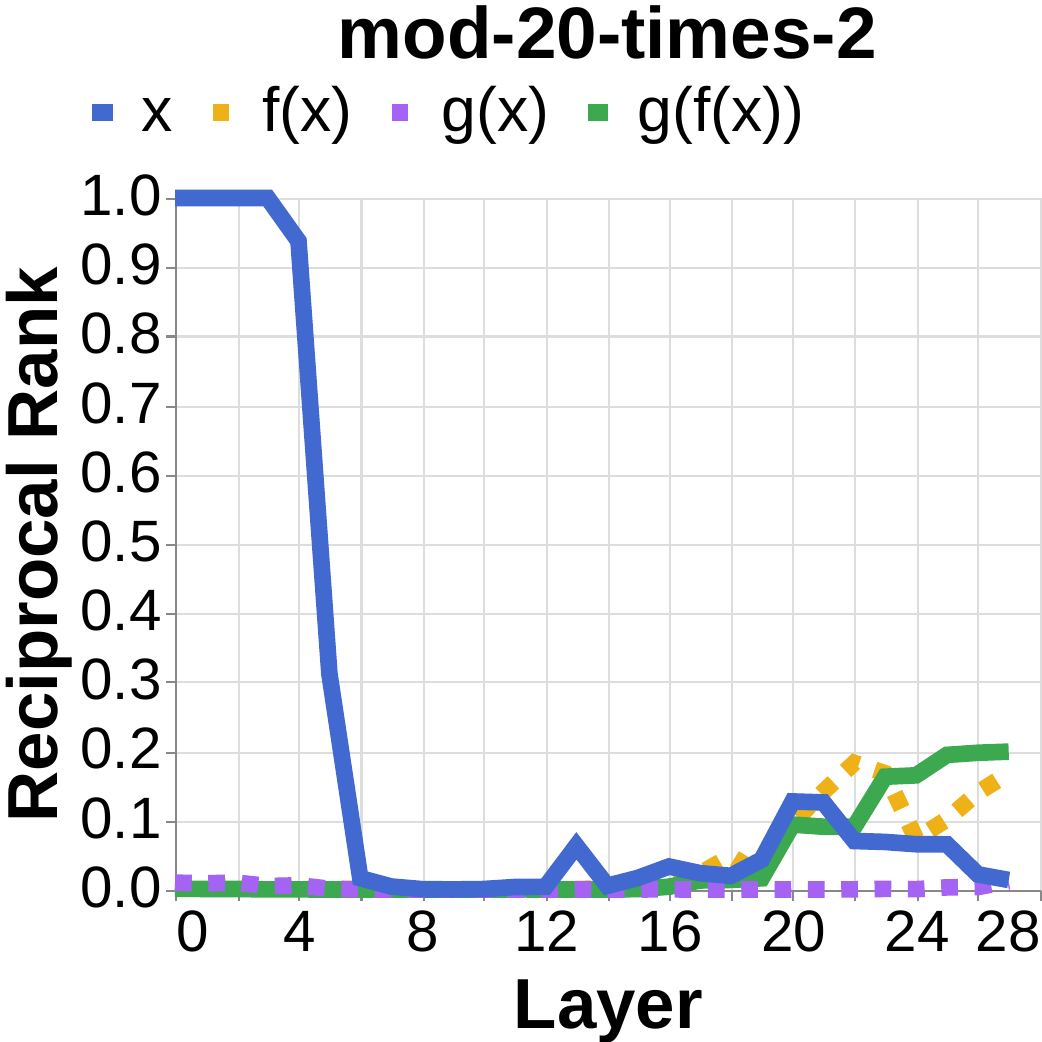}
        & \includegraphics[width=0.25\linewidth]{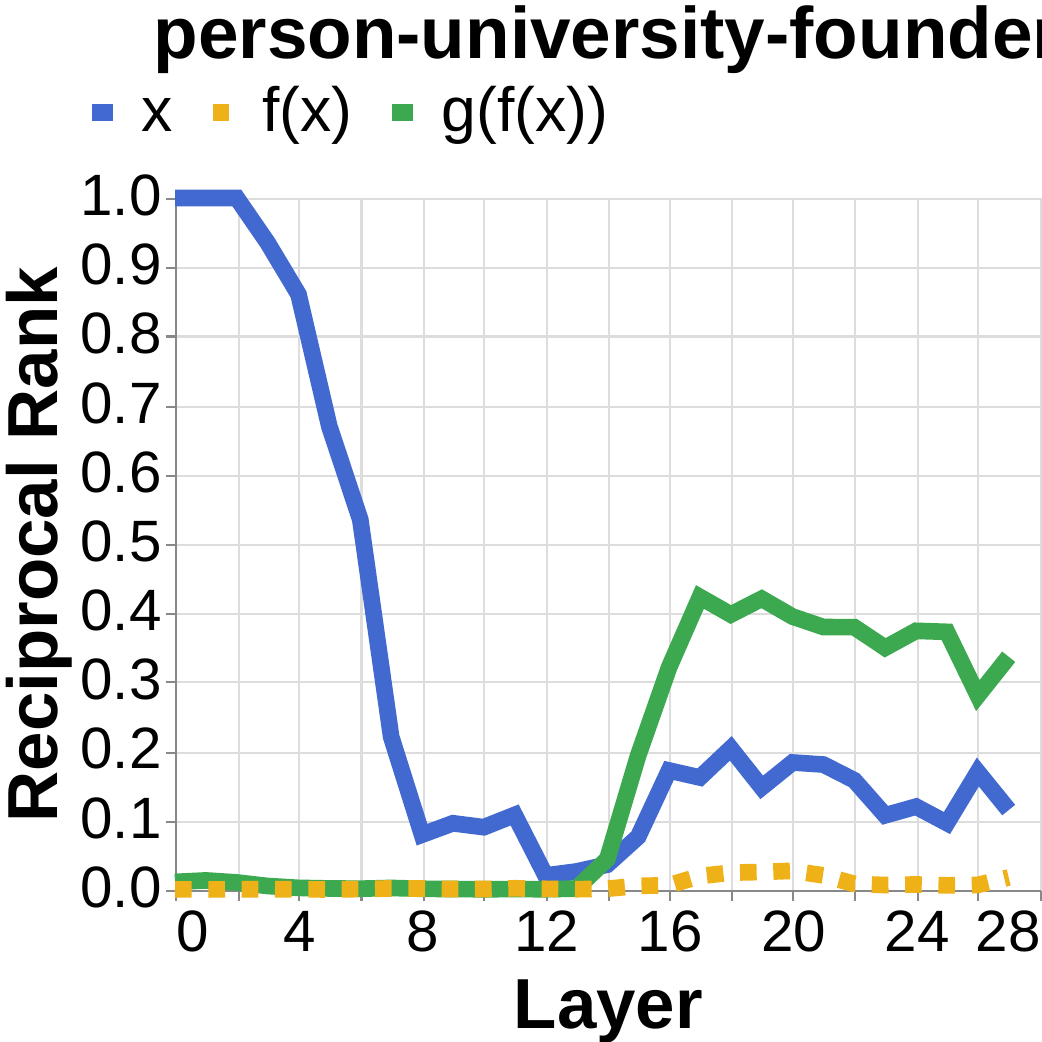}
        & \includegraphics[width=0.25\linewidth]{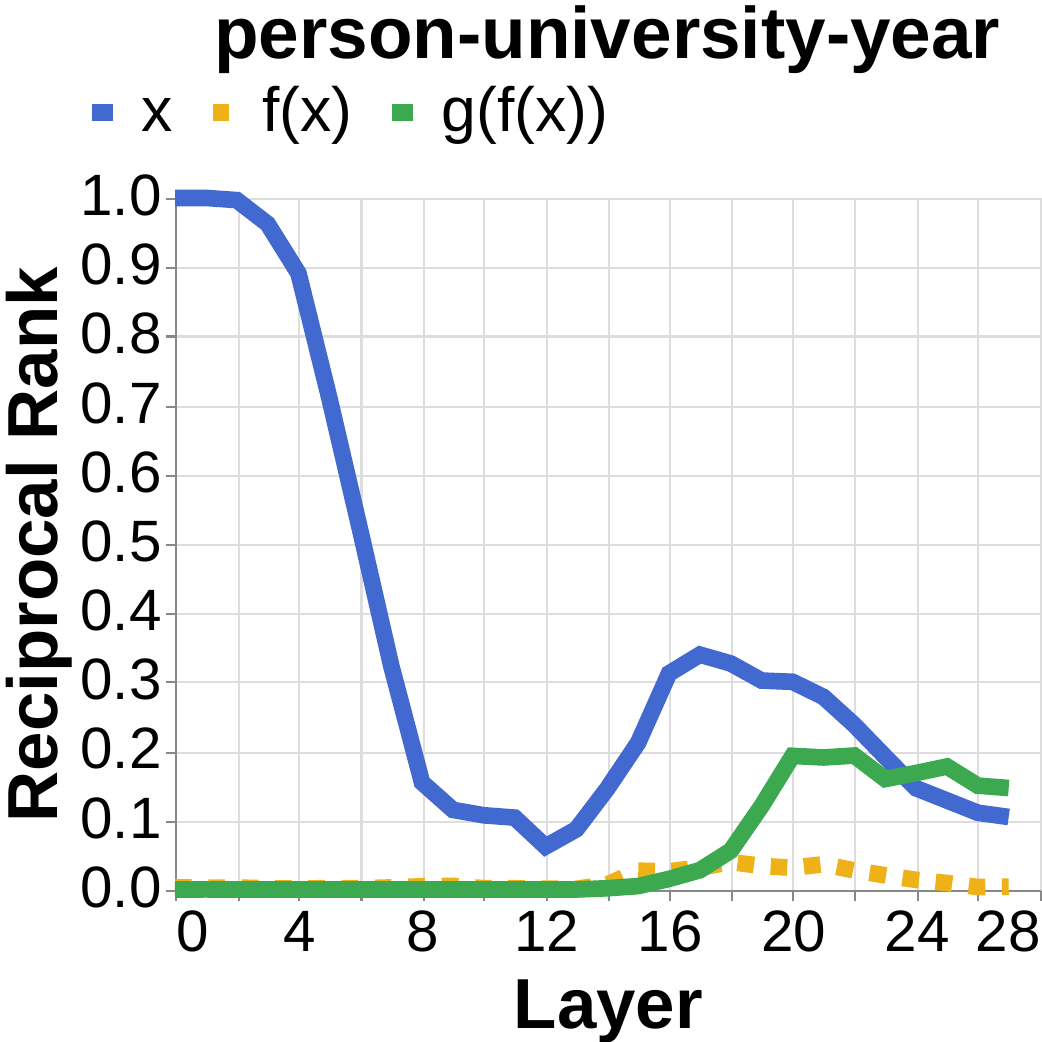}
        & \includegraphics[width=0.25\linewidth]{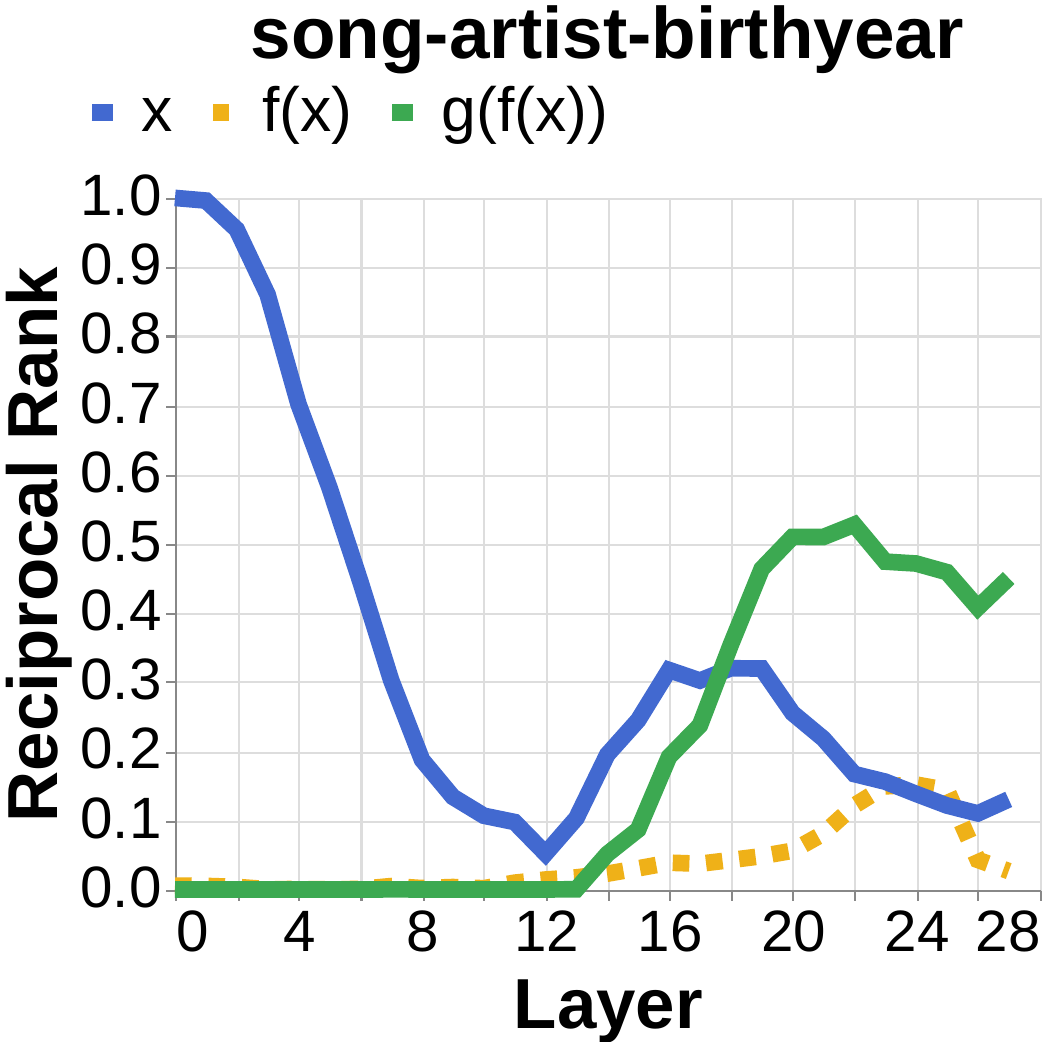} \\
    \end{tabular}
    \end{adjustbox}\end{center}
    \caption{Aggregate processing signatures for each of our tasks, in which Llama 3 (3B) correctly solves all hops but not the composition for at least 10 examples.}
    \label{fig:lens-all-incorrect}
\end{figure}

Although we see a difference in aggregate processing signatures (\cref{fig:lens-correct,fig:lens-incorrect}), where the signal for the intermediate variables is clearer in the correct cases than the incorrect cases, this does not appear to be generally true (and is more likely due to data imbalances). We can see significant presence of the intermediate variables when considering incorrect examples, de-aggregated by task (\cref{fig:lens-all-incorrect}).

\begin{figure}[H]
    \begin{center}
    \begin{tabular}{cccc}
        \includegraphics[width=0.23\linewidth]{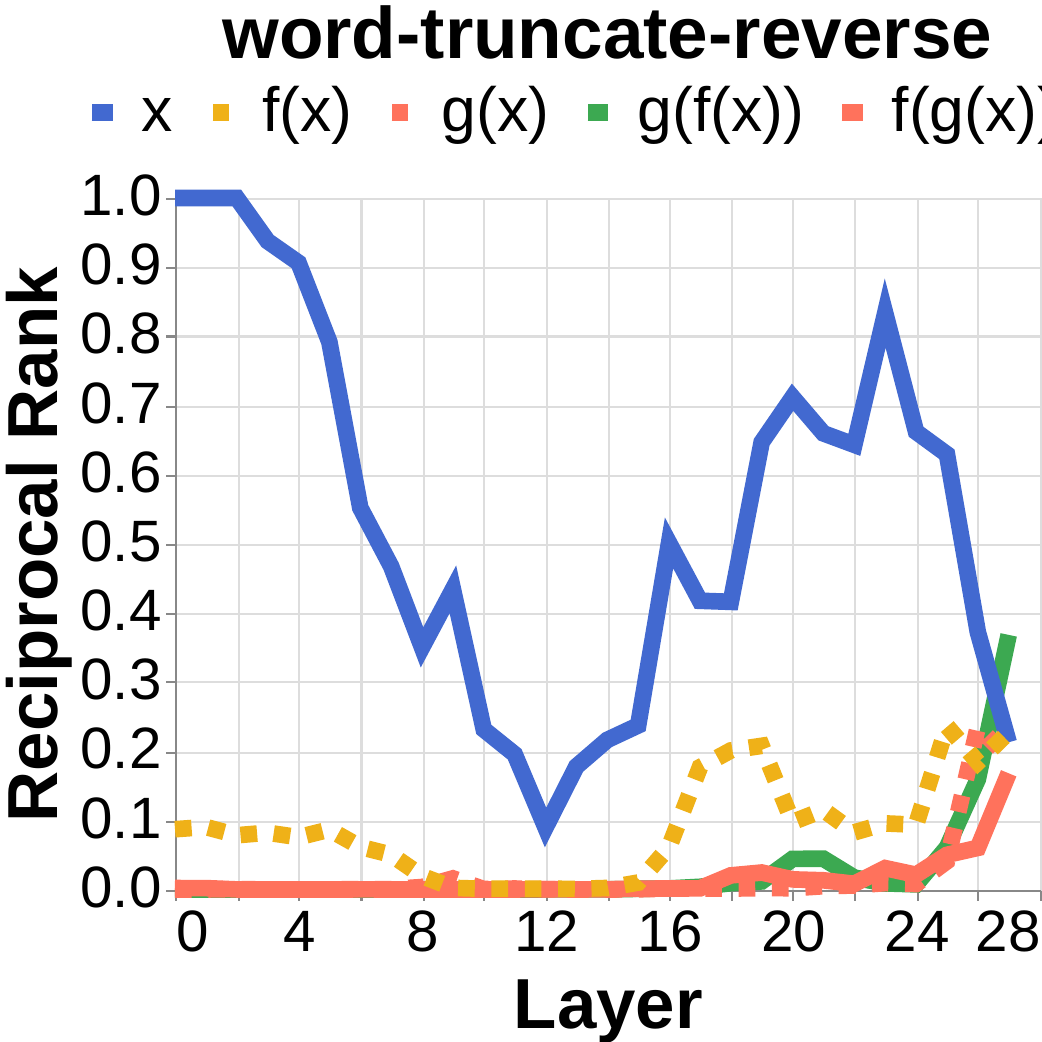} & & & \\
    \end{tabular}
    \end{center}
    \caption{Aggregate processing signatures for each of our tasks, in which Llama 3 (3B) correctly solves all hops but not the composition for less than 10 examples.}
    \label{fig:lens-excluded-incorrect}
\end{figure}

\section{Token Identity Patchscope}\label{app:token-identity}

Here, we repeat the analyses in \cref{sec:lens,sec:linearity}, but use the token identity patchscope \citep{ghandeharioun2024:patchscopes} instead of logit lens. This patchscope is specifically designed (e.g. by prompt, patching strategy) to outperform logit lens at decoding the next token prediction and involves more of the model's computation in doing so. We should be aware that this design objective is not perfectly aligned with our purpose (decoding intermediate variables), but still offers a way to explore features stored in an alternative representational subspace.

We would specifically like to use this method to decode a representation into vocabulary-space logits. To do so, we prompt a model with the "token identity prompt", in which random tokens are repeated twice each, such as "\texttt{[A] [A] ; [B] [B] ; ... ; [?]}". We patch our representation of interest into the residual stream of this forward pass (at the corresponding layer and final token position). The language modeling logits resulting from our intervention then serve as the decoding for our representation.

We generally find similarities with our logit lens analyses: in tasks with "compositional" processing signatures, we continue to see growth of the signals for the intermediate variables with or before that for $g(f(x))$. Please zoom in to observe simultaneous growth, which may be difficult to see due to overlapping lines. And, although these plots may show growth of $f(x)$ and $g(f(x))$ in the same layers, recall that these computations can occur in different (e.g. earlier or later) residual streams (\cref{app:implementation}).

\begin{figure}[H]
    \begin{center}\begin{adjustbox}{width=\linewidth}
    \begin{tabular}{cccc}
        \includegraphics[width=0.25\linewidth]{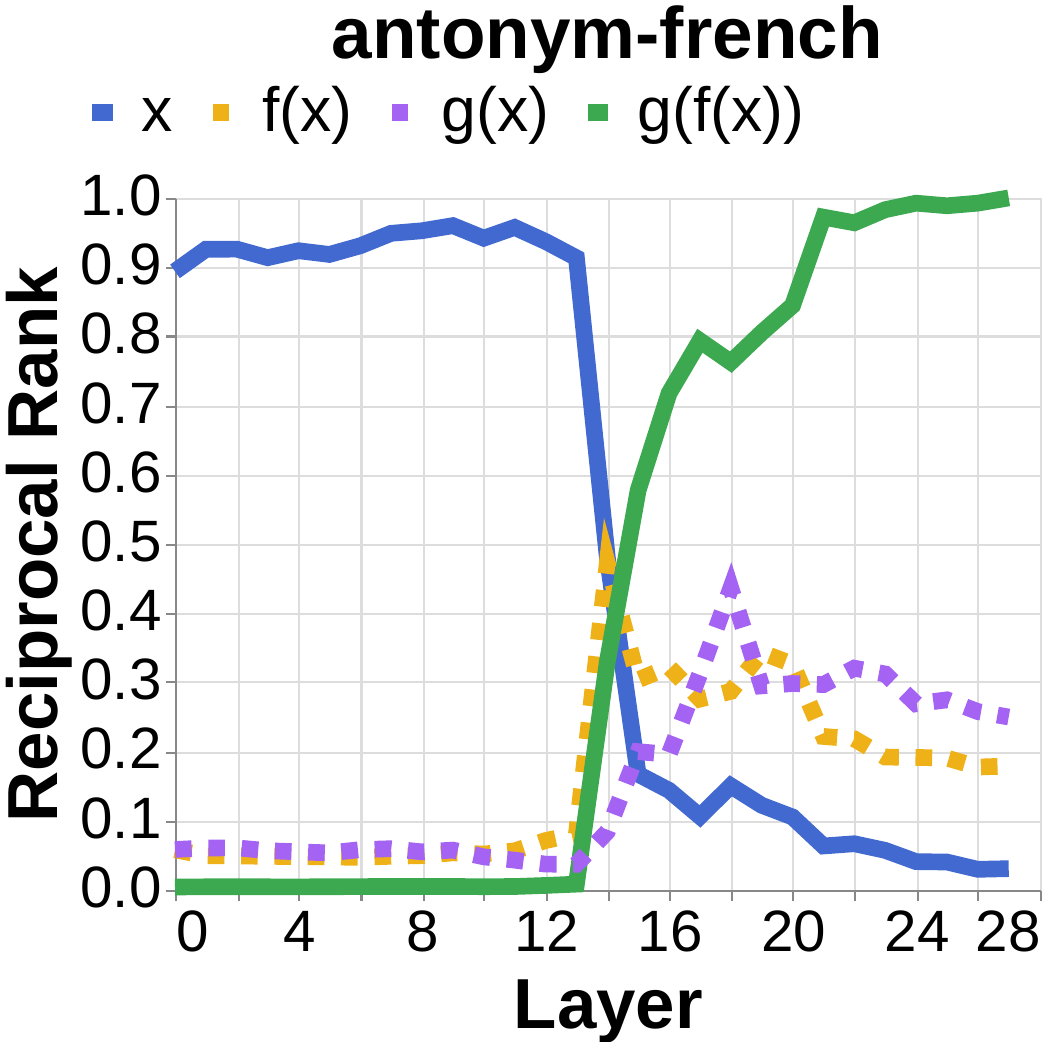}
        & \includegraphics[width=0.25\linewidth]{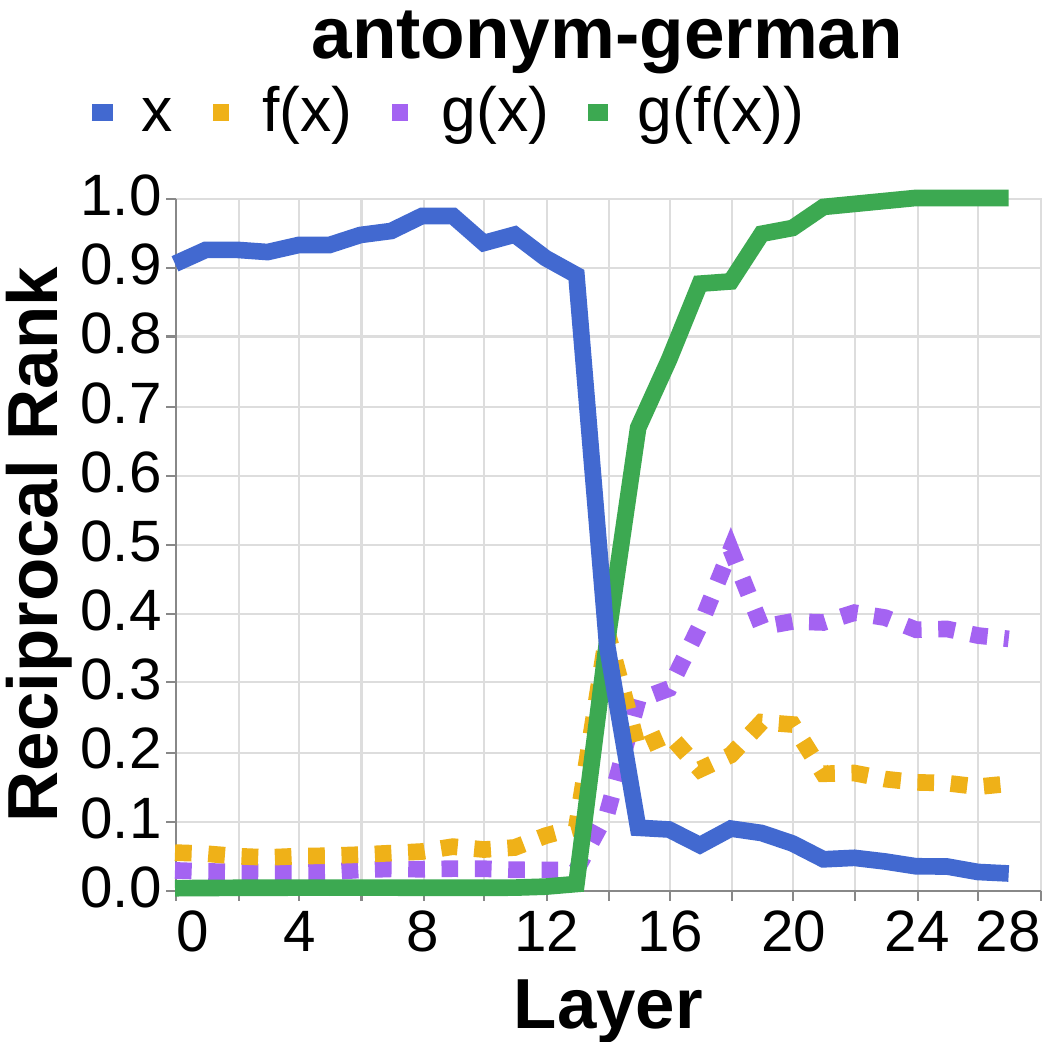}
        & \includegraphics[width=0.25\linewidth]{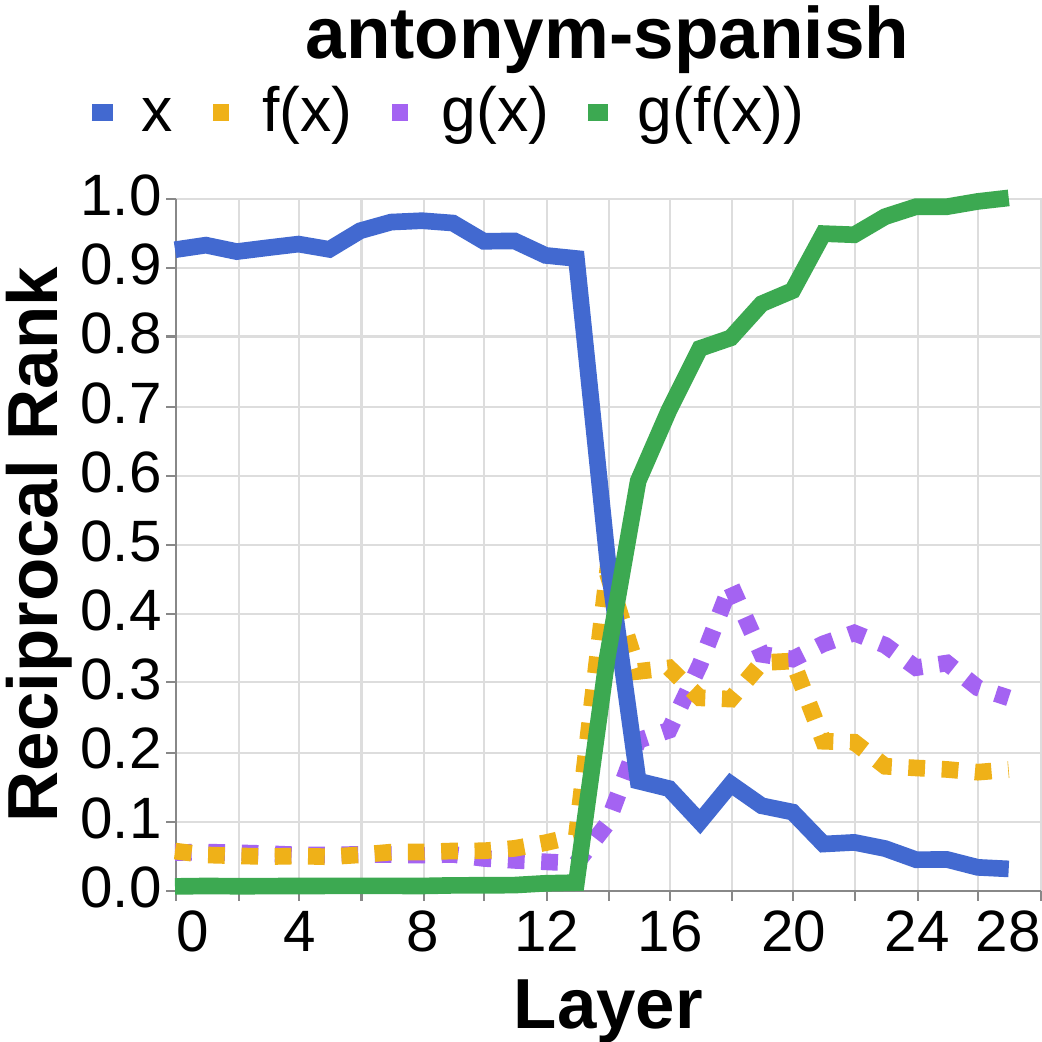}
        & \includegraphics[width=0.25\linewidth]{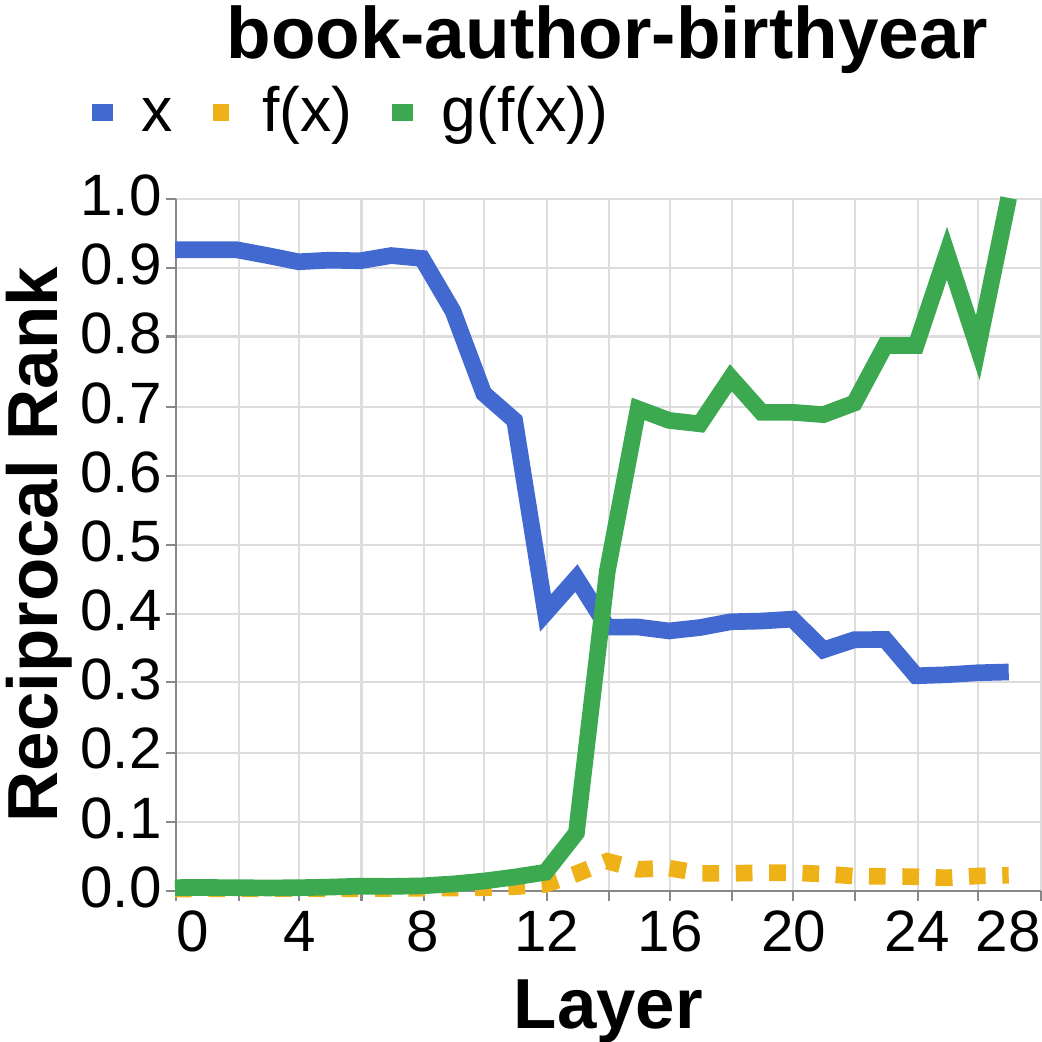} \\
        \includegraphics[width=0.25\linewidth]{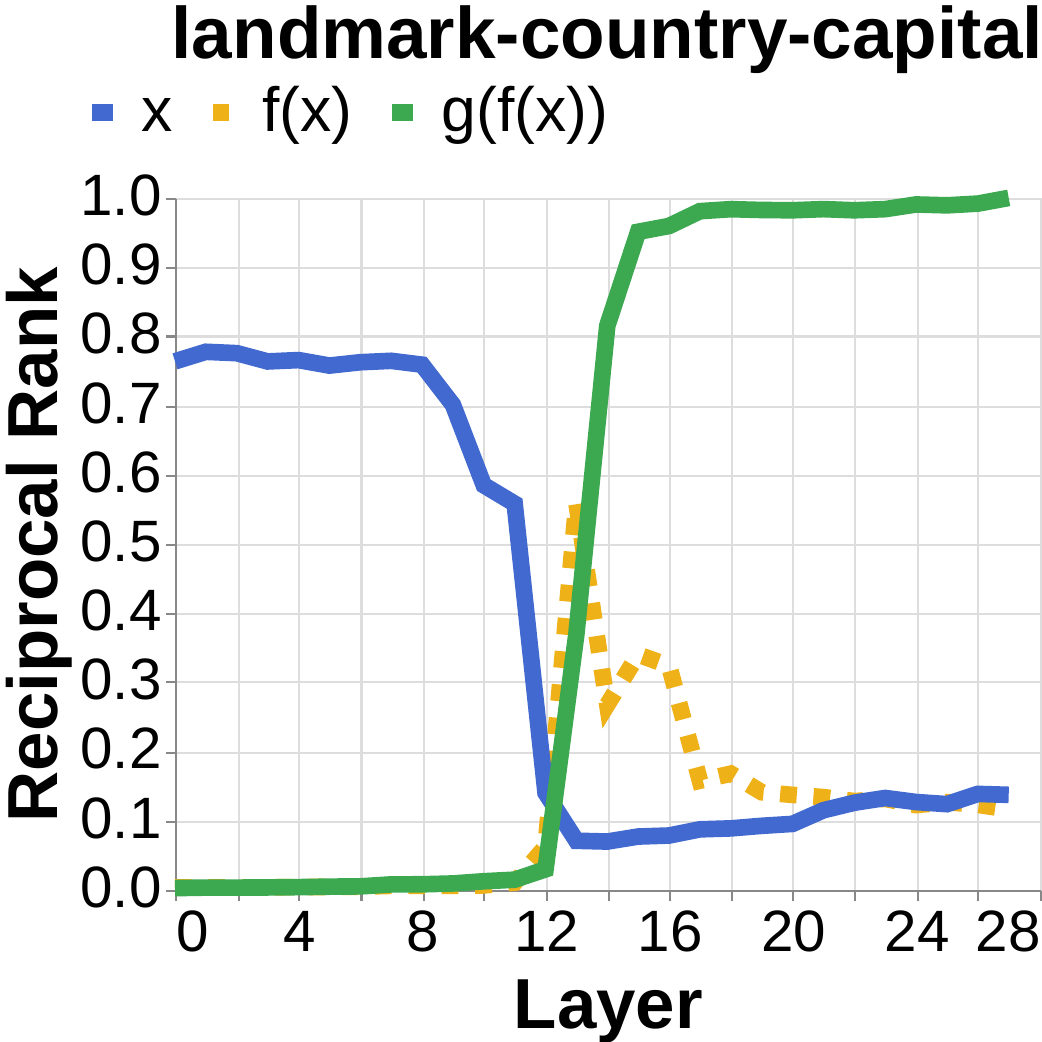}
        & \includegraphics[width=0.25\linewidth]{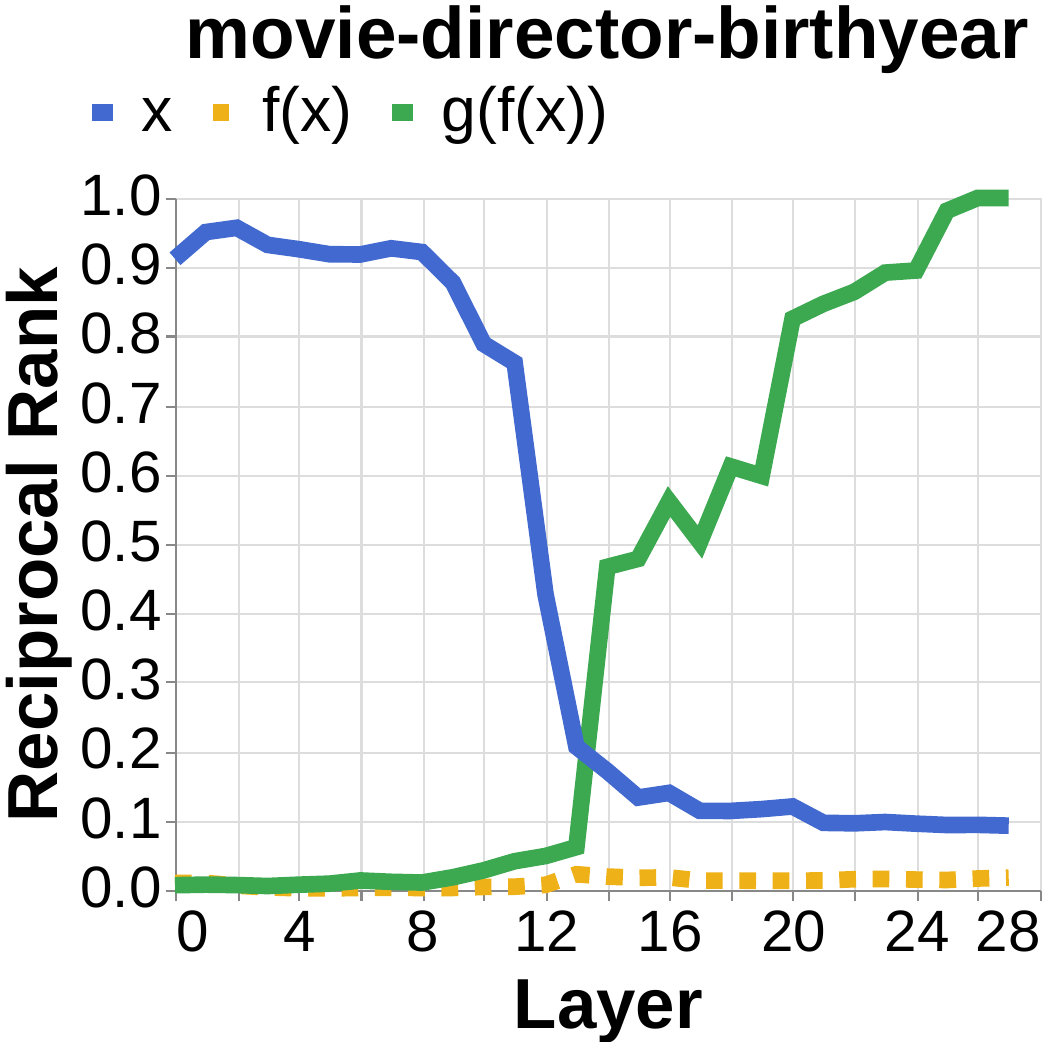}
        & \includegraphics[width=0.25\linewidth]{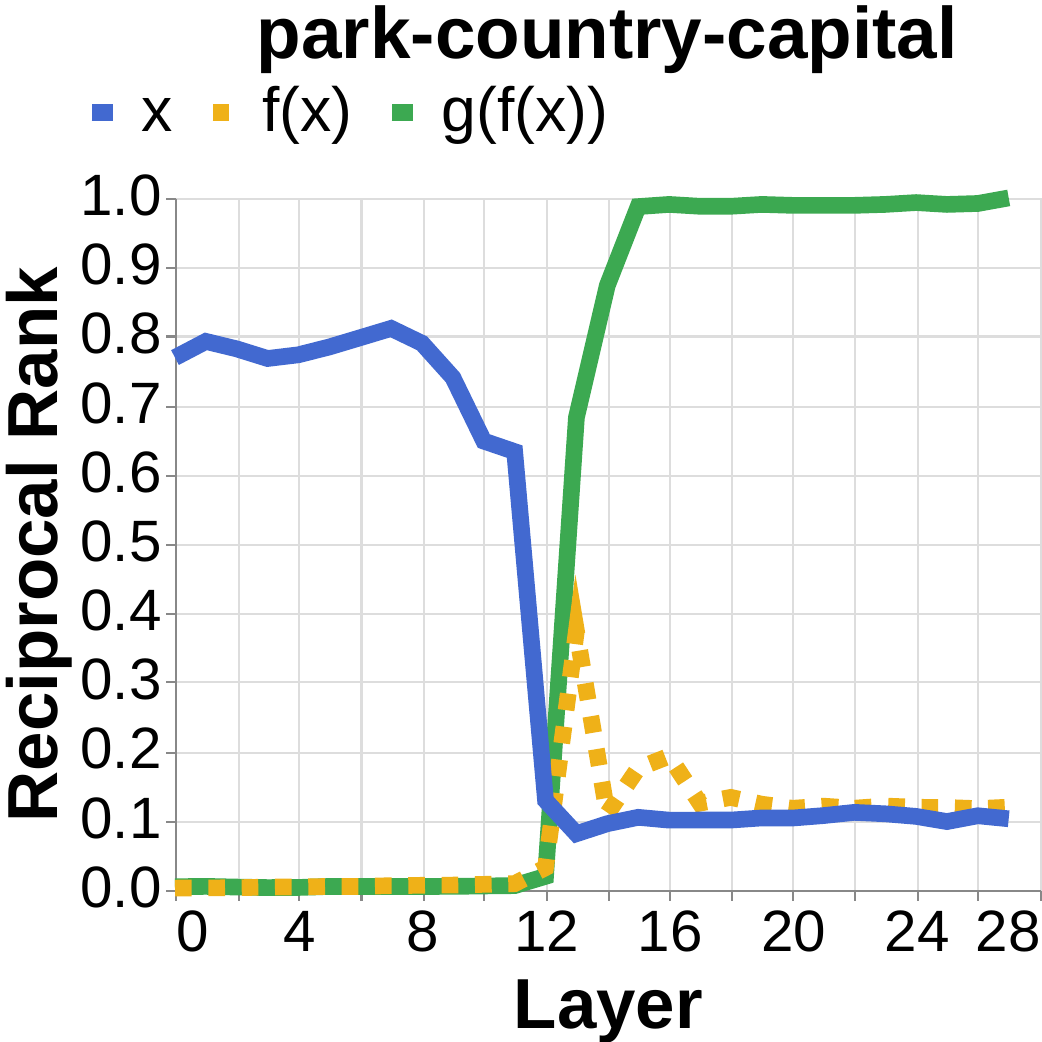}
        & \includegraphics[width=0.25\linewidth]{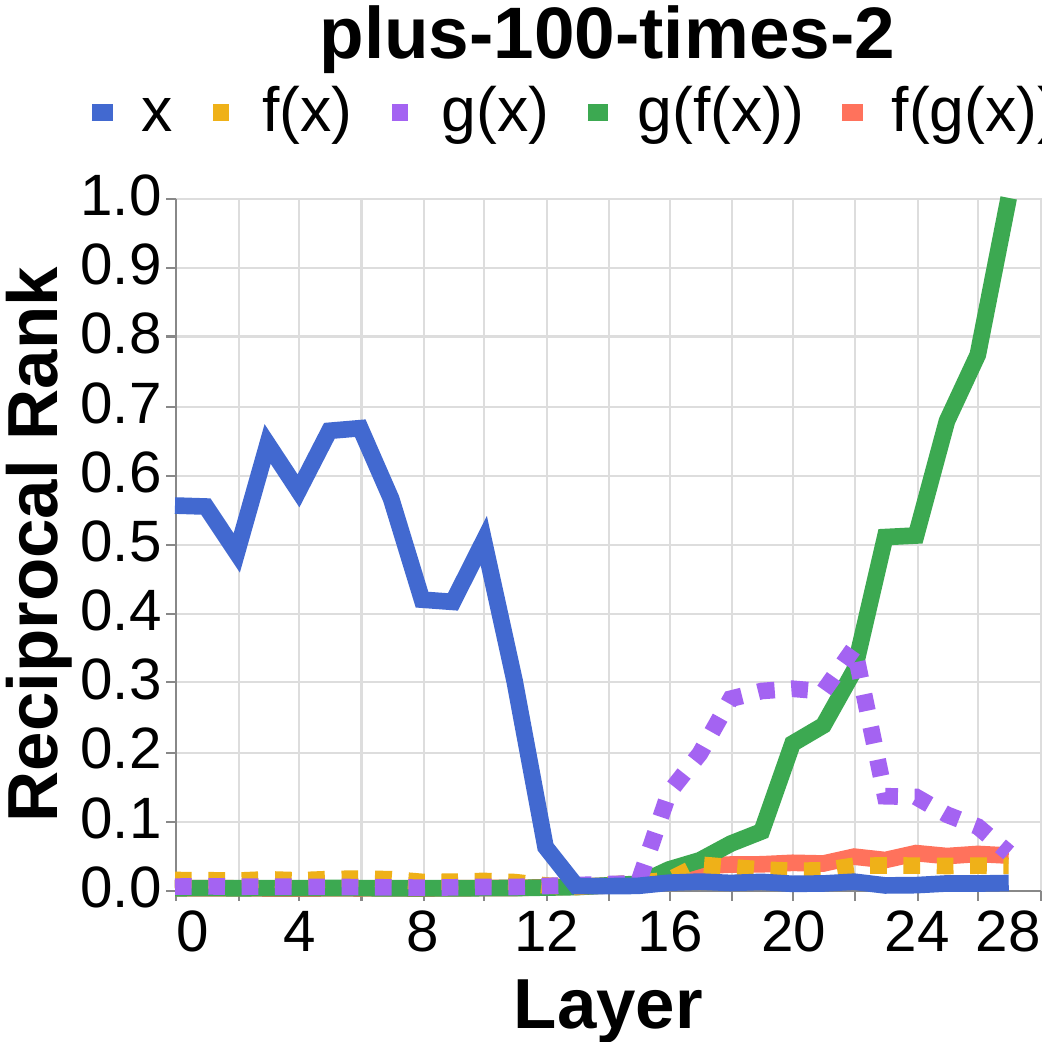} \\
        \includegraphics[width=0.25\linewidth]{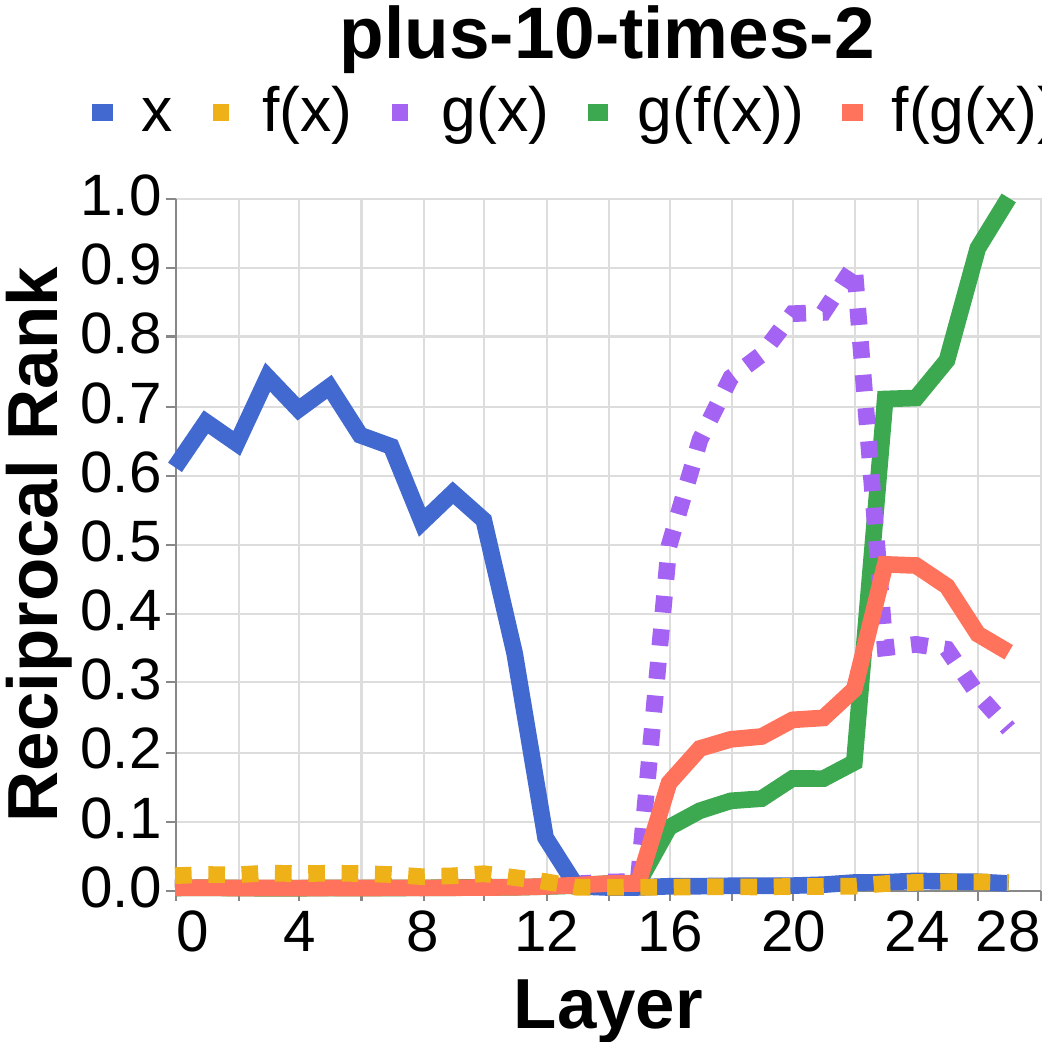}
        & \includegraphics[width=0.25\linewidth]{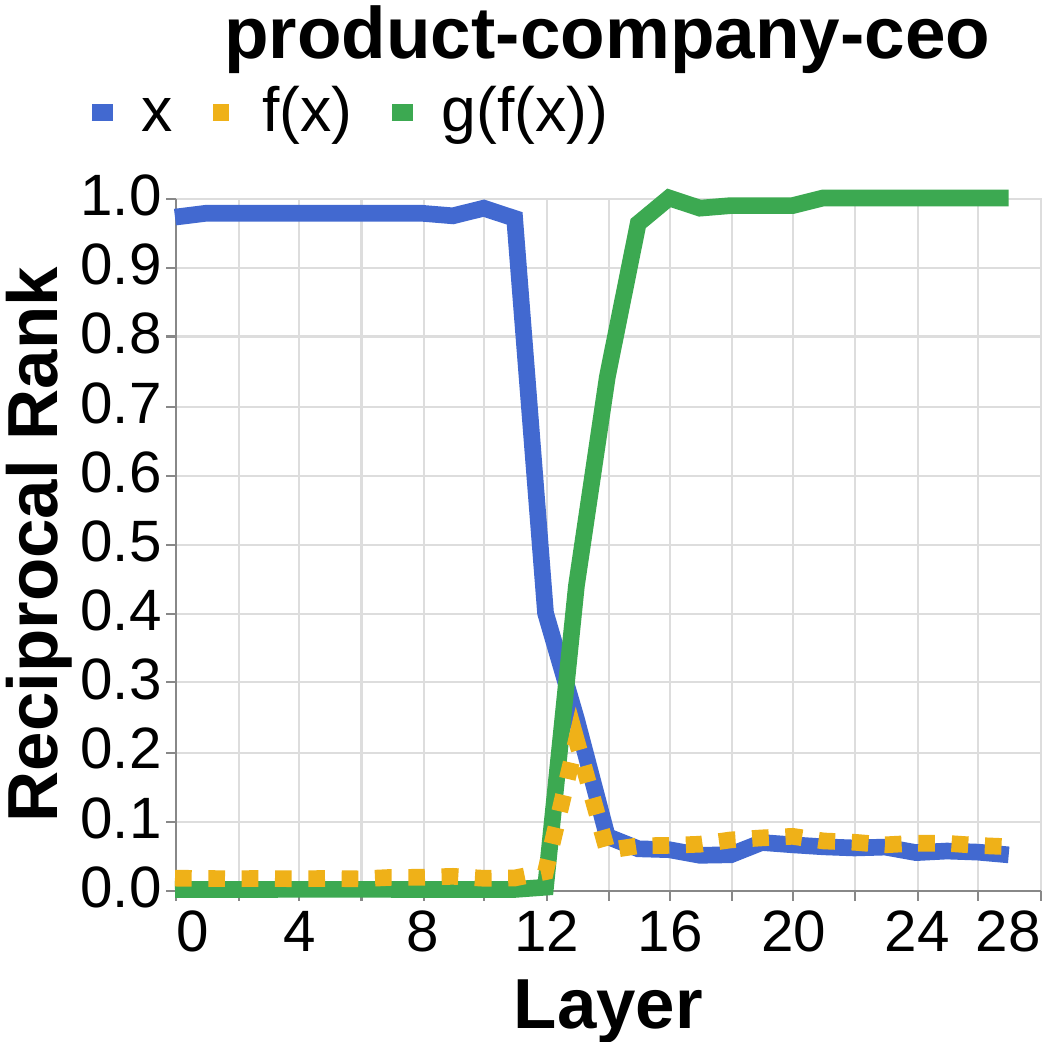}
        & \includegraphics[width=0.25\linewidth]{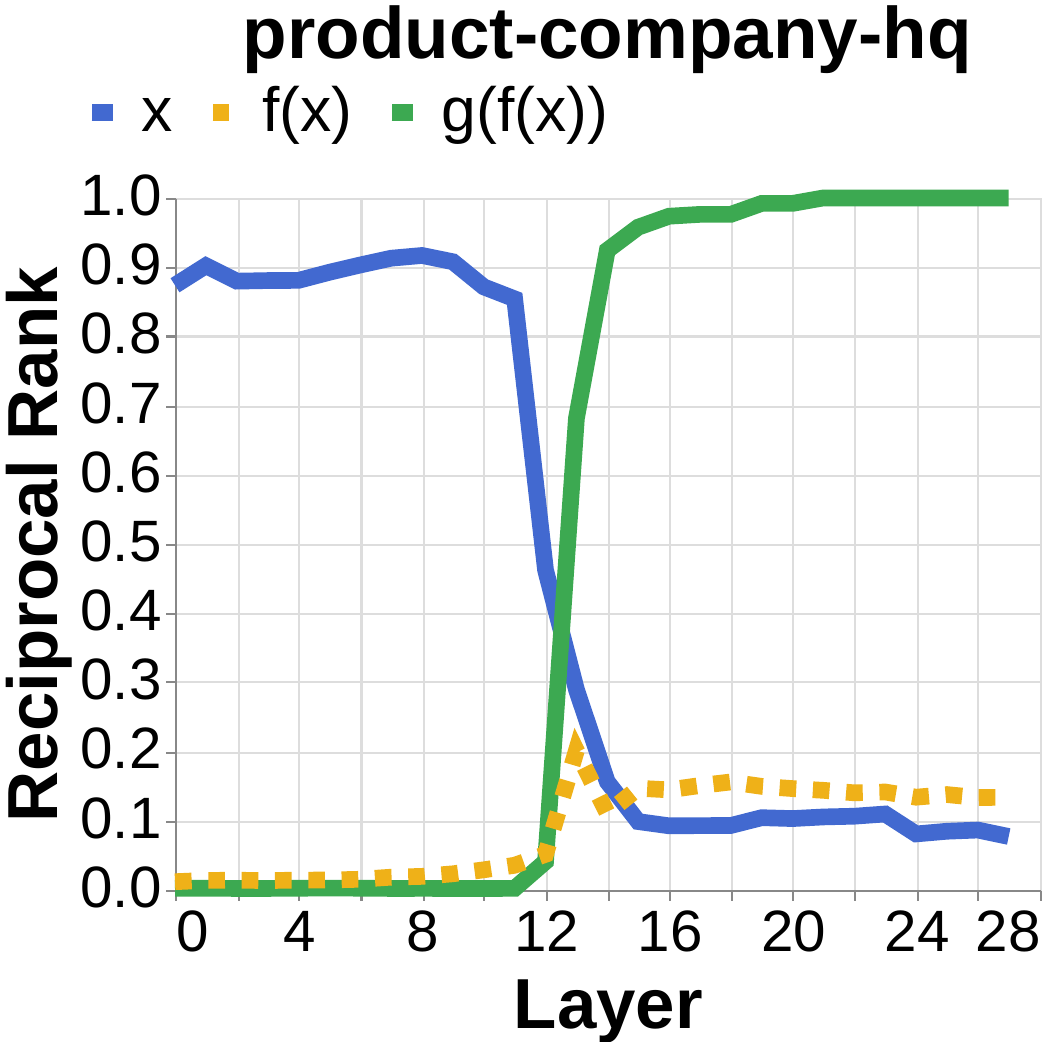}
        & \includegraphics[width=0.25\linewidth]{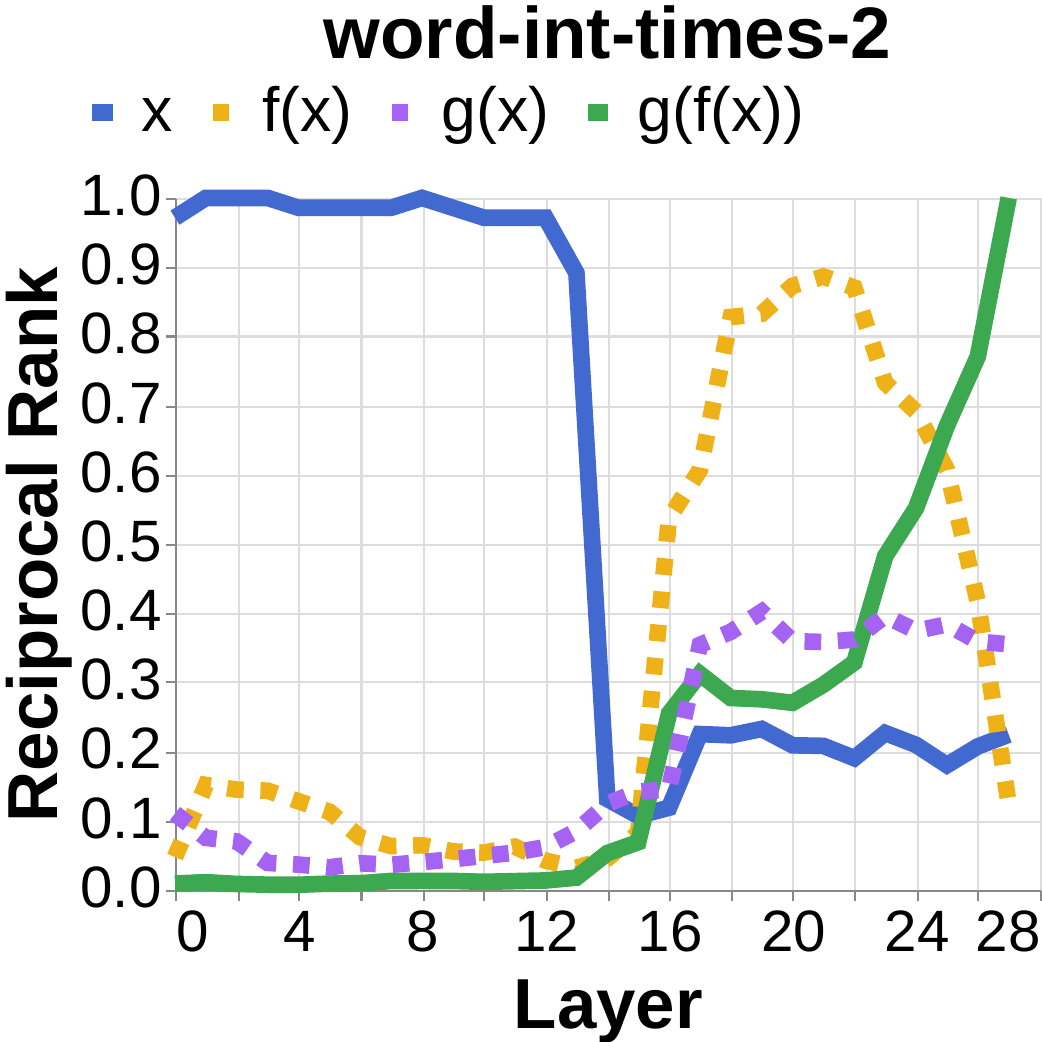} \\
    \end{tabular}
    \end{adjustbox}\end{center}
    \caption{Aggregate processing signatures (using the token identity patchscope) for each of our tasks, in which Llama 3 (3B) correctly solves all hops and the composition for at least 10 examples.}
\end{figure}

\begin{figure}[H]
    \begin{center}\begin{adjustbox}{width=\linewidth}
    \begin{tabular}{cccc}
        \includegraphics[width=0.25\linewidth]{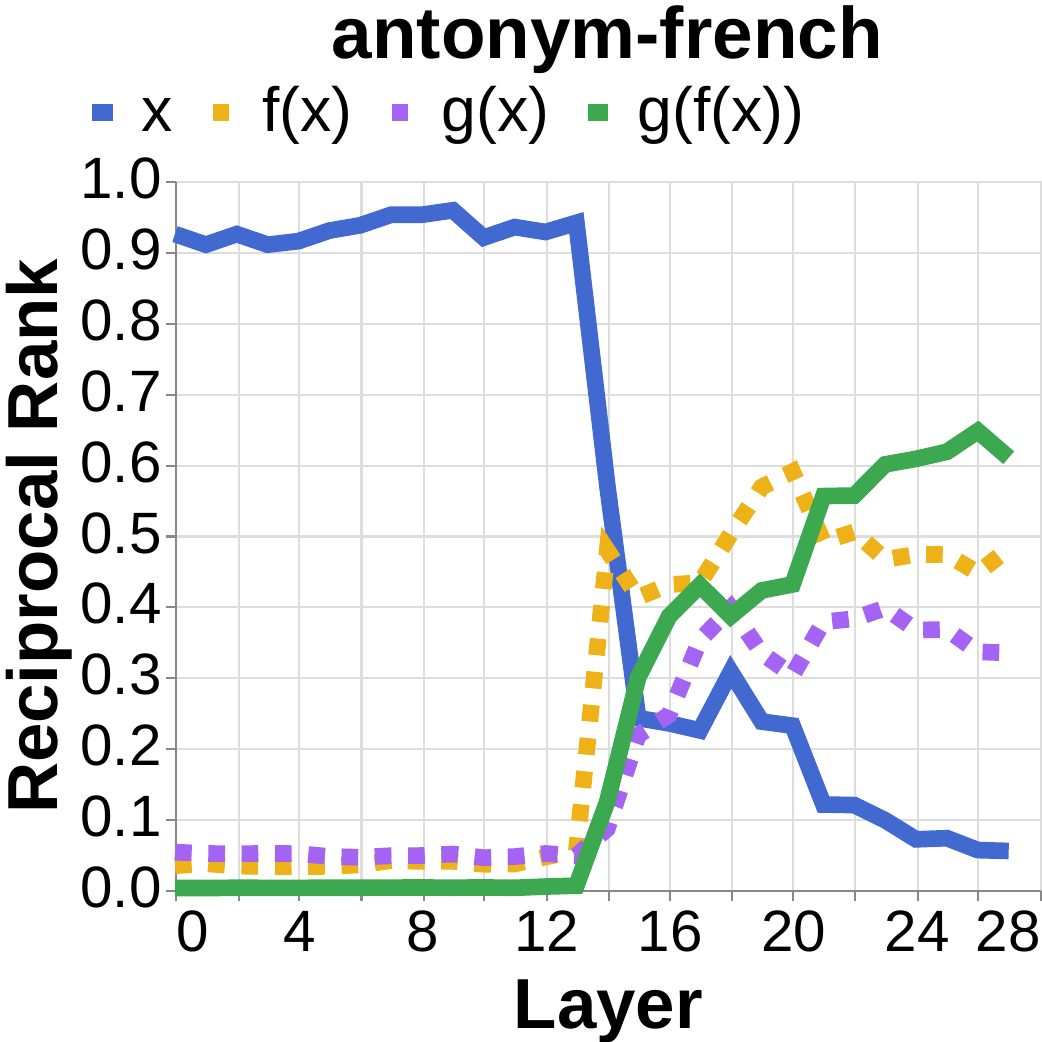}
        & \includegraphics[width=0.25\linewidth]{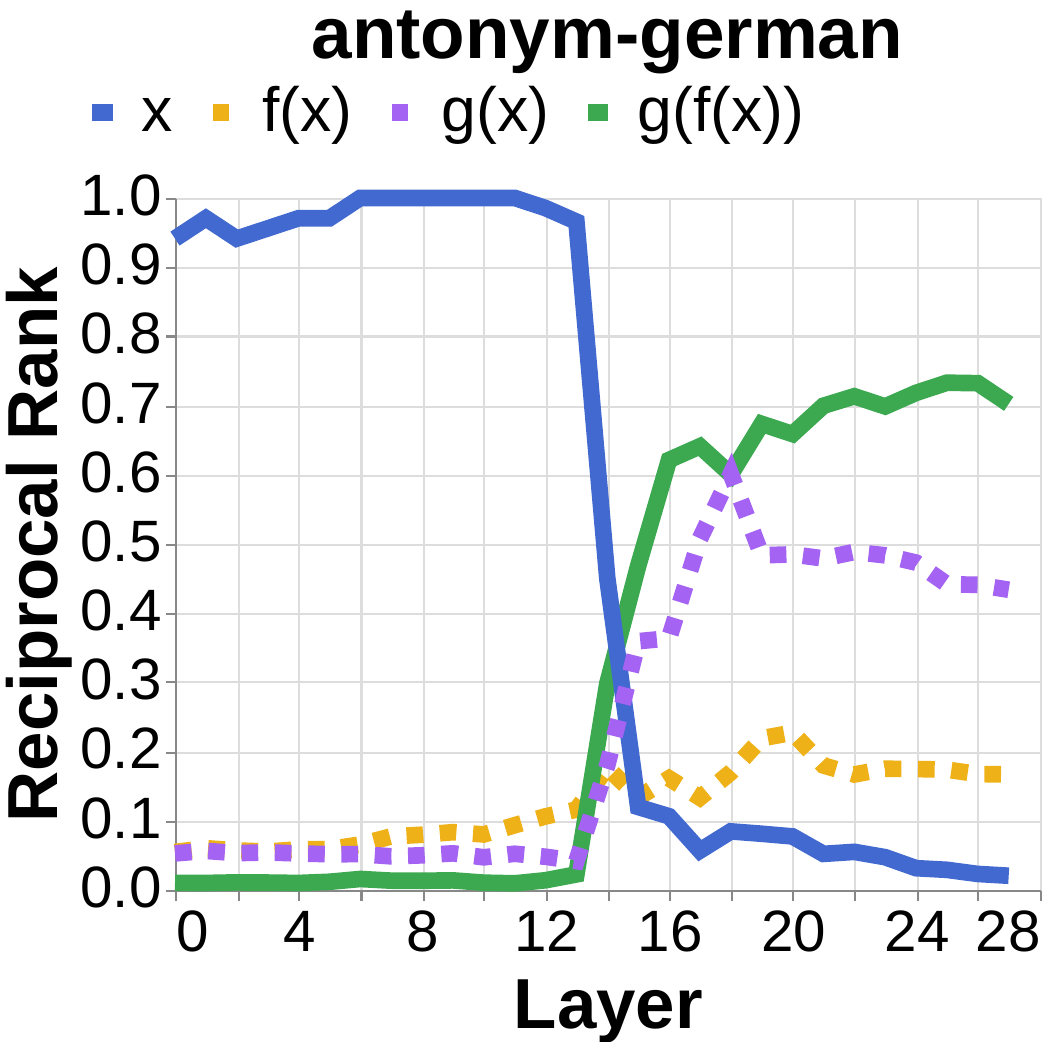}
        & \includegraphics[width=0.25\linewidth]{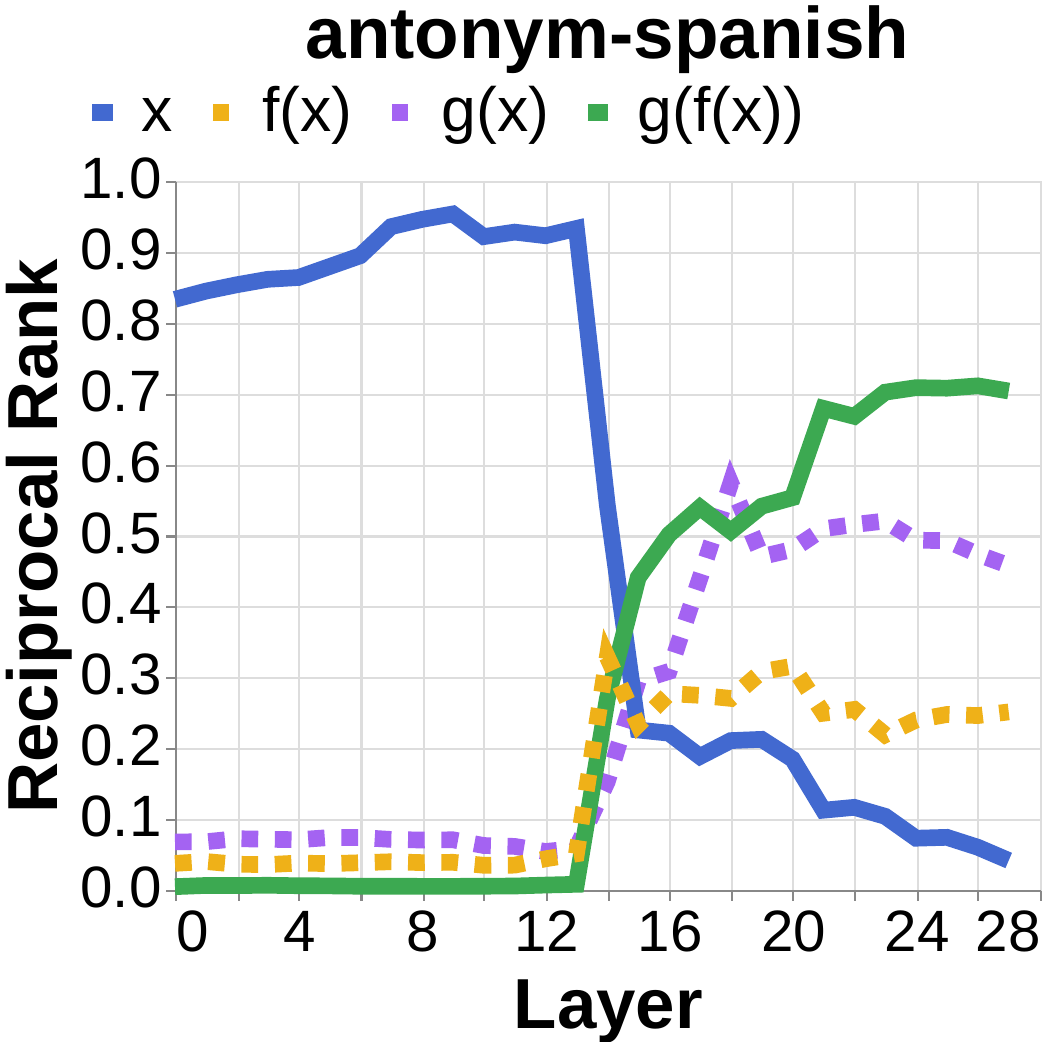}
        & \includegraphics[width=0.25\linewidth]{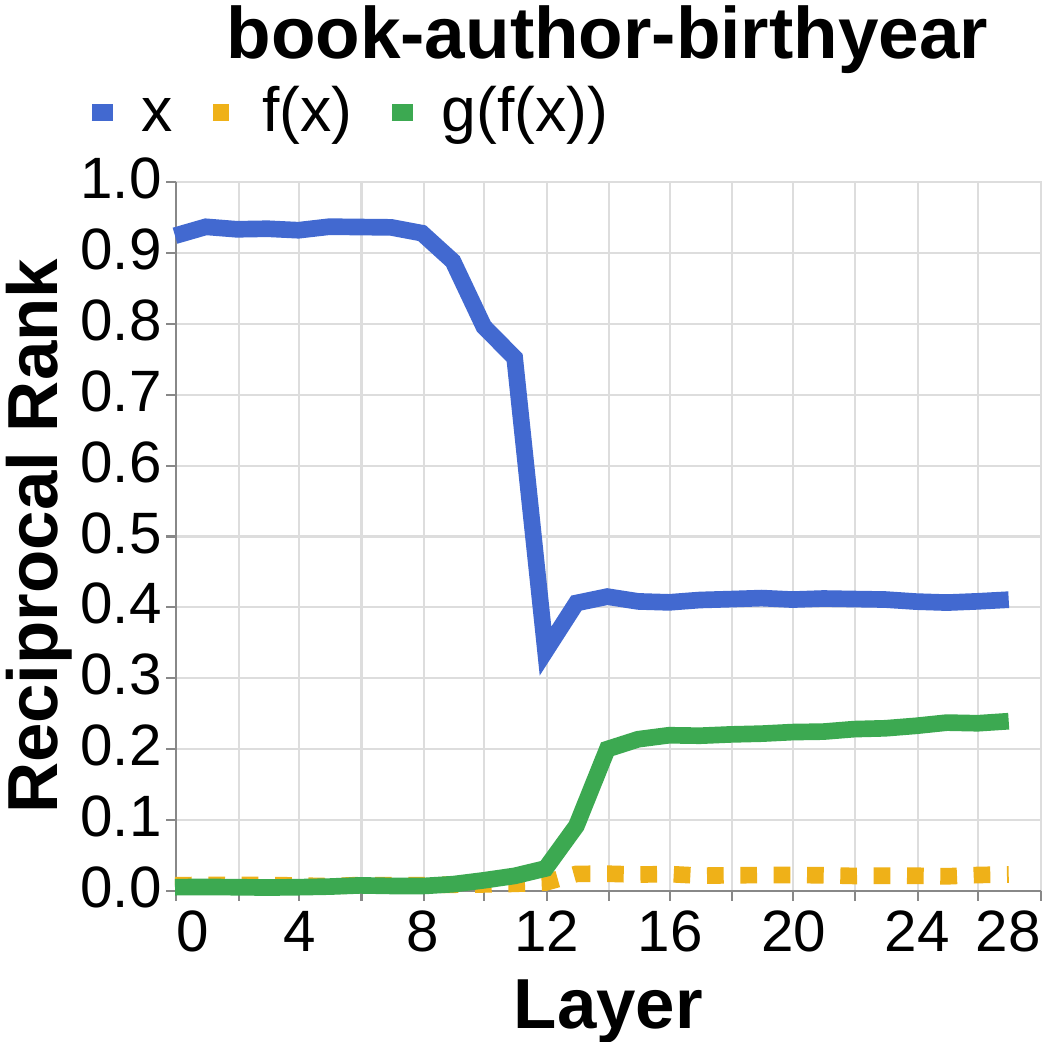} \\
        \includegraphics[width=0.25\linewidth]{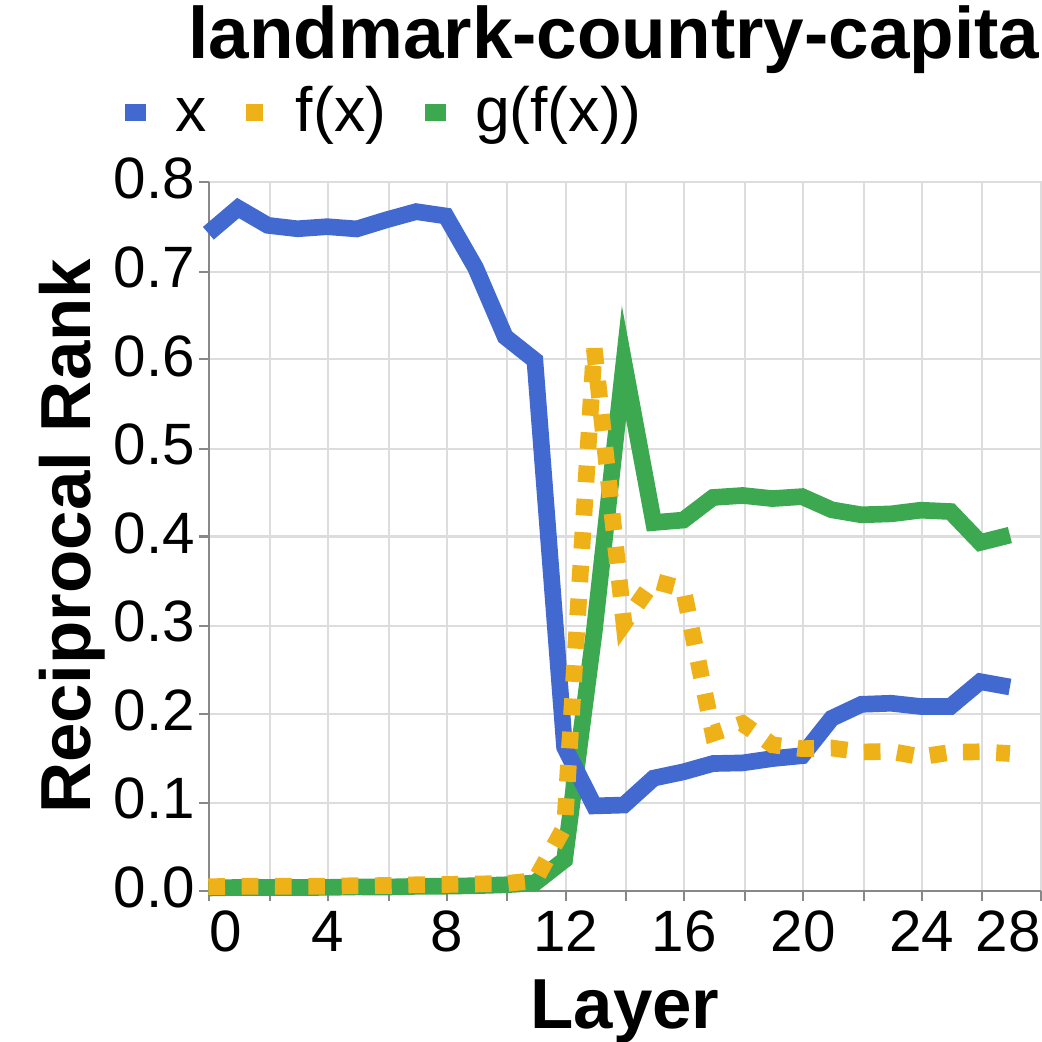}
        & \includegraphics[width=0.25\linewidth]{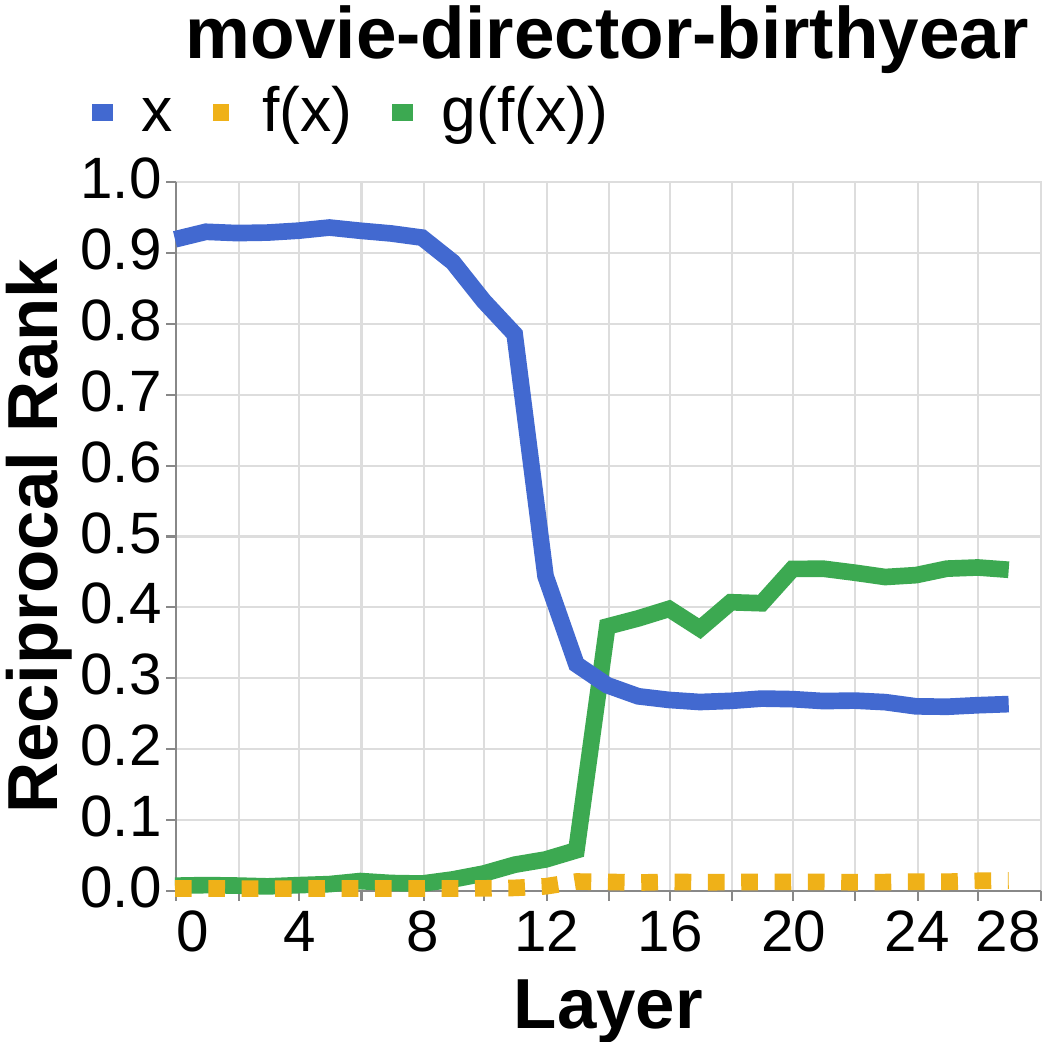}
        & \includegraphics[width=0.25\linewidth]{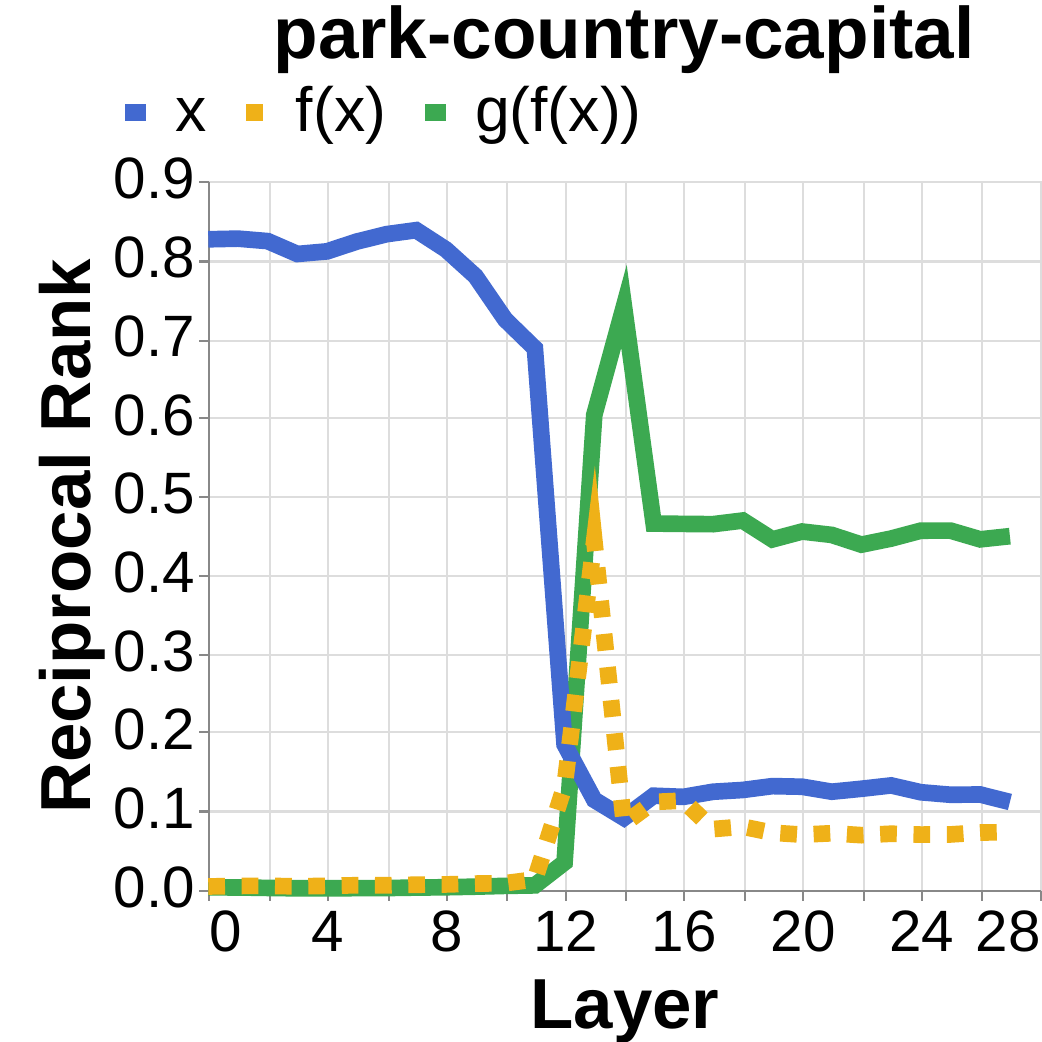}
        & \includegraphics[width=0.25\linewidth]{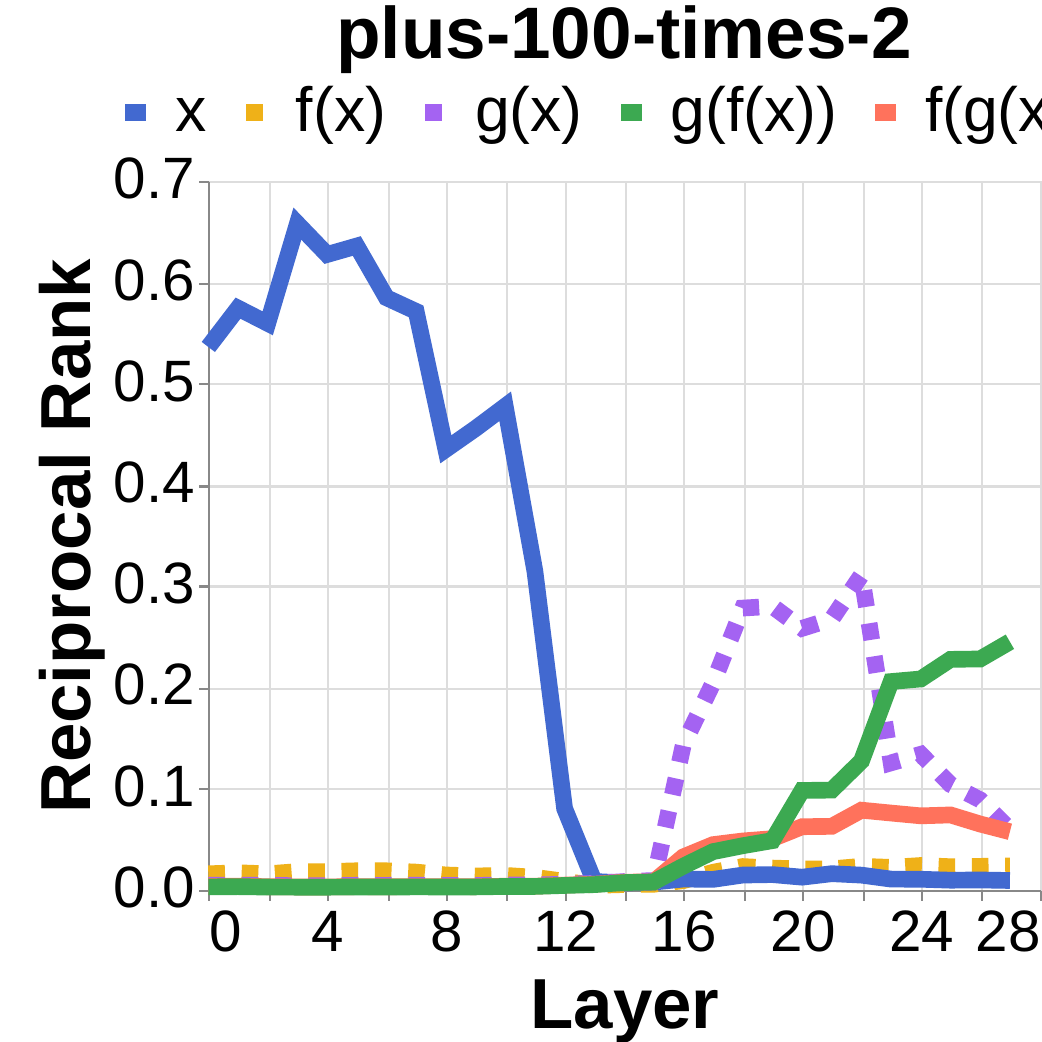} \\
        \includegraphics[width=0.25\linewidth]{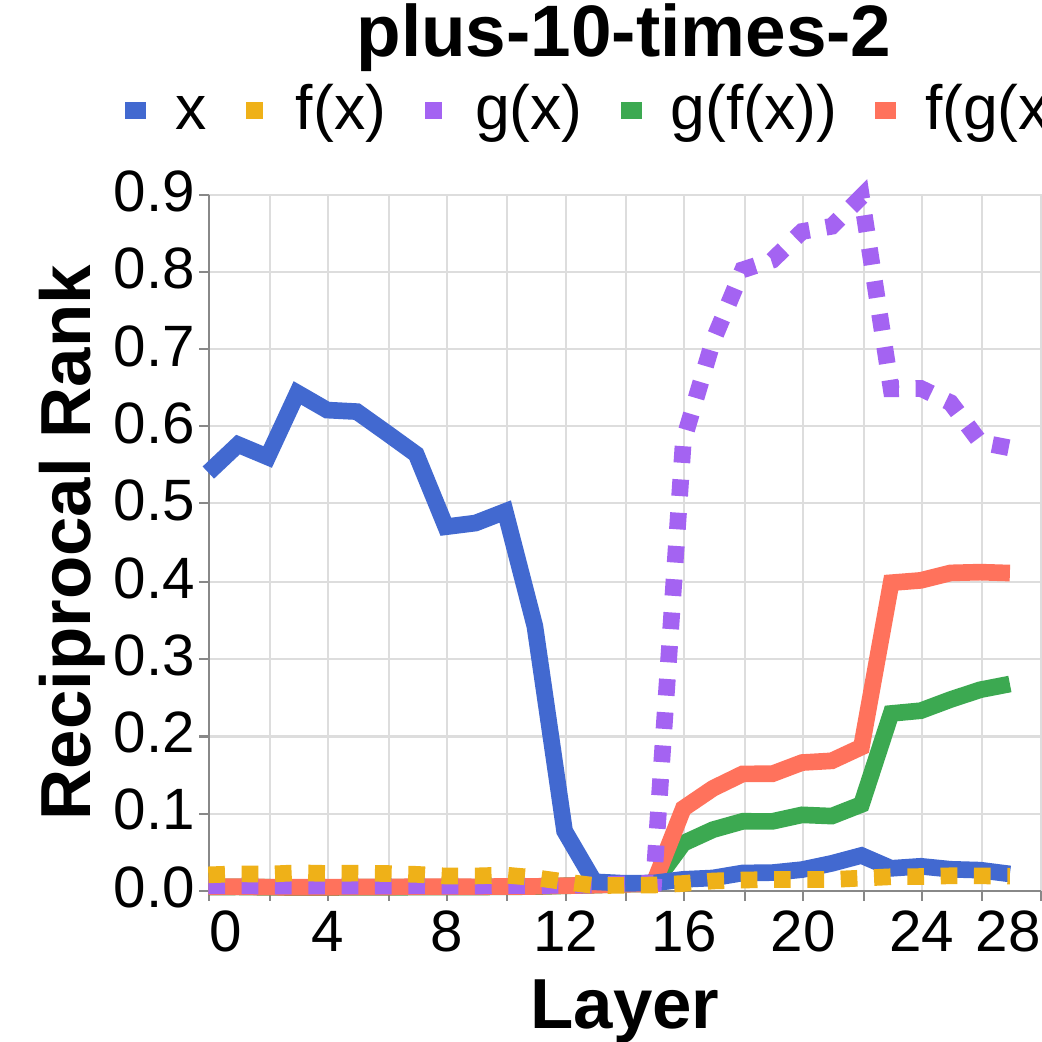}
        & \includegraphics[width=0.25\linewidth]{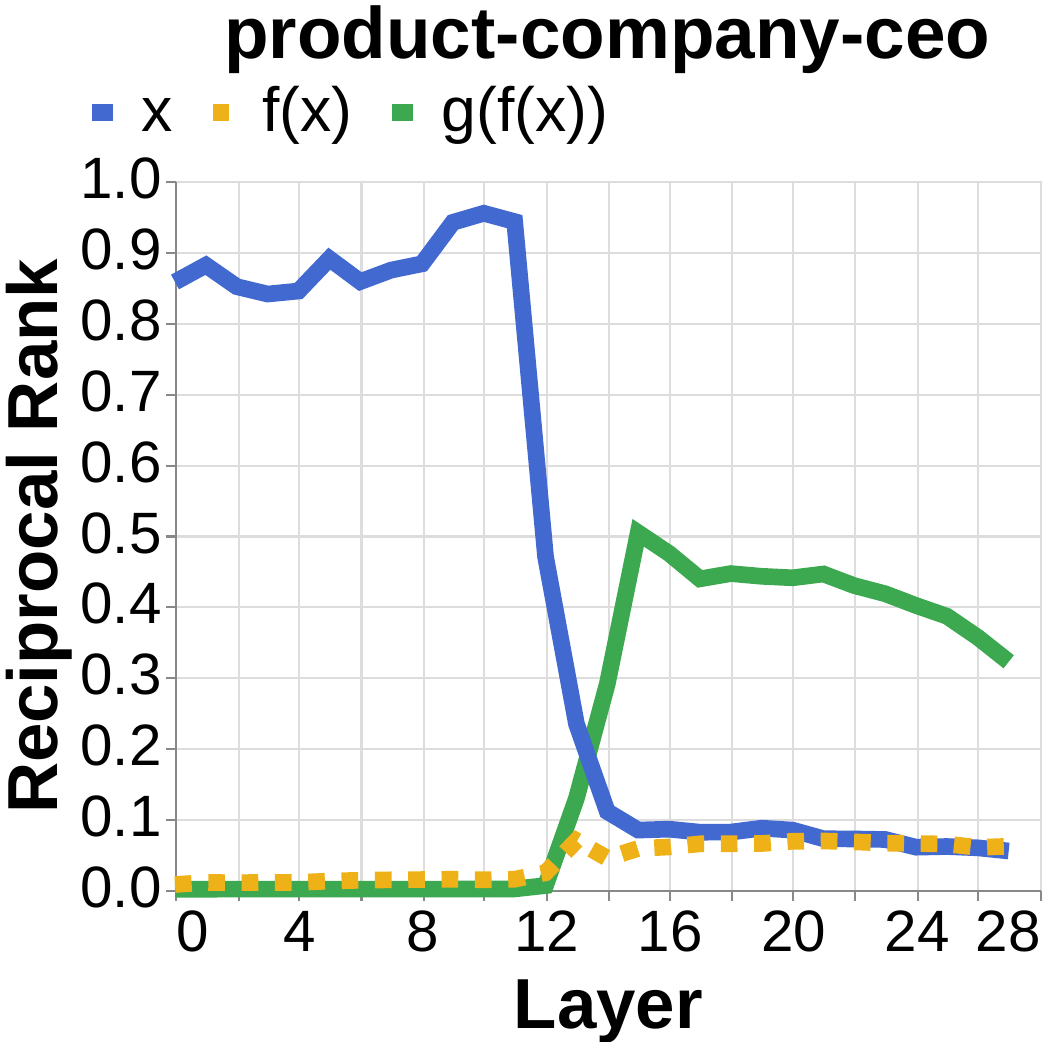}
        & \includegraphics[width=0.25\linewidth]{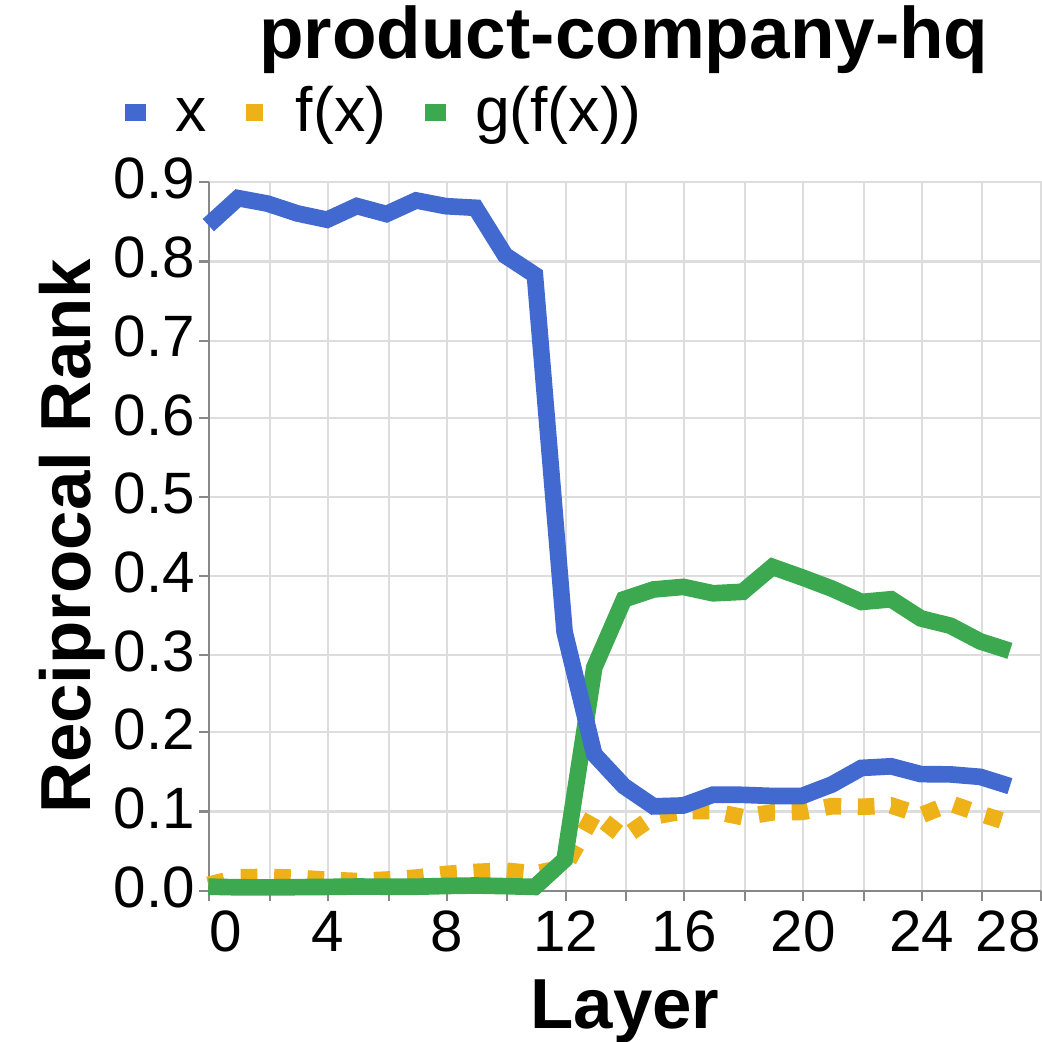}
        & \includegraphics[width=0.25\linewidth]{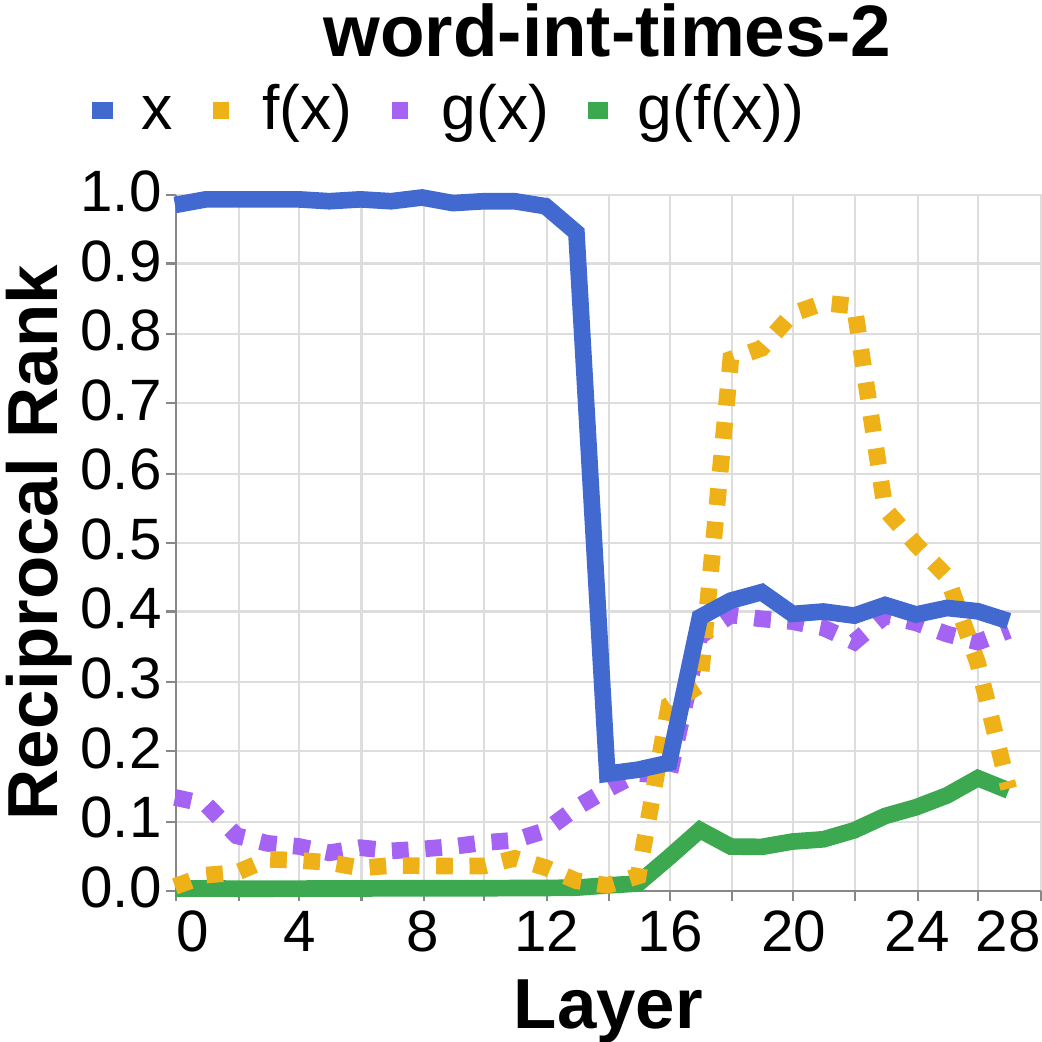} \\
        \includegraphics[width=0.25\linewidth]{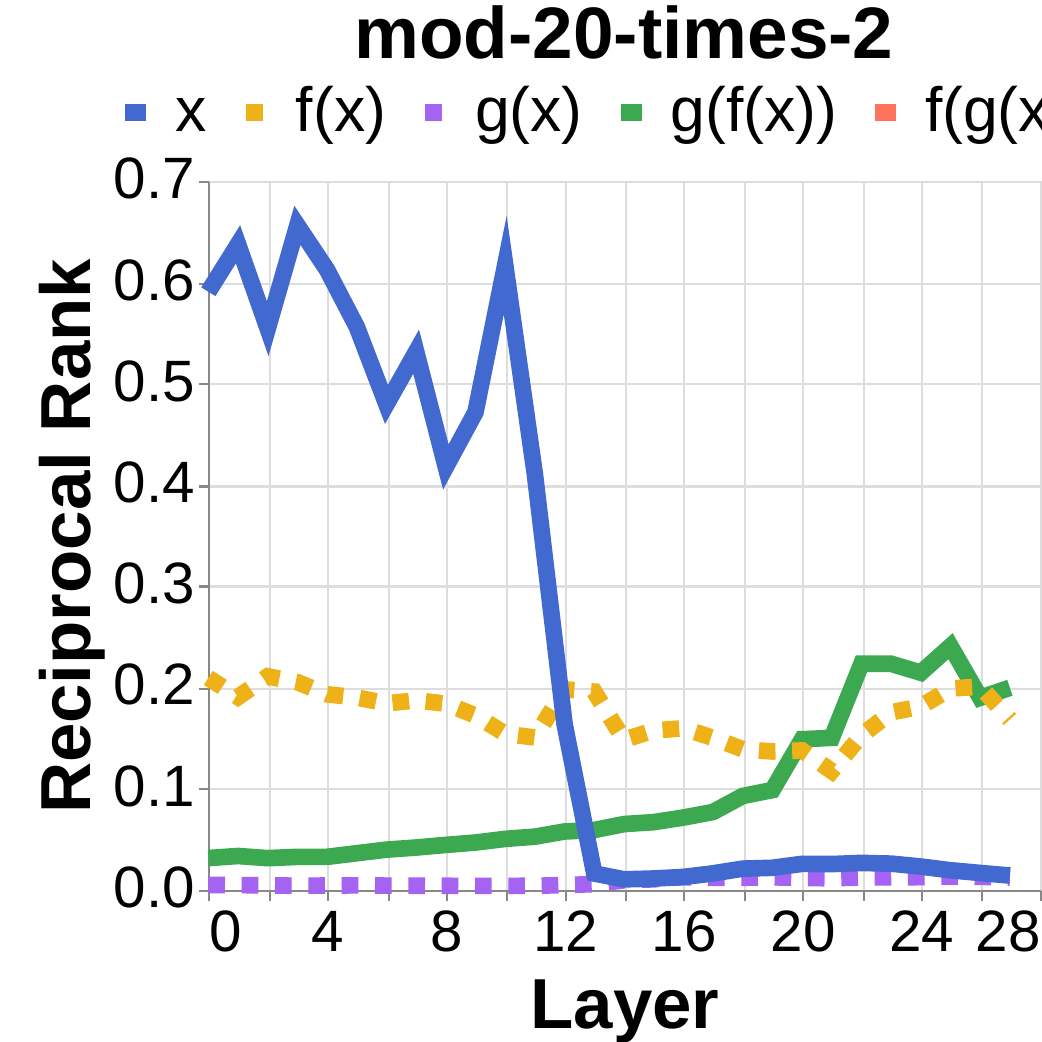}
        & \includegraphics[width=0.25\linewidth]{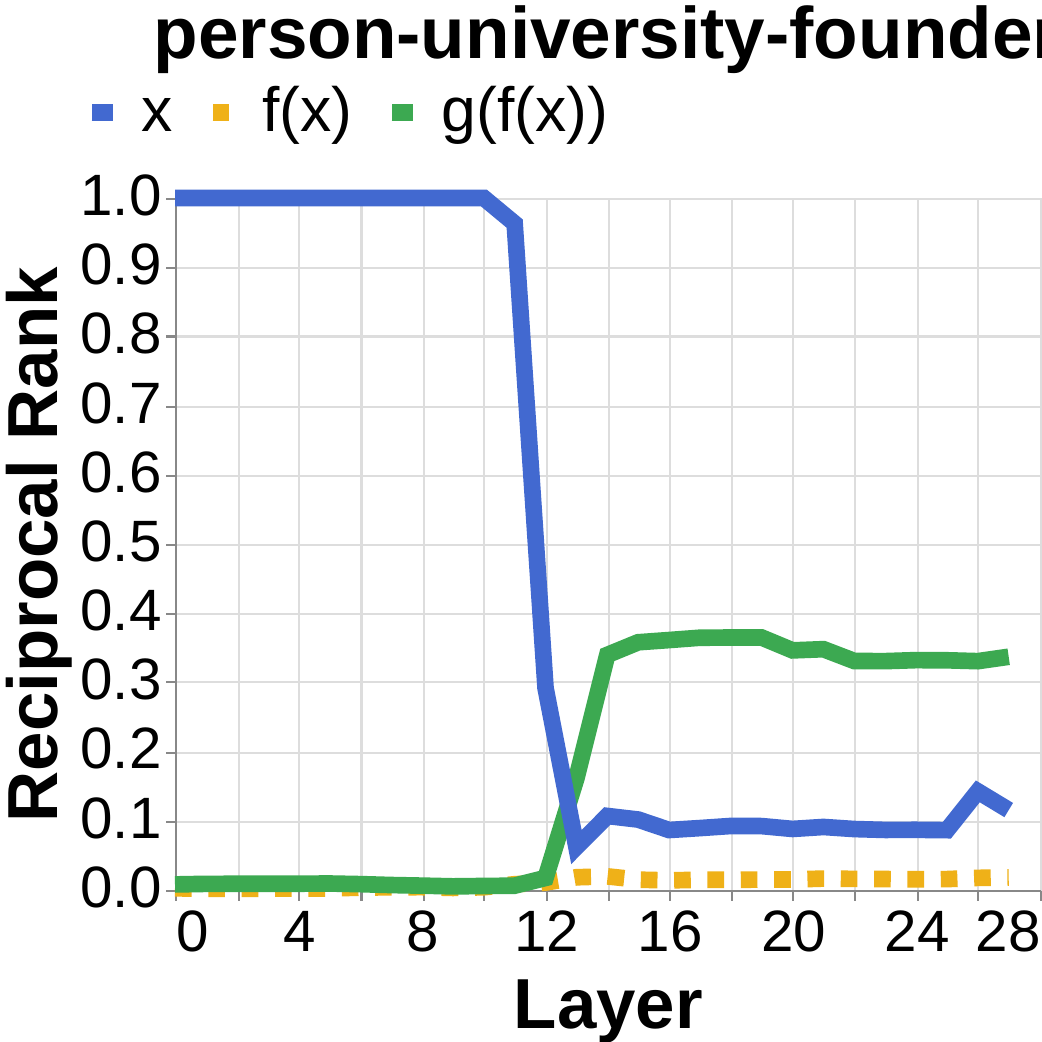}
        & \includegraphics[width=0.25\linewidth]{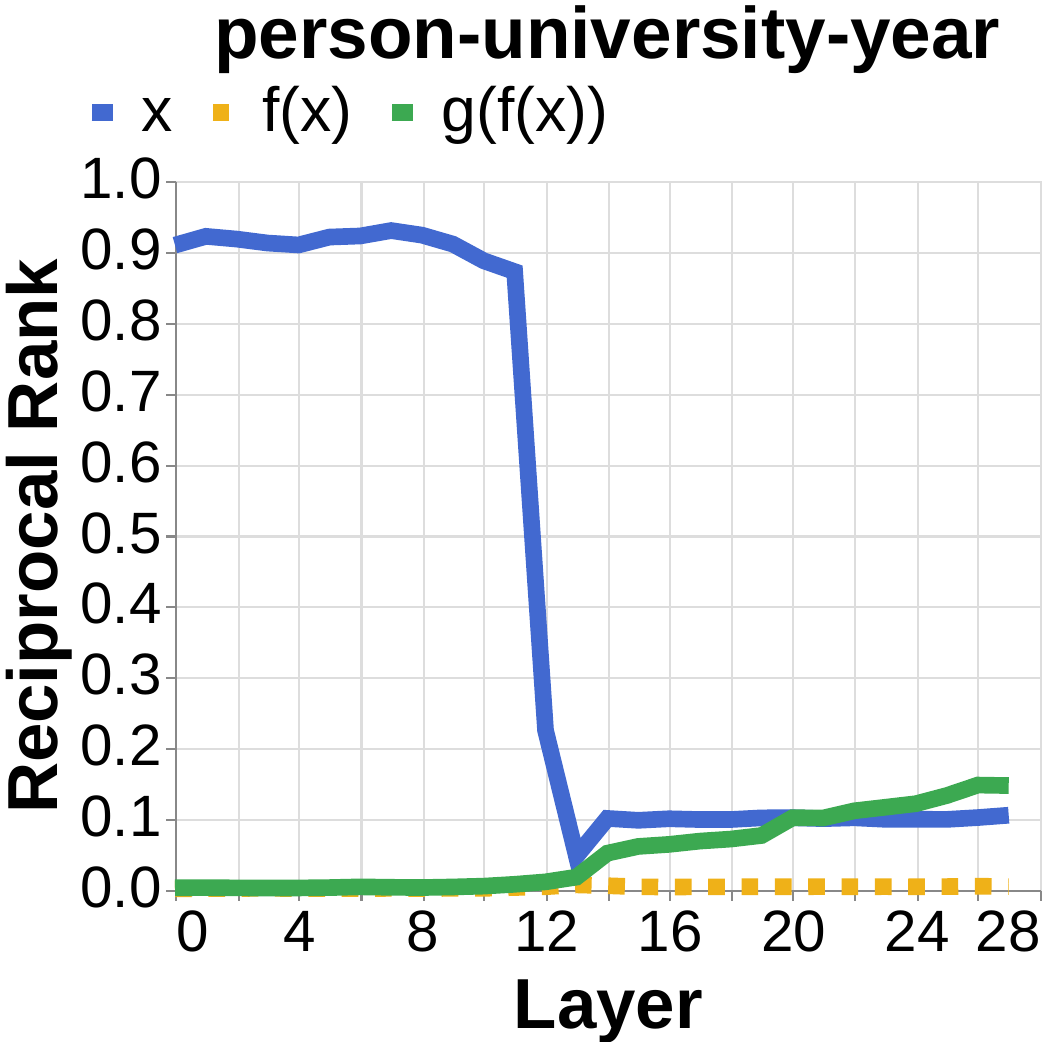}
        & \includegraphics[width=0.25\linewidth]{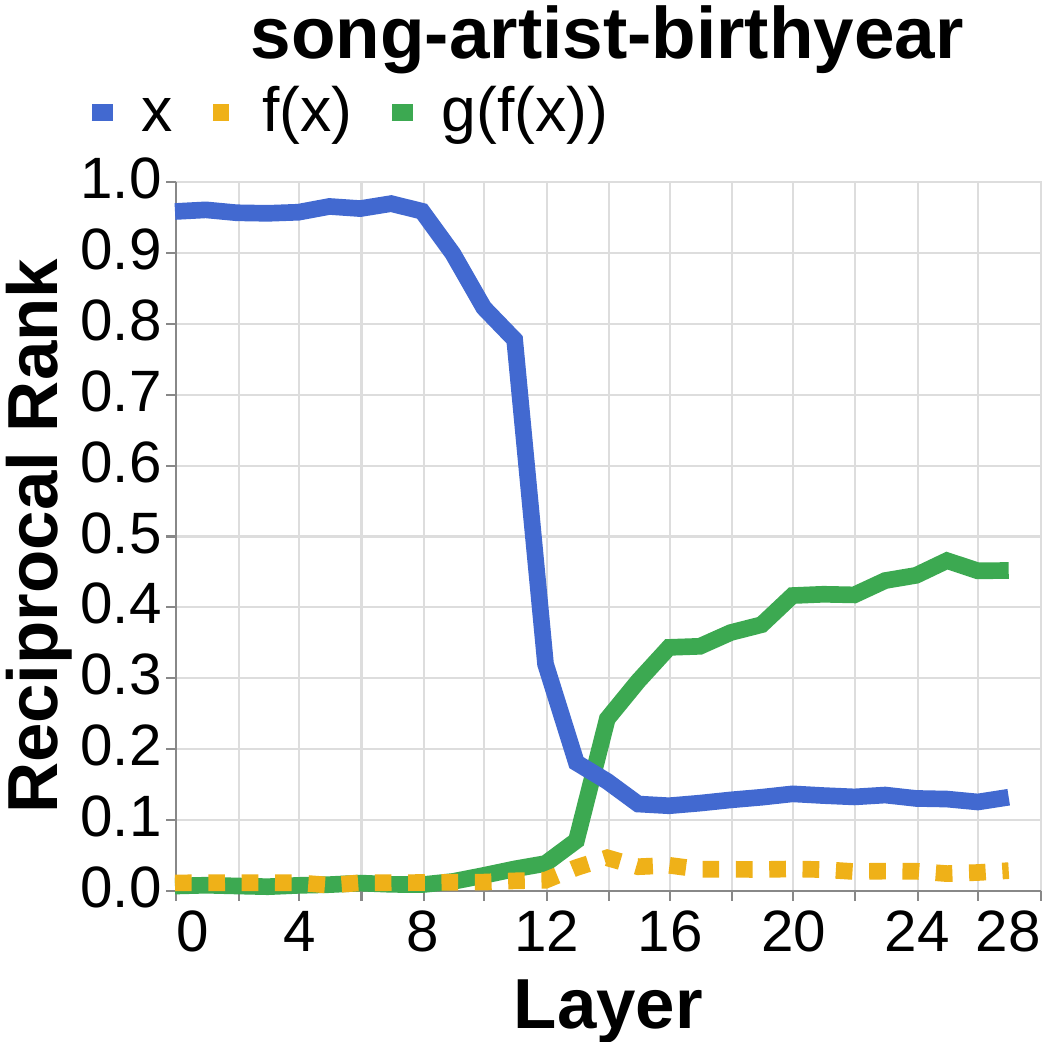} \\
    \end{tabular}
    \end{adjustbox}\end{center}
    \caption{Aggregate processing signatures (using the token identity patchscope) for each of our tasks, in which Llama 3 (3B) correctly solves all hops but not the composition for at least 10 examples.}
\end{figure}

\begin{figure}[H]
    \centering
    \includegraphics[width=0.4\linewidth]{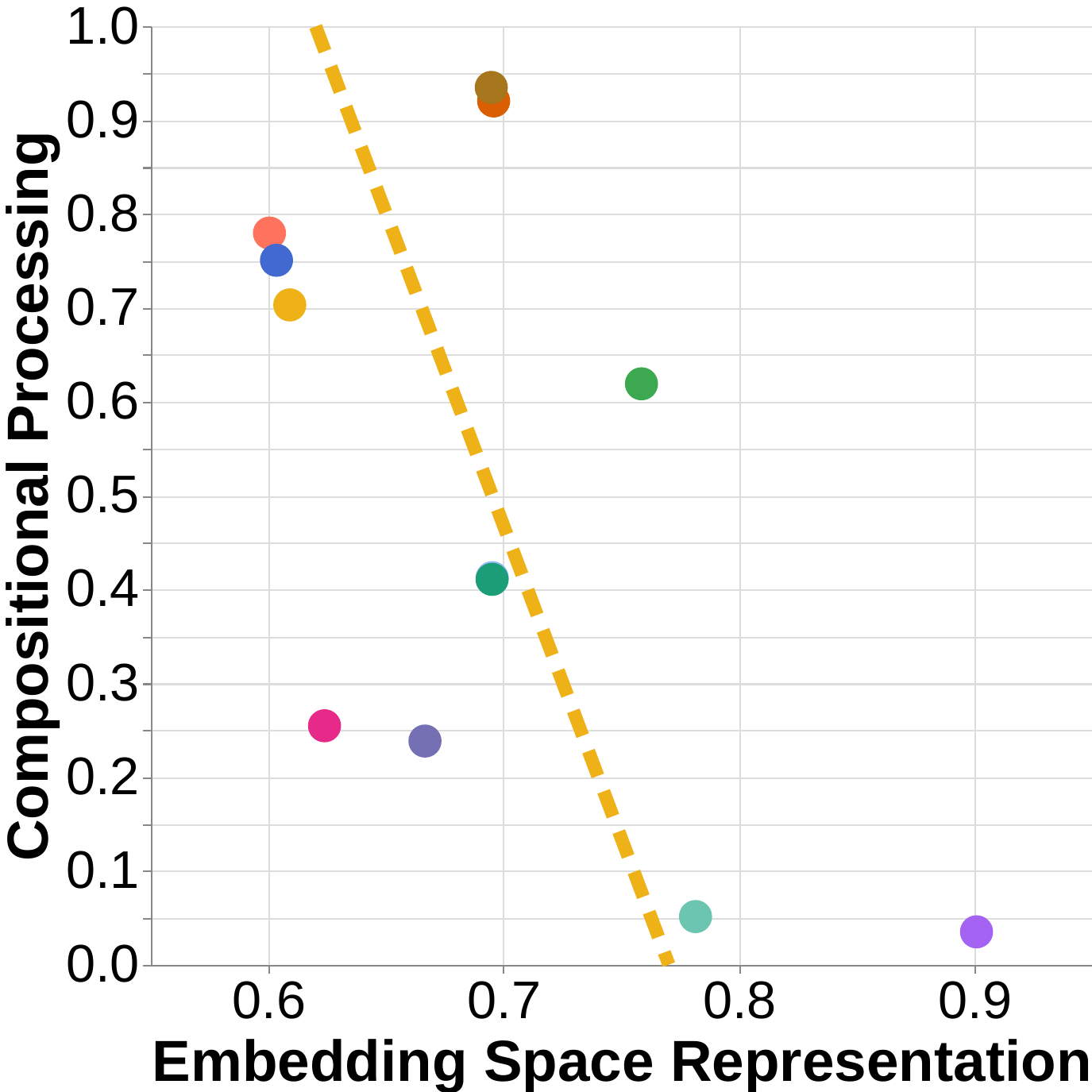}
    \caption{Correlation across tasks ($r^2 = \protect\input{artifacts/llama_3_3b/token_identity/corr/lens_task}$) for embedding space task representations and compositional processing (using the  metric from the token identity patchscope). This correlation is weaker than in \cref{fig:task-lens-correlation}, which might simply indicate that the feature of our interest (intermediate variables in our task) is better aligned with the representational subspace corresponding to logit lens.}
\end{figure}

\section{Causality of Intermediate Variables}\label{app:patching}

We would like to determine whether the variables, $f(x)$ and $g(x)$, we identify in models' intermediate representations have a causal effect on the outcome. To do so, we use activation patching \citep{vig2020:patching}, a common method for conducting causal interventions in interpretability, and patch representations across tasks.

We first identify tasks with the same $f$ but different $g$, such as \texttt{antonym-spanish} ($g \circ f$) and \texttt{antonym-german} ($g' \circ f$). For some $x$ and $x'$, we extract a single intermediate representation from the forward pass of $g'(f(x'))$ and patch it into the forward pass of $g(f(x))$. On average (over many $x$ and $x'$), we measure the causal effects on the predictions $g(f(x))$, $g(f(x'))$, $g'(f(x))$, and $g'(f(x'))$.

We extract the representation from $g'(f(x'))$ at the position and layer where $f(x)$ or $g(x)$ have the highest reciprocal rank (and only use instances where this value is at least $0.5$). We patch this representation into the forward pass for $g(f(x))$ at the median location where intermediate values are highest (layer 18 and 71st percentile query token position; identified among variables that reach RR $\geq 0.5$). We apply this intervention to two groups: instances with intermediate values that reach a peak RR $\leq 0.2$ and $\geq 0.5$. In other words, instances with direct or compositional processing signatures.

\begin{figure}[H]
    \begin{center}
    \includegraphics[width=0.92\linewidth]{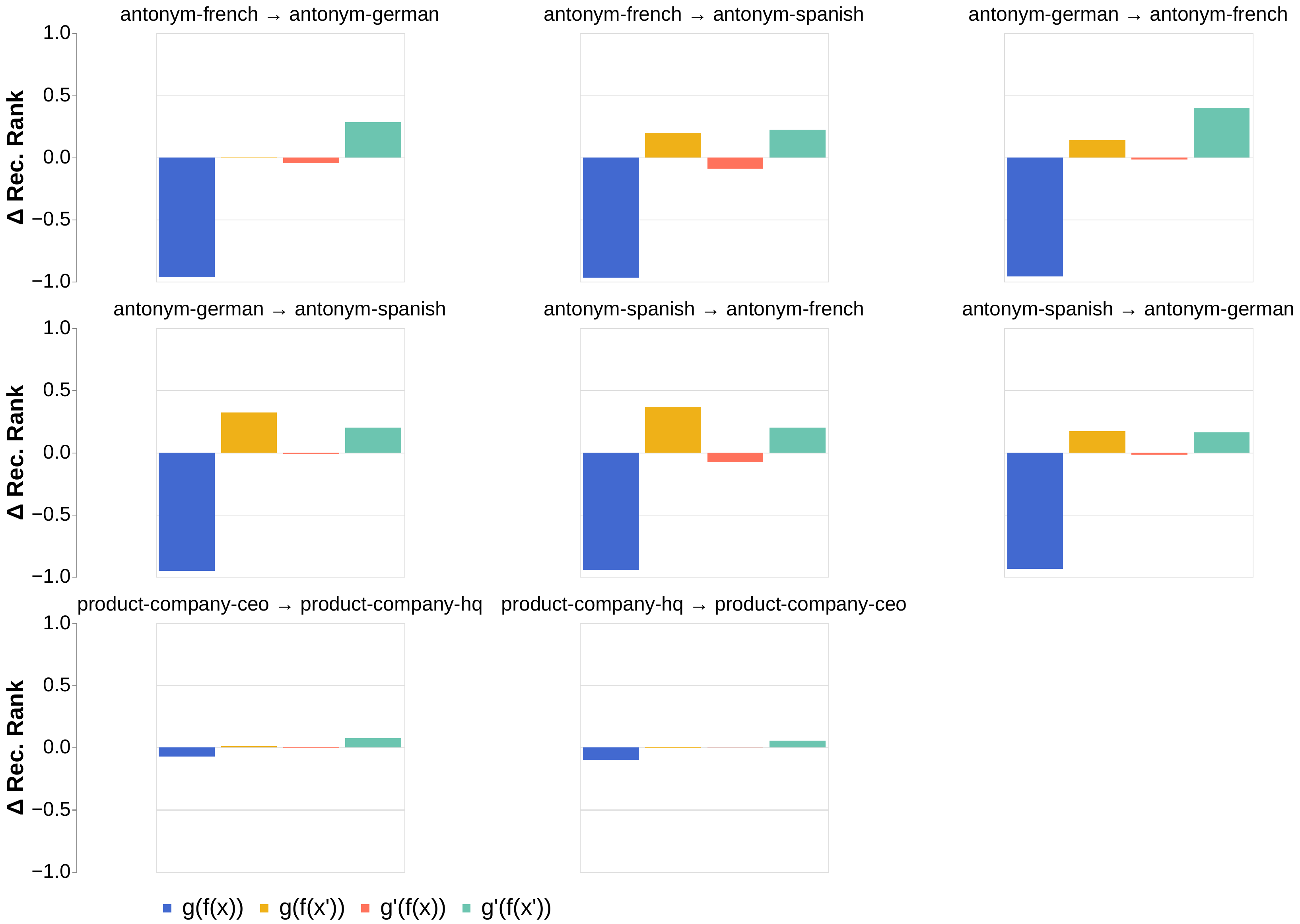}
    \end{center}
    \caption{Causal effects on predicted values after patching from $g'(f(x'))$ to $g(f(x))$ for instances with compositional processing signatures.}
\end{figure}

\begin{figure}[H]
    \begin{center}
    \includegraphics[width=0.8\linewidth]{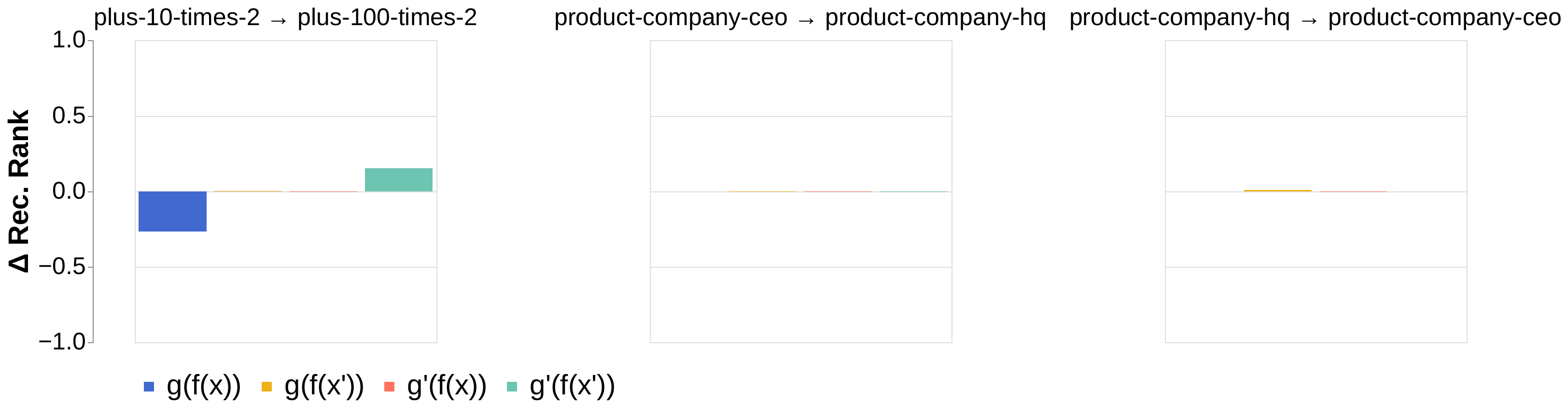}
    \end{center}
    \caption{Causal effects on predicted values after patching from $g'(f(x'))$ to $g(f(x))$ for instances with direct processing signatures.}
\end{figure}

The Antonym--Translation tasks (which tend to have compositional signatures) show the most significant causal effect: on average, $g(f(x))$ and $g'(f(x))$ decrease by -0.95 and -0.4, and $g(f(x'))$ and $g'(f(x'))$ increase by 0.20 and 0.24. The effect on $g(f(x'))$ clearly implicates the existence and causality of $f(x')$ in the patched activation; that on $g'(f(x'))$ indicates the additional existence of either itself or the function vector \citep{todd2024:fv} for $g'$ in that representation. The causal effects on compositional instances of \texttt{product-company-hq} and \texttt{product-company-ceo} are smaller.

But we can also see a clear difference between the causal effects on the compositional and direct instances. Indeed, the effects on \texttt{product-company-hq} and \texttt{product-company-ceo} are larger in their compositional instances. Patching activations from \texttt{plus-10-times-2} into \texttt{plus-100-times-2} primarily decreases $g(f(x))$ and increases $g'(f(x'))$, perhaps only implying the existence of the representation for $g'(f(x'))$ in the patched activation.

\section{Relationship between Mechanisms \& Geometry on Additional Models}\label{app:task-lens-models}

\begin{figure}[H]
    \centering
    \begin{tabular}{cc}
    \includegraphics[height=0.33\linewidth]{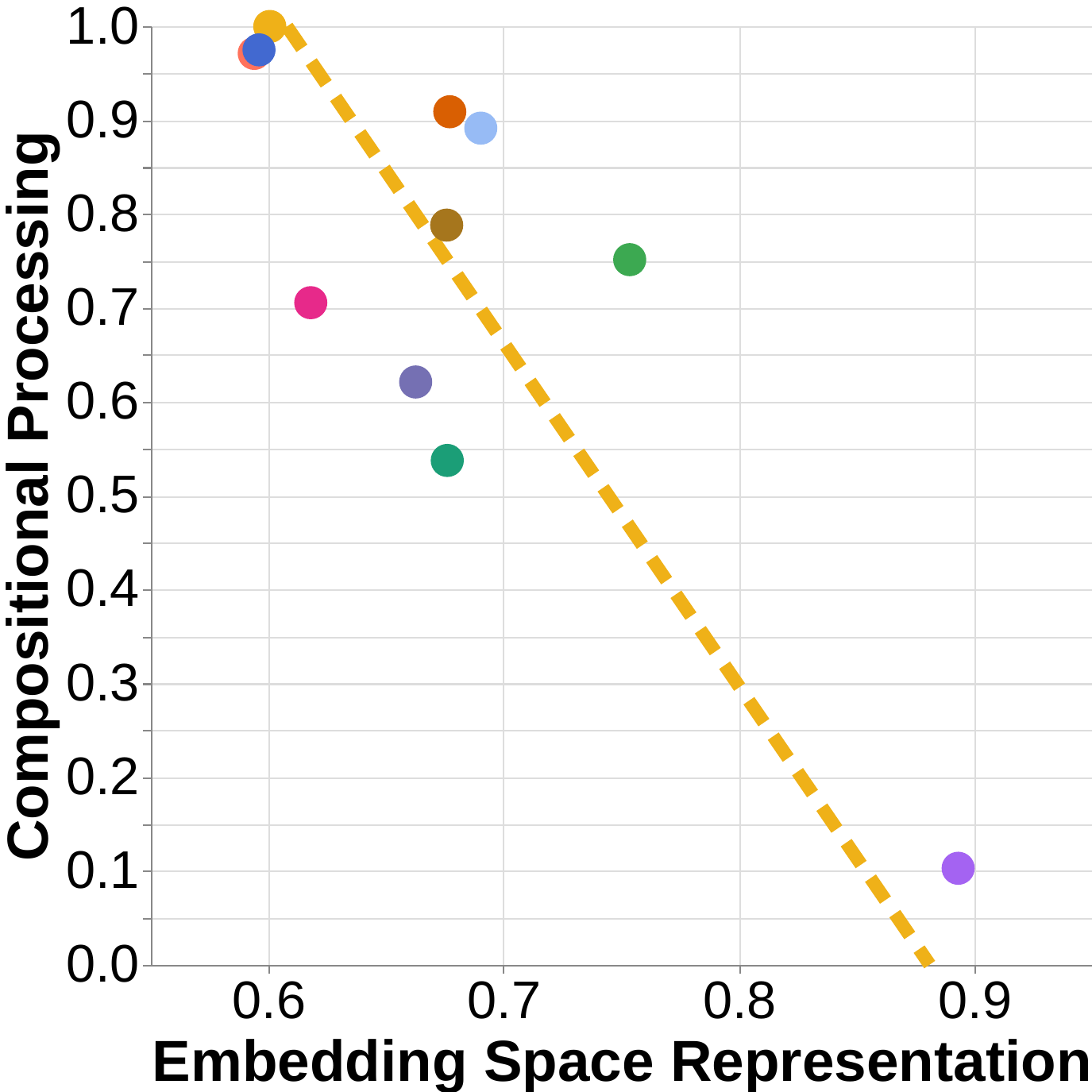} &  \includegraphics[height=0.33\linewidth]{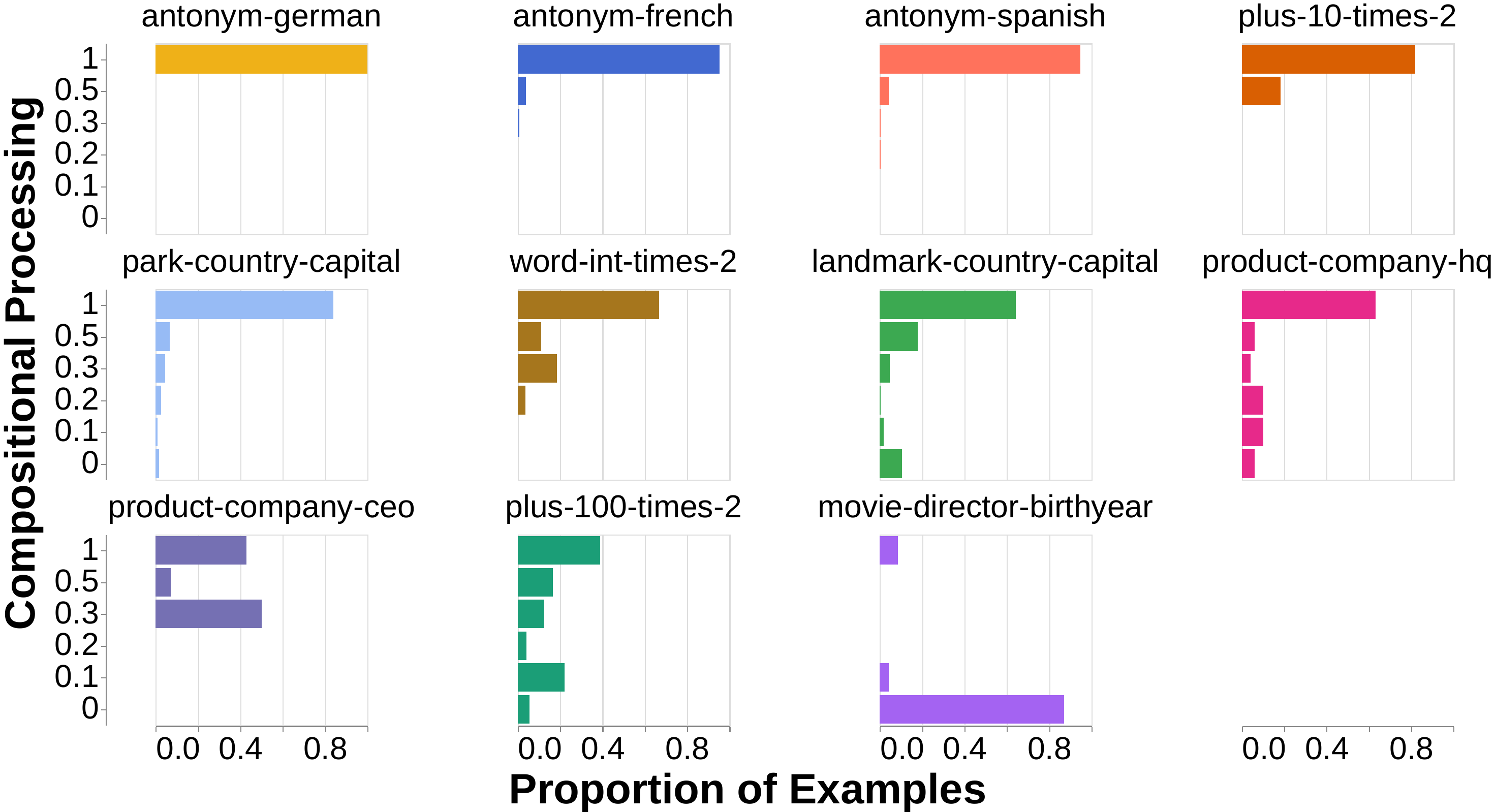} \\
    (a) & (b) \\
    \end{tabular}
    \caption{For Llama 3 (3B) Instruct. Correlation has $r^2 = \protect\input{artifacts/llama_3_3b_instruct/corr/lens_task}$. Accuracy is weakly correlated with these compositional processing ($r^2 = \protect\input{artifacts/llama_3_3b_instruct/corr/acc_lens}$) and representation ($r^2 = \protect\input{artifacts/llama_3_3b_instruct/corr/acc_task}$) metrics. \protect\input{artifacts/llama_3_3b_instruct/intermediate_var_bimodal_density} of examples have very low or high indicators for the compositional mechanism.}
\end{figure}

\begin{figure}[H]
    \centering
    \begin{tabular}{cc}
    \includegraphics[height=0.33\linewidth]{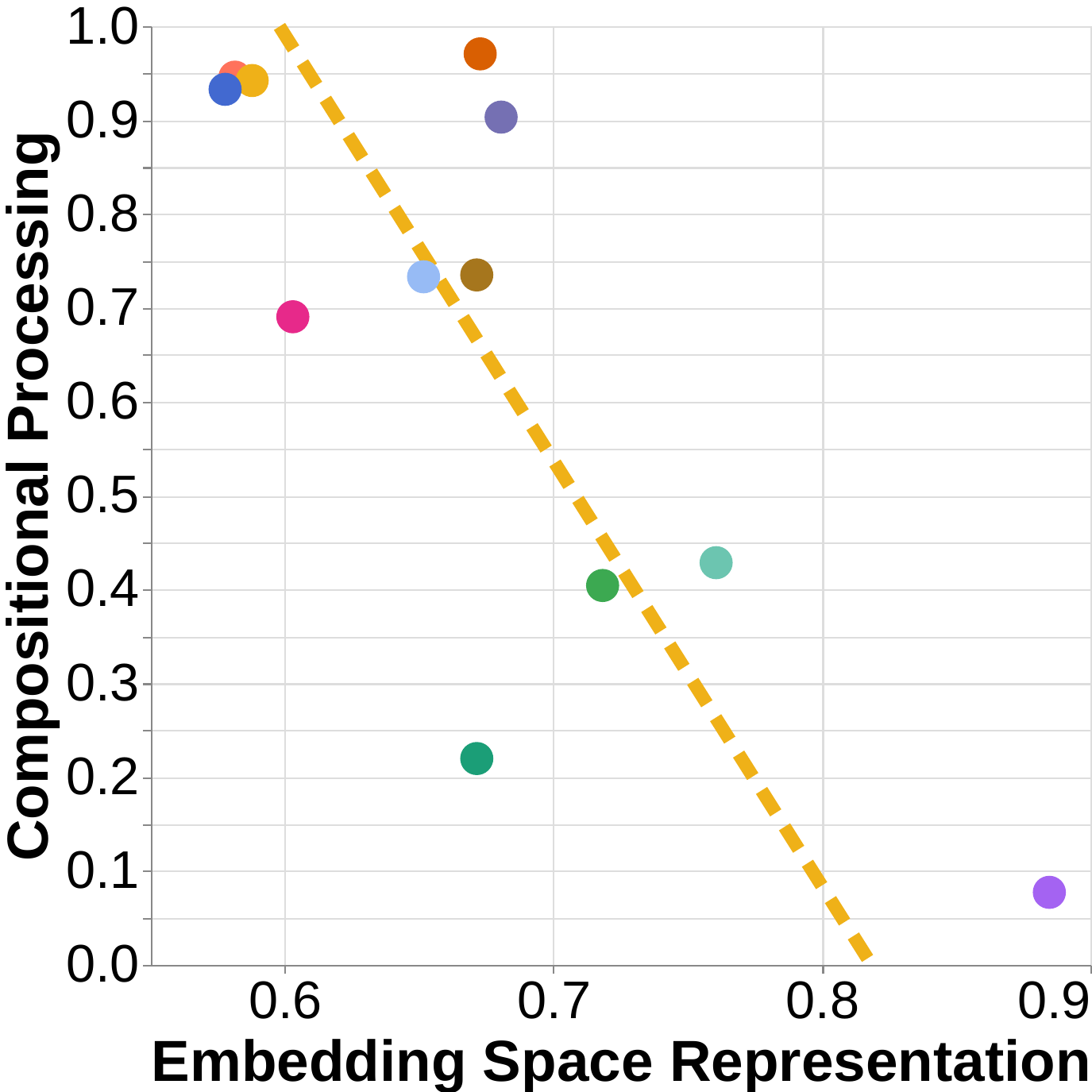} &  \includegraphics[height=0.33\linewidth]{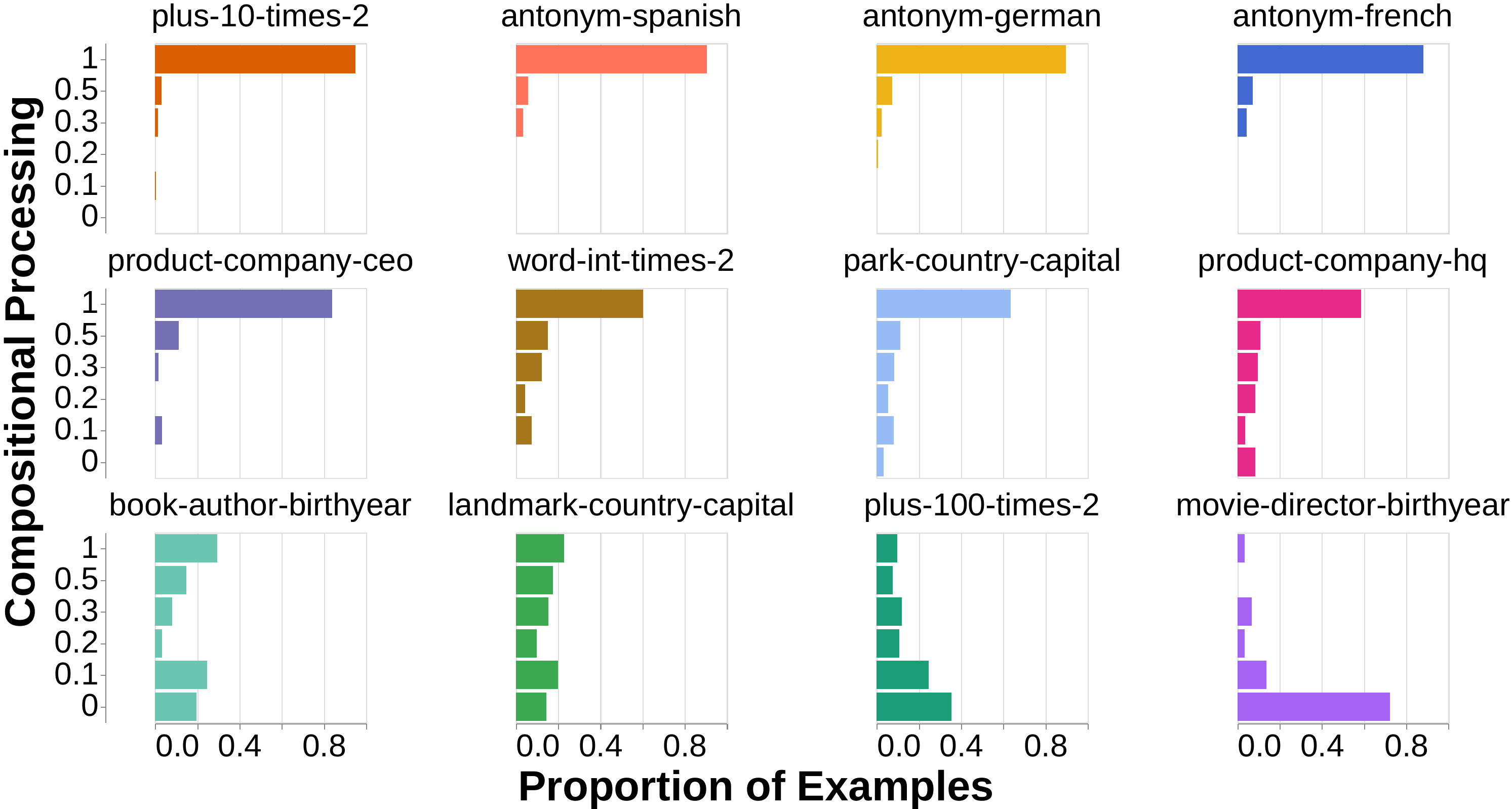} \\
    (a) & (b) \\
    \end{tabular}
    \caption{For Llama 3 (8B). Correlation has $r^2 = \protect\input{artifacts/llama_3_8b/corr/lens_task}$. Accuracy is weakly correlated with these compositional processing ($r^2 = \protect\input{artifacts/llama_3_8b/corr/acc_lens}$) and representation ($r^2 = \protect\input{artifacts/llama_3_8b/corr/acc_task}$) metrics. \protect\input{artifacts/llama_3_8b/intermediate_var_bimodal_density} of examples have very low or high indicators for the compositional mechanism.}
\end{figure}

\begin{figure}[H]
    \centering
    \begin{tabular}{cc}
    \includegraphics[height=0.33\linewidth]{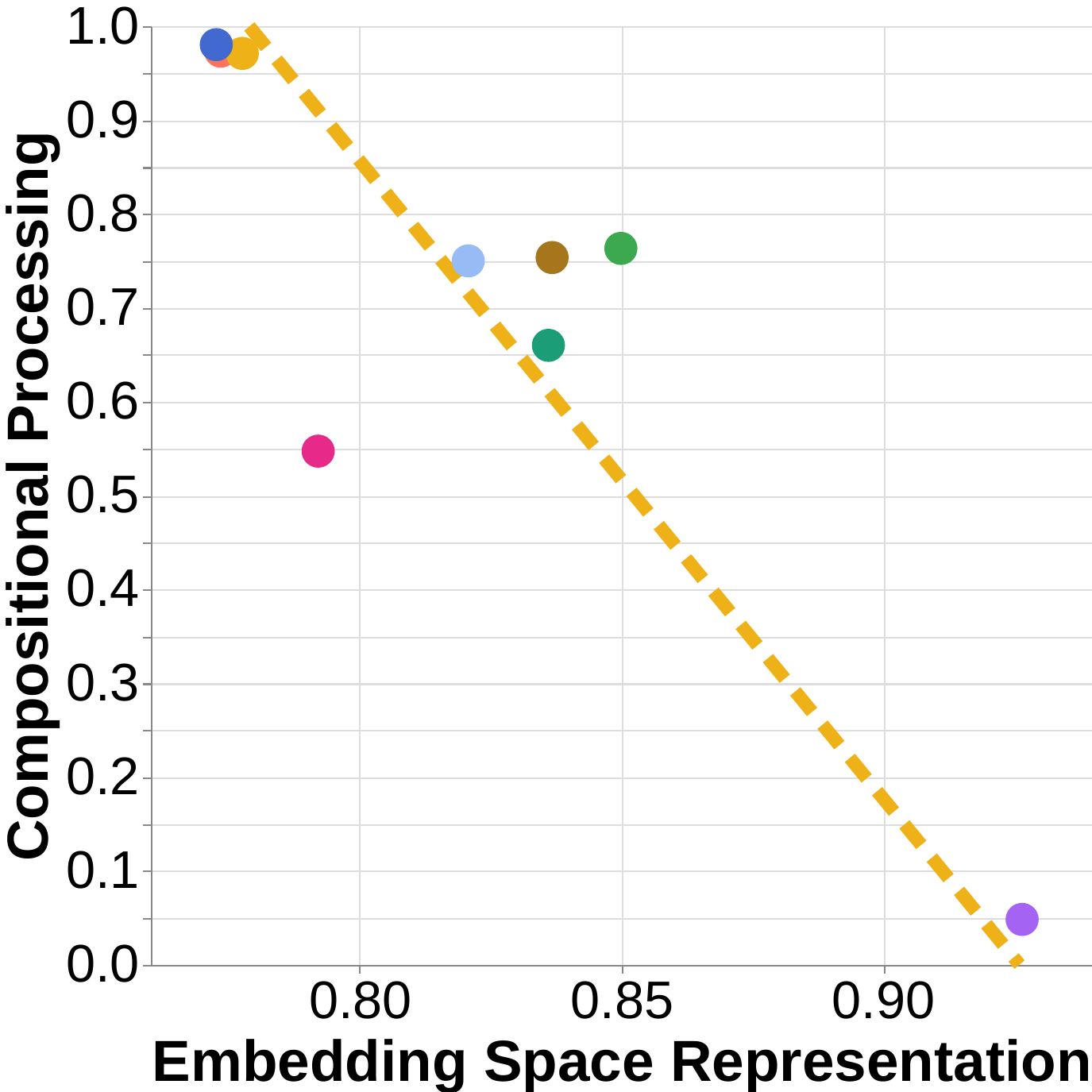} &  \includegraphics[height=0.33\linewidth]{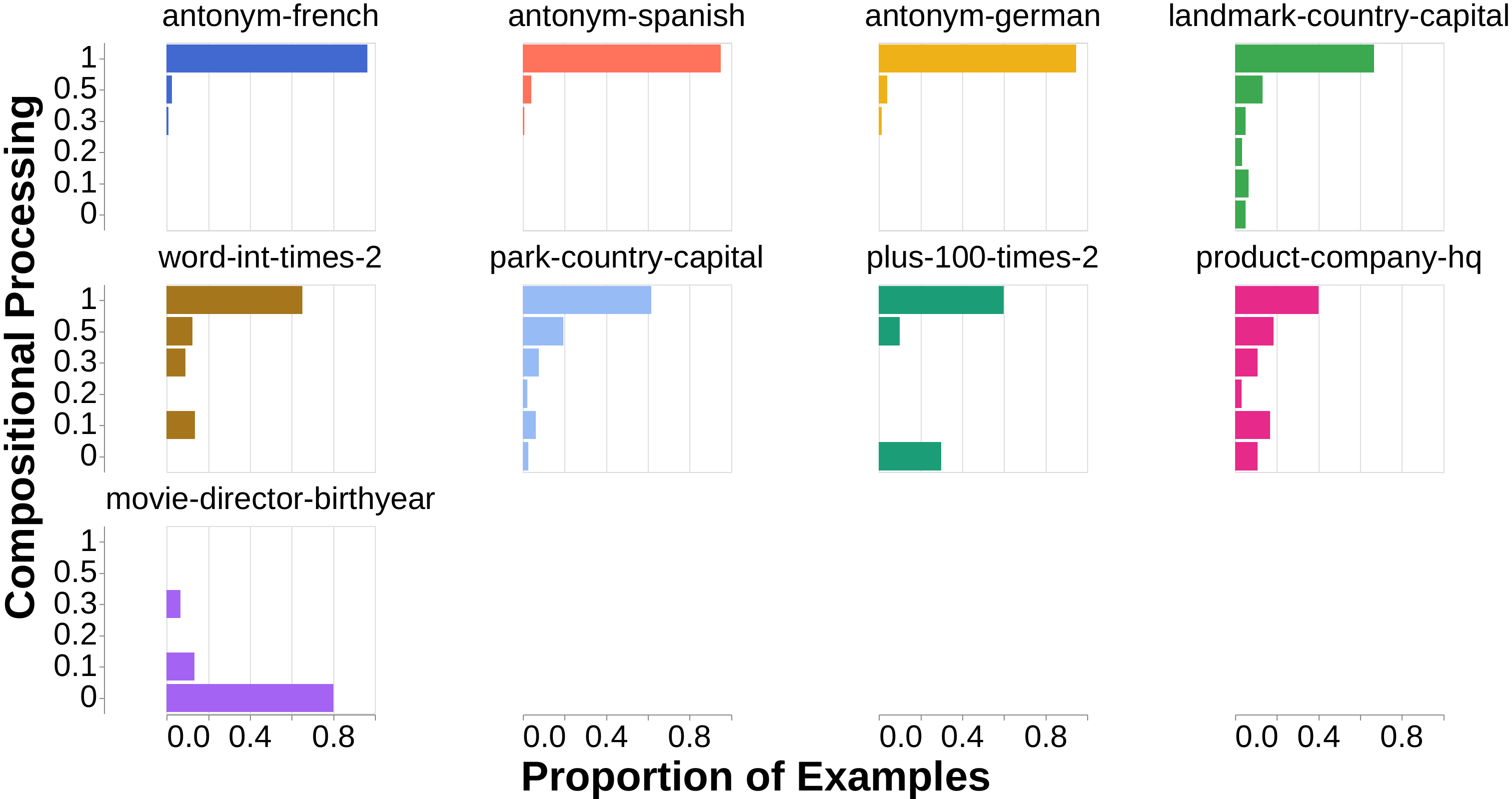} \\
    (a) & (b) \\
    \end{tabular}
    \caption{For OLMo 2 (7B). Correlation has $r^2 = \protect\input{artifacts/olmo_2_7b/corr/lens_task}$. Accuracy is weakly correlated with these compositional processing ($r^2 = \protect\input{artifacts/olmo_2_7b/corr/acc_lens}$) and representation ($r^2 = \protect\input{artifacts/olmo_2_7b/corr/acc_task}$) metrics. \protect\input{artifacts/olmo_2_7b/intermediate_var_bimodal_density} of examples have very low or high indicators for the compositional mechanism.}
\end{figure}

\begin{figure}[H]
    \centering
    \begin{tabular}{cc}
    \includegraphics[height=0.33\linewidth]{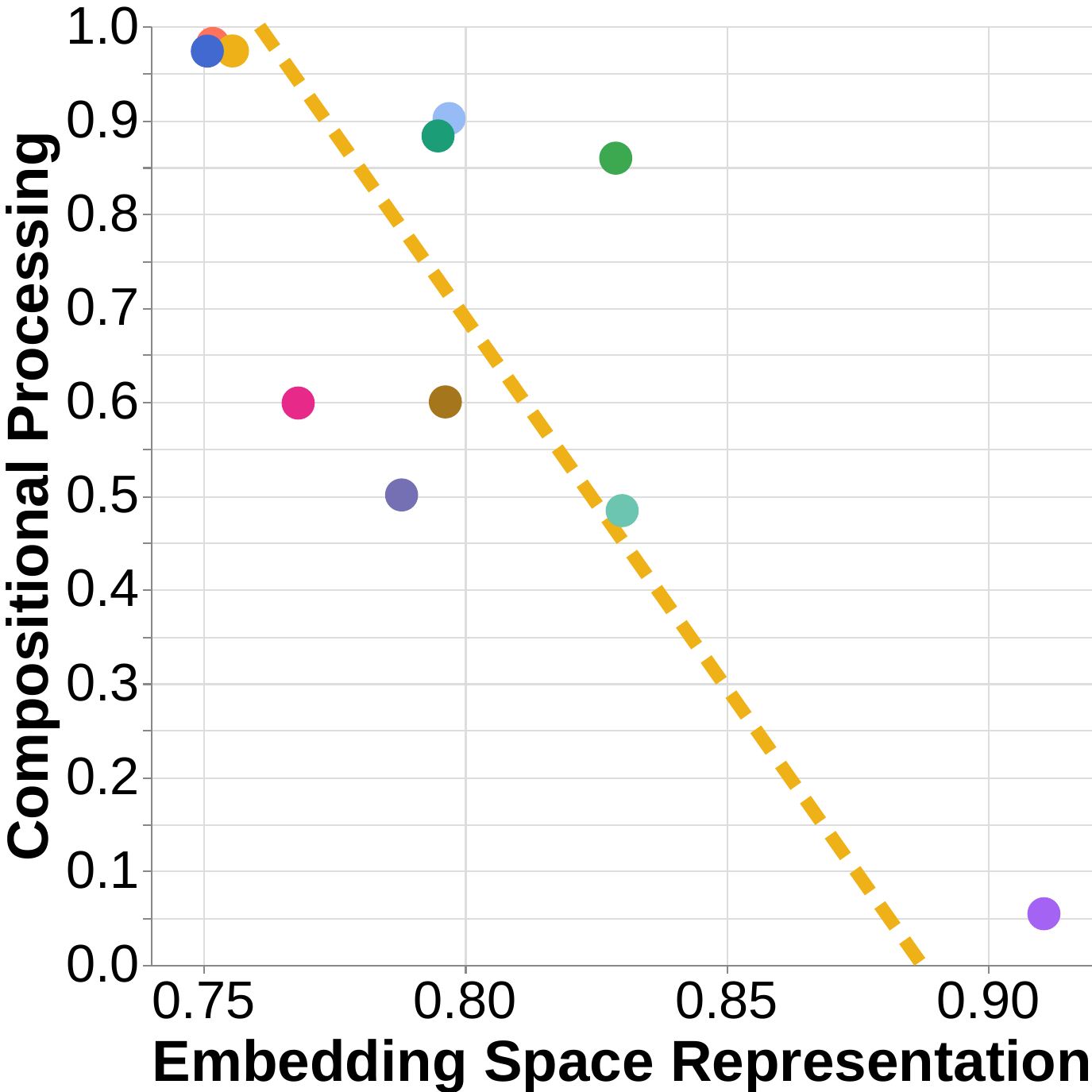} &  \includegraphics[height=0.33\linewidth]{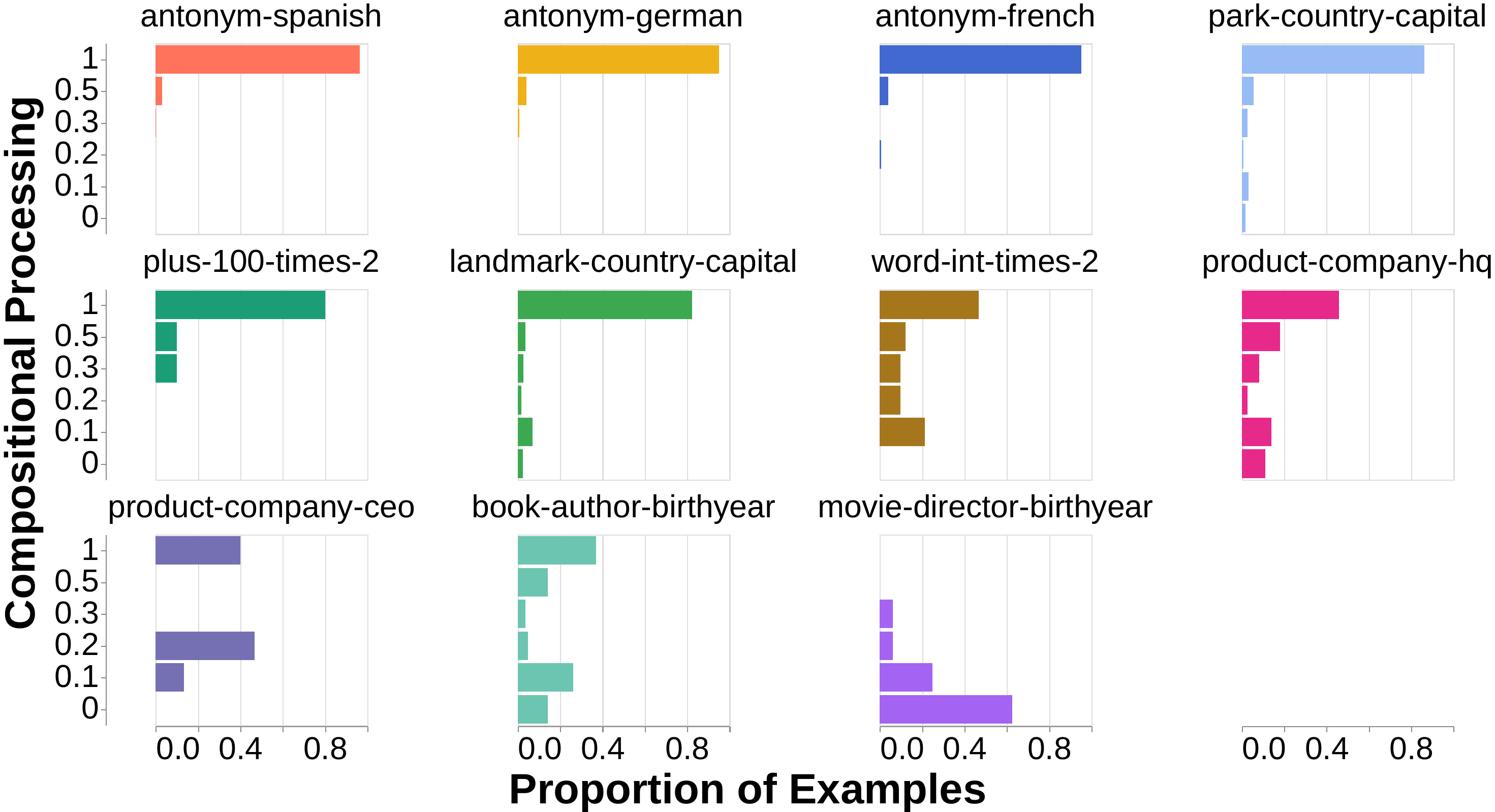} \\
    (a) & (b) \\
    \end{tabular}
    \caption{For OLMo 2 (13B). Correlation has $r^2 = \protect\input{artifacts/olmo_2_13b/corr/lens_task}$. Accuracy is weakly correlated with these compositional processing ($r^2 = \protect\input{artifacts/olmo_2_13b/corr/acc_lens}$) and representation ($r^2 = \protect\input{artifacts/olmo_2_13b/corr/acc_task}$) metrics. \protect\input{artifacts/olmo_2_13b/intermediate_var_bimodal_density} of examples have very low or high indicators for the compositional mechanism.}
\end{figure}

\newpage
\section{Relationship between Mechanisms and Representations of Hops}\label{app:hops}

Similarly to the experiment in \cref{sec:linearity,fig:task-lens-correlation}, we investigate the relationship between our compositionality heuristic and embedding space representations for variations of the hops (rather than of the compositional task) with the Llama 3 (3B) model. The representation of the second hop shown here also correlates strongly with the mechanism, but not as strongly as the translation-based hypothesis in \cref{sec:linearity}.

\begin{figure}[H]
    \begin{center}
        \includegraphics[width=0.8\linewidth]{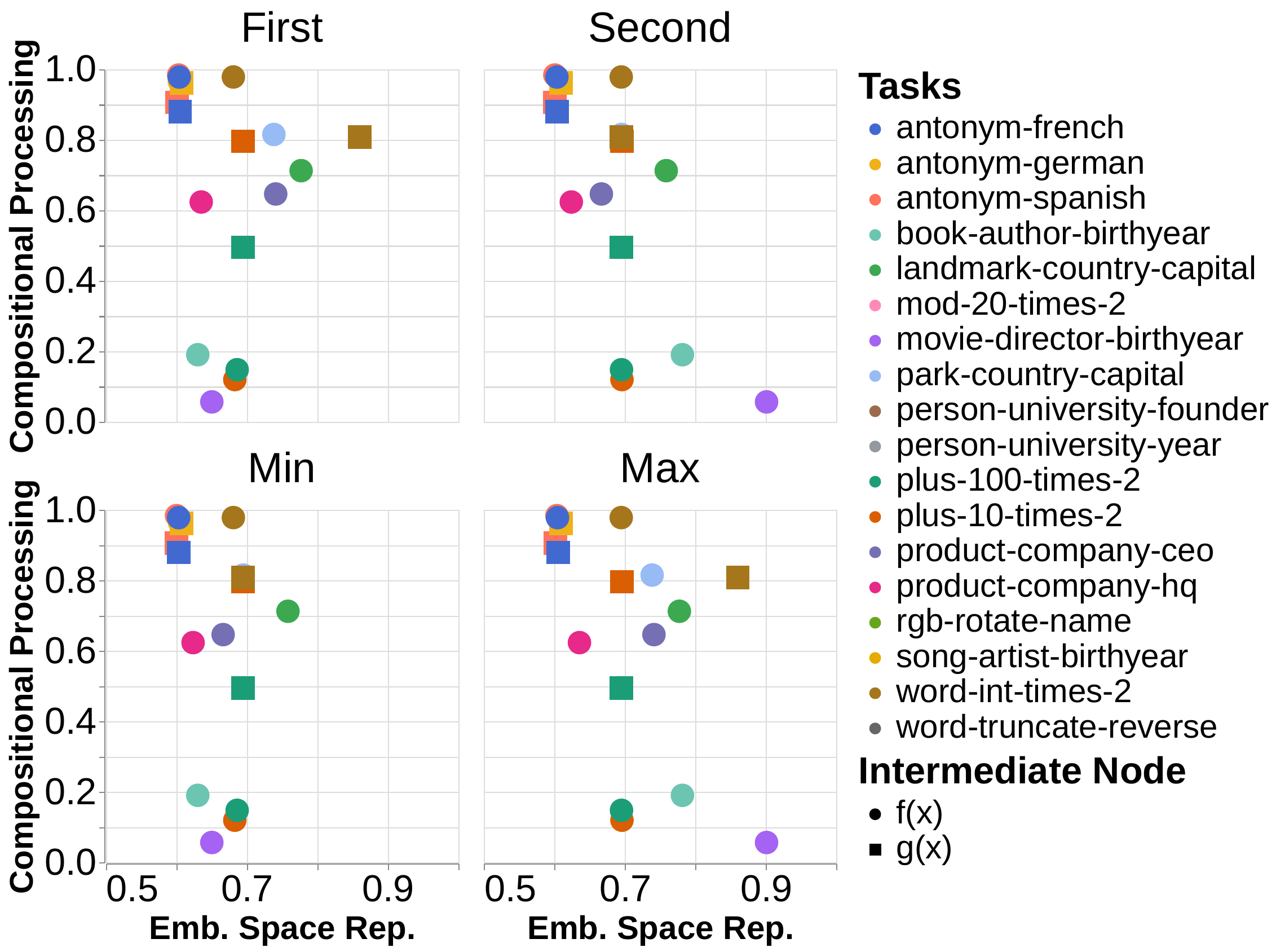}
    \end{center}
    \caption{Relationships between compositional processing and embedding space representations for the hops. $r^2 = \protect\input{artifacts/llama_3_3b/corr/lens_first_hop}$ against the representation of the first hop; $r^2 = \protect\input{artifacts/llama_3_3b/corr/lens_second_hop}$ against the second hop; $r^2 = \protect\input{artifacts/llama_3_3b/corr/lens_min_hop}$ using the minimum representation between the hops; and $r^2 = \protect\input{artifacts/llama_3_3b/corr/lens_max_hop}$ using the maximum.}
    \label{fig:linearity-correlations}
\end{figure}

\newpage
\section{Relationship between Mechanisms and Linear Task Representations}\label{app:linear}

We repeat the experiments in \cref{sec:linearity,app:hops,app:token-identity}, but test for linear representations of tasks (instead of translations). Across models, we find an average correlation of $r^2 = 0.53$ between compositional task representations and mechanisms, which is also strong but not as promising as our translation-based relationship. Future work will need to identify under which conditions are tasks better represented with translations or linear transformations and which task representation models actually compute within their layer-wise processing.

\begin{figure}[H]
    \centering
    \begin{tabular}{ccc}
    Llama 3 (3B) & Llama 3 (3B) Instruct & Llama 3 (8B) \\
    \includegraphics[height=0.33\linewidth]{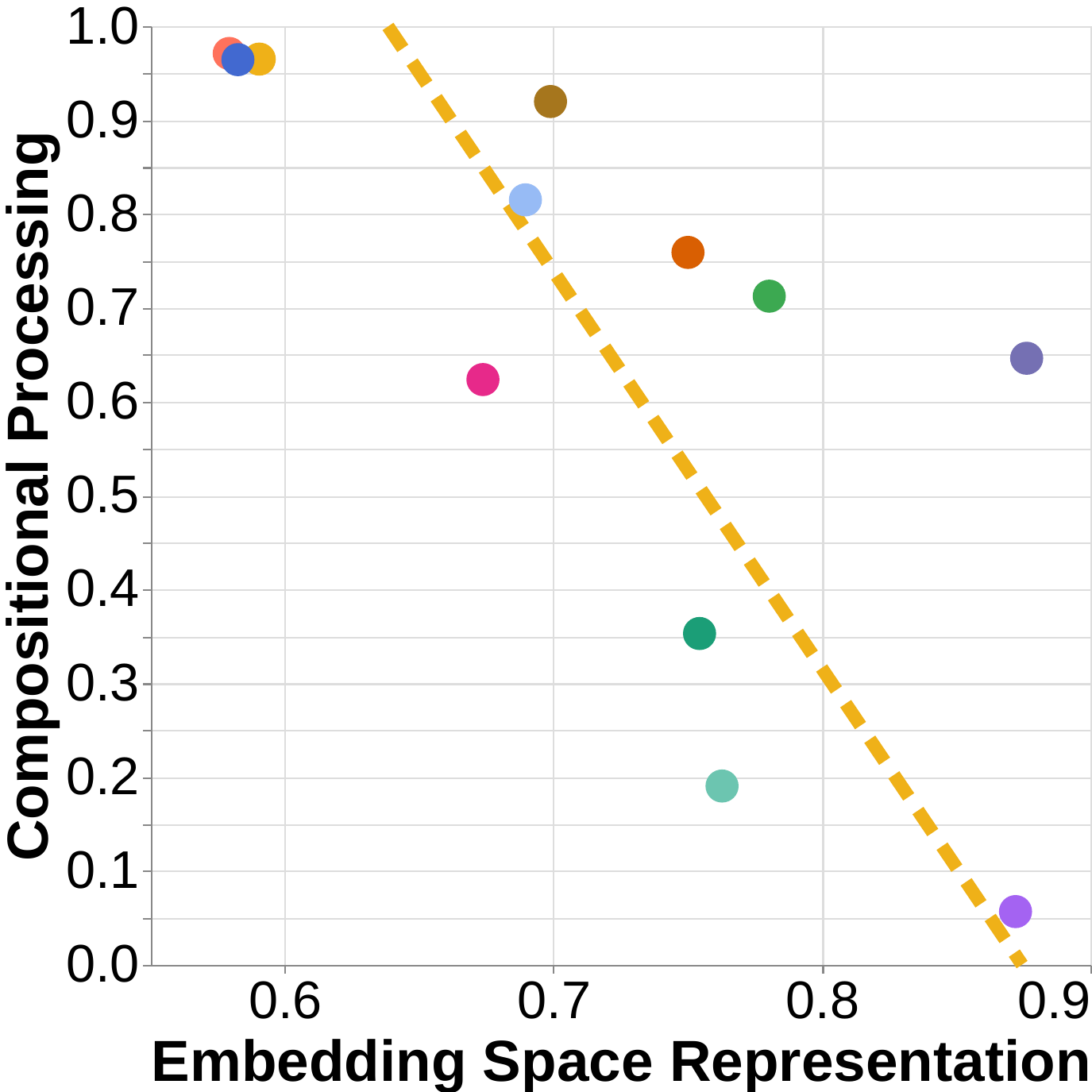} & \includegraphics[height=0.33\linewidth]{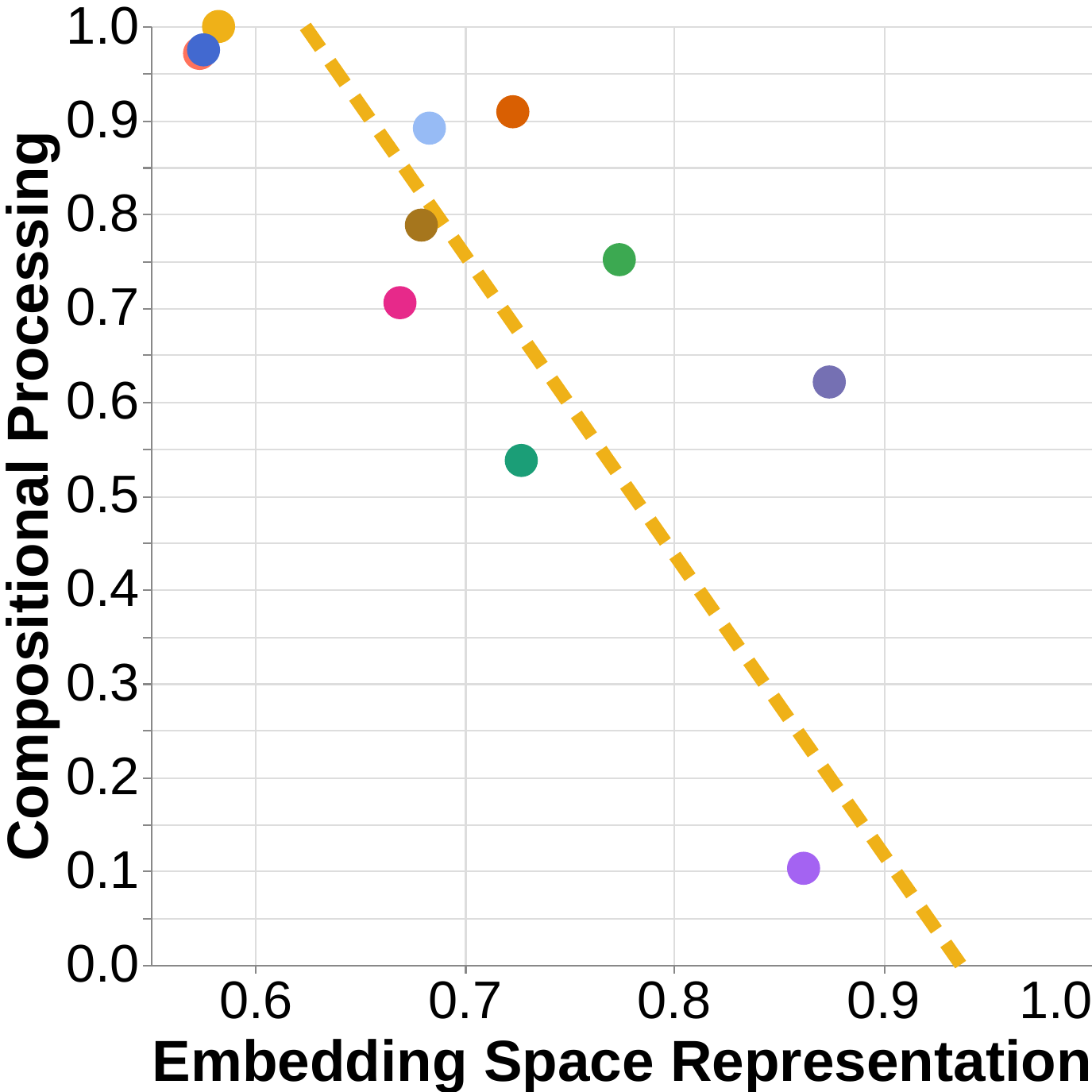} & \includegraphics[height=0.33\linewidth]{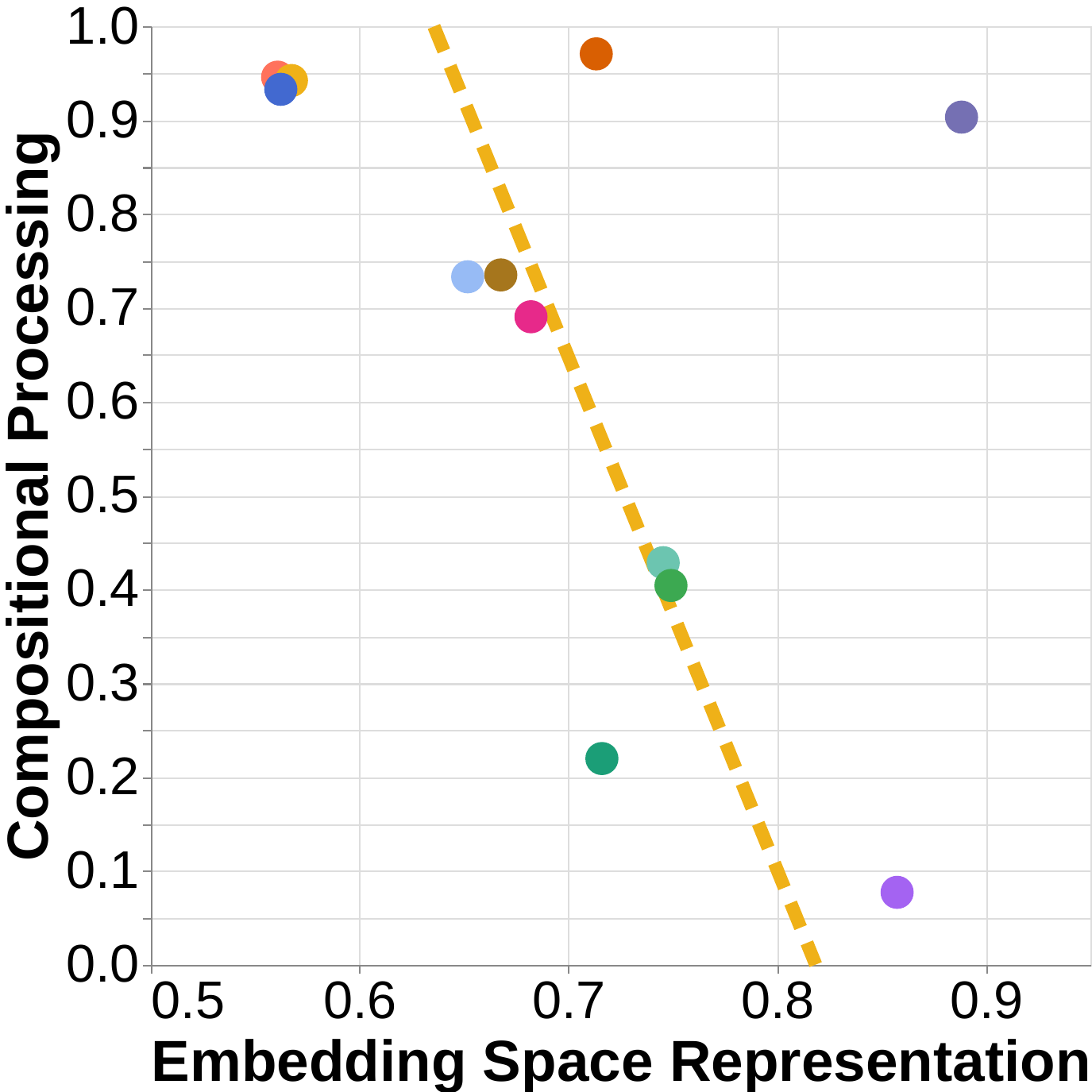} \\ \\
    OLMo 2 (7B) & OLMo 2 (13B) & \\
    \includegraphics[height=0.33\linewidth]{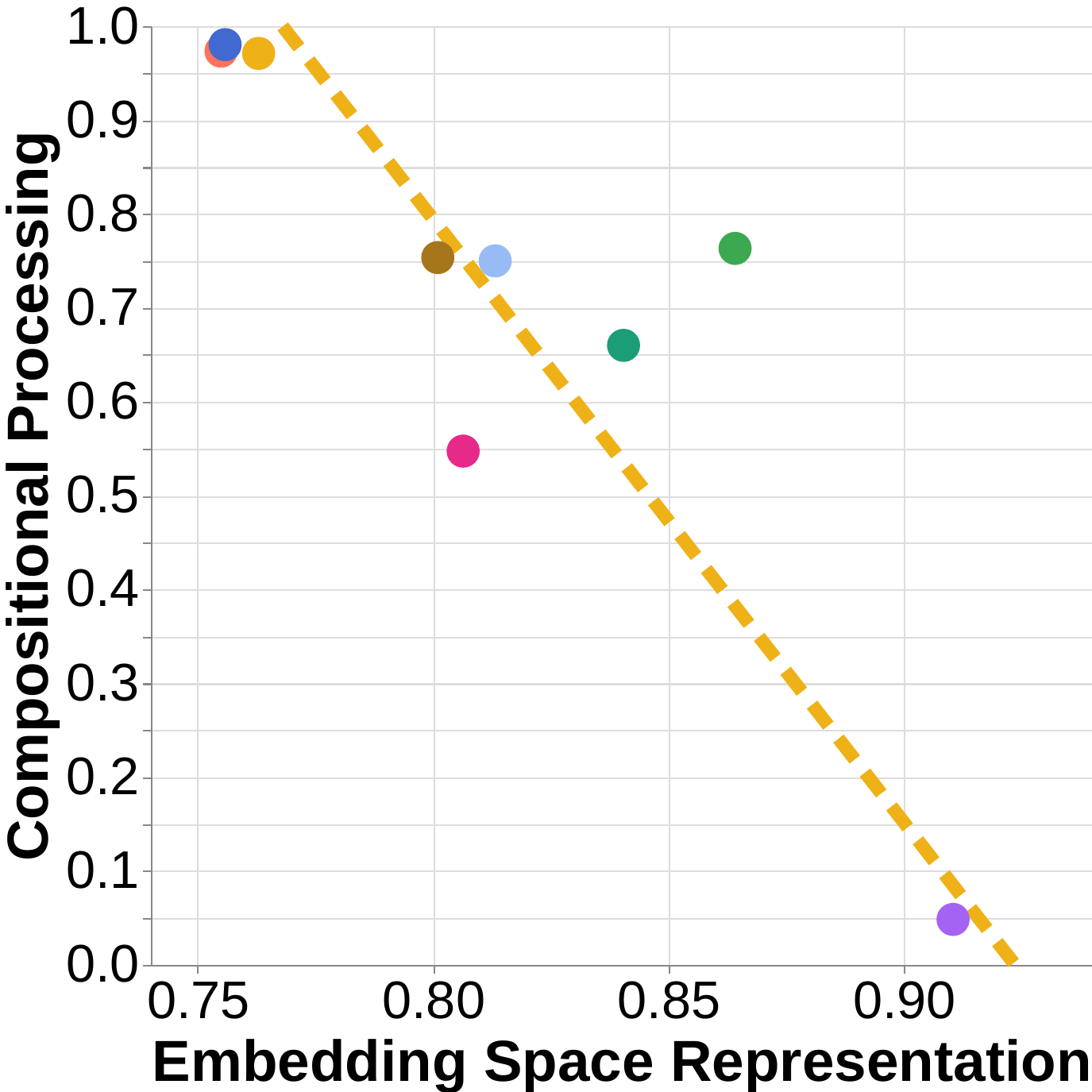} & \includegraphics[height=0.33\linewidth]{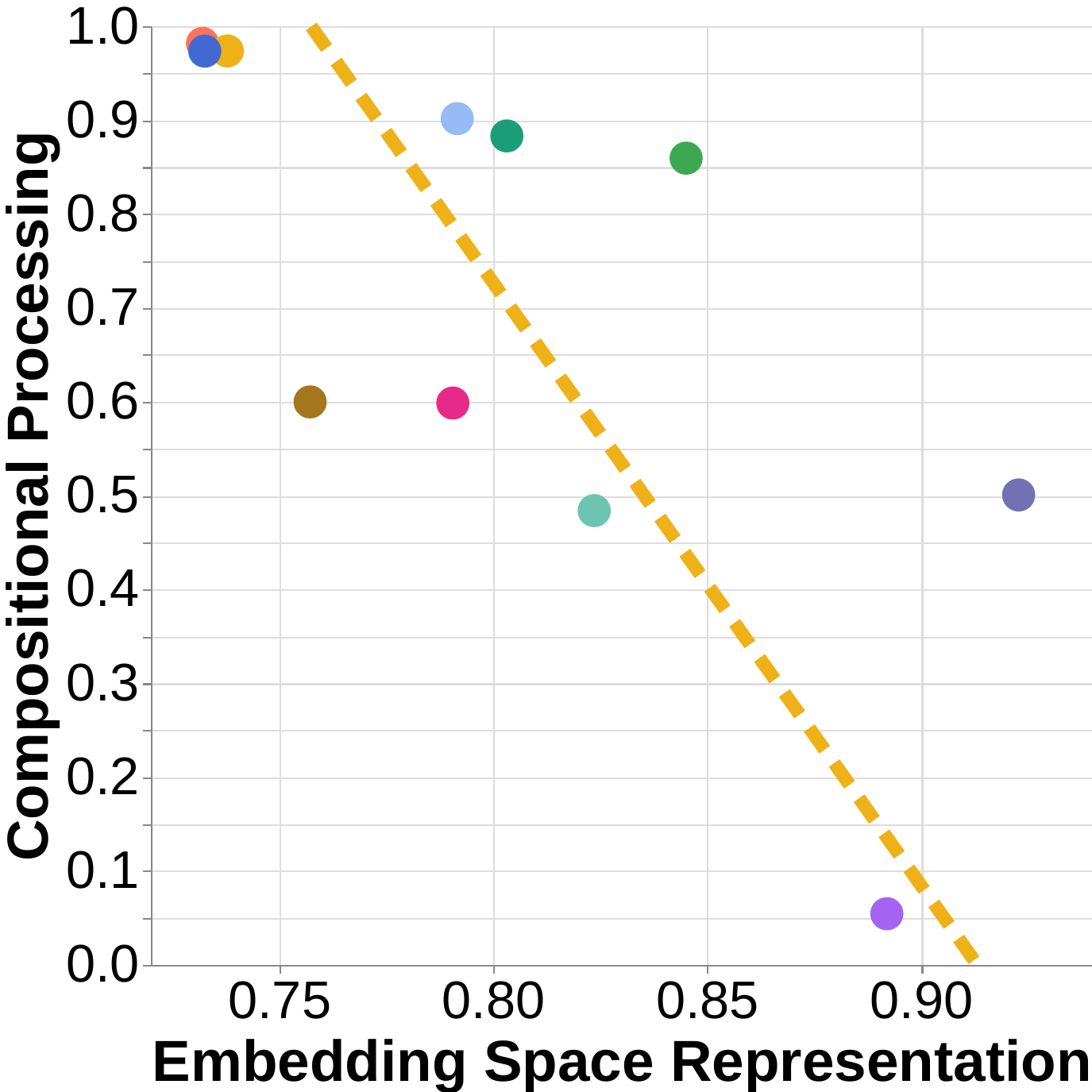} & \\
    \end{tabular}
    \caption{Llama 3 (3B) correlates with $r^2 = \protect\input{artifacts/linear/llama_3_3b/corr/lens_task}$, Llama 3 (3B) Instruct with $r^2 = \protect\input{artifacts/linear/llama_3_3b_instruct/corr/lens_task}$, Llama 3 (8B) correlates with $r^2 = \protect\input{artifacts/linear/llama_3_8b/corr/lens_task}$, OLMo 2 (7B) with $r^2 = \protect\input{artifacts/linear/olmo_2_7b/corr/lens_task}$, and OLMo 2 (13B) with $r^2 = \protect\input{artifacts/linear/olmo_2_13b/corr/lens_task}$.}
\end{figure}

\begin{figure}[H]
    \begin{center}
        \includegraphics[width=0.8\linewidth]{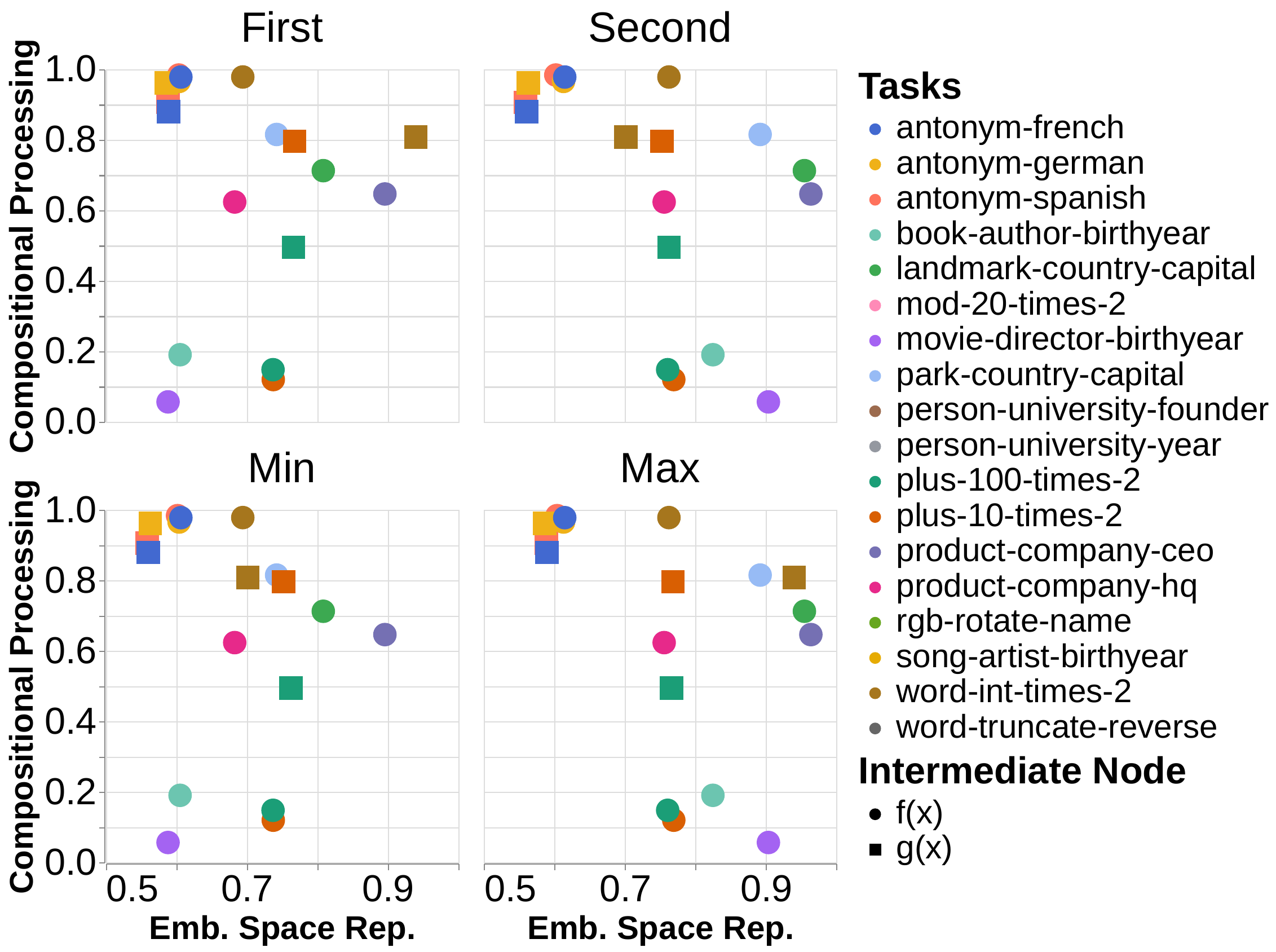}
    \end{center}
    \caption{Relationships between compositional processing and linear embedding space representations for the hops. $r^2 = \protect\input{artifacts/linear/llama_3_3b/corr/lens_first_hop}$ against the representation of the first hop; $r^2 = \protect\input{artifacts/linear/llama_3_3b/corr/lens_second_hop}$ against the second hop; $r^2 = \protect\input{artifacts/linear/llama_3_3b/corr/lens_min_hop}$ using the minimum representation between the hops; and $r^2 = \protect\input{artifacts/linear/llama_3_3b/corr/lens_max_hop}$ using the maximum.}
    \label{fig:hop-correlations-linear}
\end{figure}

\begin{figure}[H]
    \centering
    \includegraphics[width=0.5\linewidth]{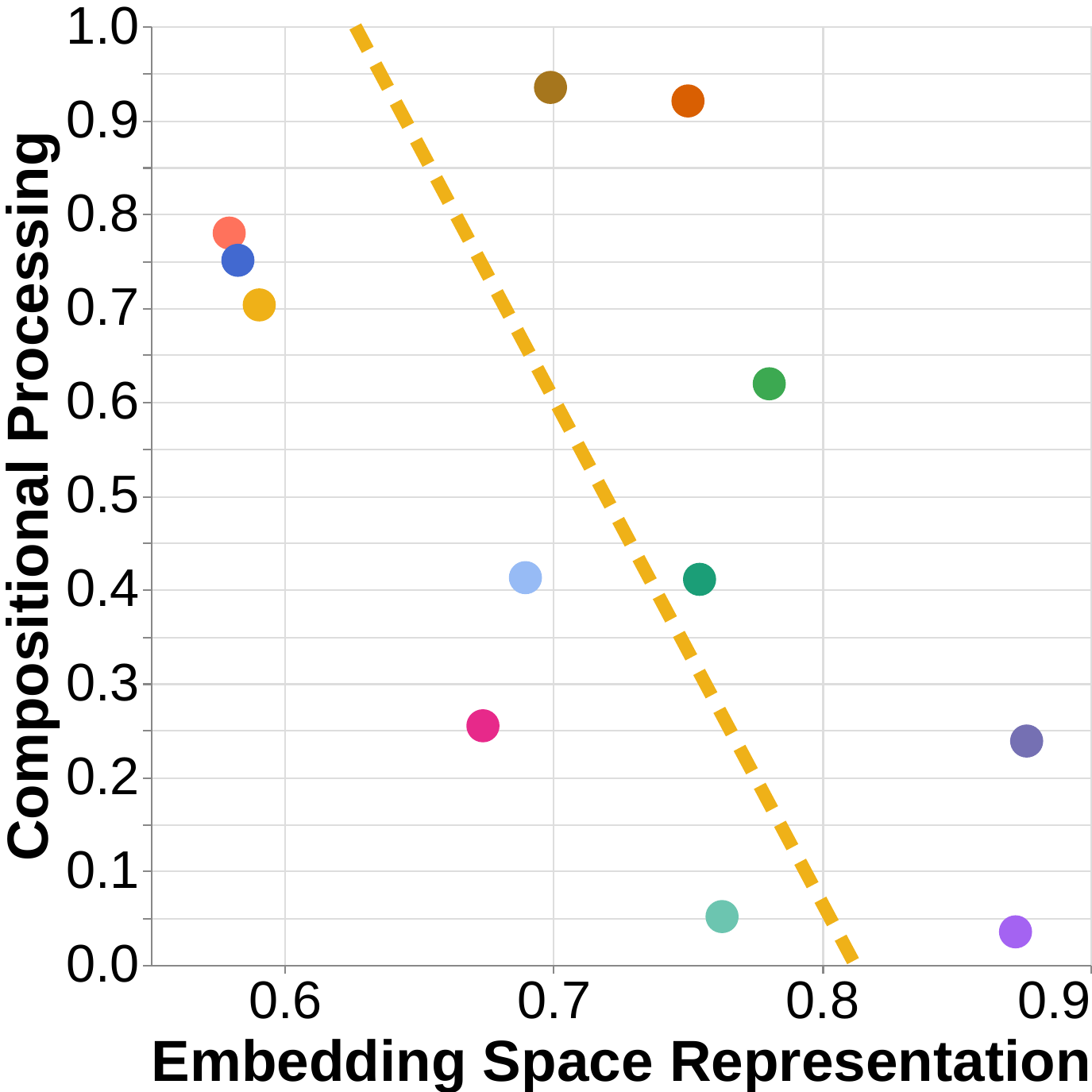}
    \caption{Correlation across tasks ($r^2 = \protect\input{artifacts/linear/llama_3_3b/token_identity/corr/lens_task}$) for linear embedding space representations and compositional processing (using the token identity patchscope metric).}
\end{figure}


\end{document}